\documentclass[11pt,twoside]{article}
\pdfoutput=1
\usepackage{enumitem}
\usepackage{varwidth} 
\usepackage{tikz}
\usetikzlibrary{arrows,fit,shapes,automata,topaths,petri}
\usetikzlibrary{decorations.pathreplacing,decorations.pathmorphing,calc,backgrounds,fit,plotmarks}
\usetikzlibrary{positioning,patterns}
\usepackage{pgfplots}
\usepackage{tikz-3dplot}
\usepackage{amssymb,graphicx,xspace}
\usepackage{graphicx}
\usepackage{subfig}
\usepackage{url}
\usepackage{relsize} 

\definecolor{light-gray}{gray}{0.43}

\usepackage{amsmath}

\usepackage{cancel}

\usepackage{todonotes}
\usepackage{subfig}
\usepackage[percent]{overpic}
\usepackage{algorithm}

\usepackage[noend]{algorithmic}
\usepackage{eqparbox} 
\newlength\myindent
\newcommand{\alstatespace}{4pt}
\newcommand\SSTATE{\vspace{\alstatespace}\STATE}

\graphicspath{{./diagrams/}}

\makeatletter
\tikzset{nomorepostaction/.code=\let\tikz@postactions\pgfutil@empty}
\makeatother

\usepackage{mathtools}
\mathtoolsset{showonlyrefs}
\usepackage{stylefile}

\global\long\def\outputIndex{j}
\global\long\def\dataIndex{i}

\global\long\def\latentIndex{j}

\global\long\def\dataDim{p}
\global\long\def\latentDim{q}

\global\long\def\numData{n}

\global\long\def\numInducing{m}

\global\long\def\dataScalar{y}
\global\long\def\dataVector{\mathbf{\dataScalar}}
\global\long\def\dataMatrix{\mathbf{\MakeUppercase{\dataScalar}}}

\global\long\def\latentScalar{x}
\global\long\def\latentMatrix{\mathbf{\MakeUppercase{\latentScalar}}}
\global\long\def\latentVector{\mathbf{\latentScalar}}

\global\long\def\kernelScalar{k}

\global\long\def\kernelMatrix{\mathbf{\MakeUppercase{\kernelScalar}}}

\global\long\def\weightScalar{w}
\global\long\def\weightVector{{\bf \weightScalar}}

\global\long\def\numData{n}

\global\long\def\dataDim{p}

\global\long\def\mappingFunction{f}
\global\long\def\mappingFunctionVector{\mathbf{\mappingFunction}}
\global\long\def\mappingFunctionMatrix{\mathbf{\MakeUppercase{\mappingFunction}}}

\global\long\def\pseudotargetScalar{u}
\global\long\def\pseudotargetVector{\mathbf{\pseudotargetScalar}}
\global\long\def\pseudotargetMatrix{\mathbf{\MakeUppercase{\pseudotargetScalar}}}

\global\long\def\weightScalar{w}
\global\long\def\weightVector{\mathbf{\weightScalar}}

\global\long\def\eye{\mathbf{I}}

\global\long\def\gaussianSamp#1#2{\mathcal{N}\left(#1,#2\right)}
\global\long\def\gaussianDist#1#2#3{\mathcal{N}\left(#1|#2,#3\right)}

\global\long\def\KL#1#2{\text{KL}\left( #1\,\|\,#2 \right)}

\global\long\def\Kff{\kernelMatrix_{\mappingFunctionVector \mappingFunctionVector}}
\global\long\def\Kuu{\kernelMatrix_{\inducingVector \inducingVector}}

\global\long\def\Kfu{\kernelMatrix_{\mappingFunctionVector \inducingVector}}
\global\long\def\Kuf{\kernelMatrix_{\inducingVector \mappingFunctionVector}}

\global\long\def\tr#1{\text{tr}\left(#1\right)}

\usepackage{color}
\usepackage{verbatim}

\definecolor{brown}{rgb}{0.9,0.59,0.078}
\definecolor{ironsulf}{rgb}{0,0.7,.5}
\definecolor{lightpurple}{rgb}{0.156,0,0.245}

\ifdefined\blackBackground
\definecolor{colorOne}{rgb}{0, 1, 1}
\definecolor{colorTwo}{rgb}{1, 0, 1}
\definecolor{colorThree}{rgb}{1, 1, 0}
\definecolor{colorTwoThree}{rgb}{1, 0, 0}
\definecolor{colorOneThree}{rgb}{0, 1, 0}
\definecolor{colorOneTwo}{rgb}{0, 0, 1}
\else
\definecolor{colorOne}{rgb}{1, 0, 0}
\definecolor{colorTwo}{rgb}{0, 1, 0}
\definecolor{colorThree}{rgb}{0, 0, 1}
\definecolor{colorTwoThree}{rgb}{0, 1, 1}
\definecolor{colorOneThree}{rgb}{1, 0, 1}
\definecolor{colorOneTwo}{rgb}{1, 1, 0}
\fi

\ifdefined\blackBackground

\else

\fi

\global\long\def\refsec#1{Section \ref{#1}}

\global\long\def\bfmu{\boldsymbol{\mu}}

\global\long\def\bfepsilon{\boldsymbol{\epsilon}}
\global\long\def\bfSigma{\boldsymbol{\Sigma}}

\global\long\def\bfPsi{\boldsymbol{\Psi}}

\global\long\def\bfPhi{\boldsymbol{\Phi}}

\global\long\def\bftheta{\boldsymbol{\theta}}
\global\long\def\bfTheta{\boldsymbol{\Theta}}

\global\long\def\bfa{\mathbf{a}}

\global\long\def\bff{\mathbf{f}}

\global\long\def\bfs{\mathbf{s}}
\global\long\def\bft{\mathbf{t}}
\global\long\def\bfu{\mathbf{u}}

\global\long\def\bfw{\mathbf{w}}
\global\long\def\bfx{\mathbf{x}}
\global\long\def\bfy{\mathbf{y}}

\global\long\def\bfzero{\mathbf{0}}

\global\long\def\bfB{\mathbf{B}}
\global\long\def\bfC{\mathbf{C}}

\global\long\def\bfI{\mathbf{I}}
\global\long\def\eye{\mathbf{I}}
\global\long\def\bfK{\mathbf{K}}

\global\long\def\bfM{\mathbf{M}}

\global\long\def\bfS{\mathbf{S}}

\global\long\def\bfU{\mathbf{U}}

\global\long\def\bfW{\mathbf{W}}
\global\long\def\bfX{\mathbf{X}}
\global\long\def\bfY{\mathbf{Y}}
\global\long\def\bfZ{\mathbf{Z}}

\global\long\def\cut#1{}

\global\long\def\detail#1{}

\renewcommand{\numData}{n}            
\renewcommand{\dataDim}{p}            
\renewcommand{\latentDim}{q}          
\renewcommand{\numInducing}{m}        

\newcommand{\jn}{\dataIndex}
\newcommand{\jd}{\outputIndex}
\newcommand{\jq}{\latentIndex}

\newcommand{\ie}{i.e.\ }
\newcommand{\eg}{e.g.\ }

\newcommand{\bb}{\beta^{-1}}

\newcommand{\Ts}{_*}

\newcommand{\calY}{\mathcal{Y}}
\newcommand{\calZ}{\mathcal{Z}}

\newcommand{\numViews}{K}
\newcommand{\viewYi}{\mathcal{A}}                
\newcommand{\viewZi}{\mathcal{B}}                
\newcommand{\viewY}{\mY^{\viewYi}}       
\newcommand{\viewZ}{\mY^{\viewZi}}       
\newcommand{\viewYTs}{\mY^{\viewYi}_*}   

\newcommand{\viewy}{\vy^{\viewYi}}       
\newcommand{\viewz}{\vy^{\viewZi}}       
\newcommand{\viewyTs}{\vy^{\viewYi}_*}   
\newcommand{\viewzTs}{\vy^{\viewZi}_*}   

\newcommand{\viewynTs}{\vy^{\viewYi}\nTs}    
\newcommand{\viewznTs}{\vy^{\viewZi}\nTs}    
\newcommand{\viewyinnTs}[1]{\vy^{\viewYi}\innTs{#1}}  
\newcommand{\viewzinnTs}[1]{\vy^{\viewZi}\innTs{#1}}  
\newcommand{\viewzs}{\sy^{\viewZi}}

  \newcommand{\n}{_{\dataIndex,:}}                
  \renewcommand{\d}{_{:,\outputIndex}}            
  \newcommand{\nd}{_{\dataIndex,\outputIndex}}    
  \newcommand{\q}{_{:,\latentIndex}}              

  \newcommand{\nTs}{_{\dataIndex,*}}                  
  \newcommand{\dTs}{_{*,\outputIndex}}                
  \newcommand{\ndTs}{_{*;\dataIndex,\outputIndex}}    
  \newcommand{\qTs}{_{*,\latentIndex}}                

  \newcommand{\inn}[1]{_{#1,:}}                
  
  \newcommand{\ind}[2]{_{#1,#2}}               
  \newcommand{\inq}[2]{_{#1,#2}}               

  \newcommand{\innTs}[1]{_{#1,*}}                

\newcommand{\nN}{\numData}
\newcommand{\nQ}{\latentDim}
\newcommand{\nD}{\dataDim}
\newcommand{\nM}{\numInducing}

\newcommand{\sx}{\latentScalar}
\newcommand{\sy}{\dataScalar}

\newcommand{\vx}{\latentVector}
\newcommand{\vy}{\dataVector}
\newcommand{\vf}{\mappingFunctionVector}

\newcommand{\vu}{\pseudotargetVector}

\newcommand{\vw}{\weightVector}

\newcommand{\mX}{\latentMatrix}
\newcommand{\mY}{\dataMatrix}
\newcommand{\mF}{\mappingFunctionMatrix}

\newcommand{\mU}{\pseudotargetMatrix}
\newcommand{\mK}{\kernelMatrix}

\newcommand{\xn}{\vx\n} 
\newcommand{\xq}{\vx\q} 

\newcommand{\yn}{\vy\n} 
\newcommand{\yd}{\vy\d} 

\newcommand{\ud}{\vu\d} 

\newcommand{\fd}{\vf\d} 

\tikzstyle{block}=[rectangle,rounded corners=3mm,text centered,draw=black,very thick,text centered,minimum width=7.5em]
\tikzstyle{edge} = [draw,-triangle 45,>=angle 90]
\tikzstyle{edge2} = [black,ultra thick, -triangle 45,>=angle 90]
\tikzstyle{vertex}=[circle,thick,draw=blue!75,fill=blue!20,minimum size=8mm]
\tikzstyle{vertex2}=[circle,thick,draw=blue!50,fill=blue!10,minimum size=8mm]
\tikzstyle{vertex3}=[circle,thick,draw=black!70,fill=black!20,minimum size=8mm]
\tikzstyle{vertex4}=[rectangle,thick,draw=black!70,fill=black!20,minimum size=8mm]
\tikzstyle{vertexml}=[circle,thick,draw=black!70,fill=black!90,text=white,minimum size=20mm]
\tikzstyle{vertexrv}=[circle,thick,draw=black!70,fill=white,text=black,minimum size=20mm]
\tikzstyle{vertexobs}=[circle,thick,draw=black!70,fill=black!30,text=black,minimum size=20mm,inner sep=0pt]

\tikzstyle{mm}=[rectangle,thick,draw=black!70,fill=black!20,rounded corners=2mm]
\pgfdeclarelayer{background}
\pgfdeclarelayer{foreground}
\pgfdeclarelayer{edges}
\pgfsetlayers{background,main,edges,foreground}

\newcommand{\dep}[1]{(#1)}

\newcommand{\bfF}{\mathbf{F}}

\newcommand{\calF}{\mathcal{F}}
\newcommand{\calA}{\mathcal{A}}
\newcommand{\calB}{\mathcal{B}}

\newcommand{\calU}{\mathcal{U}}
\newcommand{\calL}{\mathcal{L}}
\newcommand{\calQ}{\mathcal{Q}}

\newcommand{\calJ}{\mathcal{J}}

\newcommand{\fig}{Figure }

\renewcommand{\Kuu}{\mK_{uu}}
\renewcommand{\Kff}{\mK_{ff}}
\renewcommand{\Kfu}{\mK_{fu}}
\renewcommand{\Kuf}{\mK_{uf}}

\ShortHeadings{Nonparametric, Nonlinear  IBFA}{Damianou, Lawrence, Ek}
\firstpageno{1}

\begin{document}
\title{Multi-view Learning as a Nonparametric Nonlinear Inter-Battery Factor Analysis}

\author{\name Andreas Damianou \email andreas.damianou@sheffield.ac.uk\\
       \addr Department of Computer Science, \\
       Sheffield Robotics and\\
       Sheffield Institute for Translational Neuroscience\\
       University of Sheffield\\
       Sheffield, United Kingdom
       \AND
       \name Neil D. Lawrence \email neil@dcs.shef.ac.uk \\
       \addr Department of Computer Science and\\
       Sheffield Institute for Translational Neuroscience\\
       University of Sheffield\\
       Sheffield, United Kingdom
       \AND
       \name Carl Henrik Ek \email carlhenrik.ek@bristol.ac.uk \\
       \addr Department of Computer Science\\
       University of Bristol\\
       Bristol, United Kingdom
}

\maketitle

\begin{abstract}
Factor analysis aims to determine latent factors, or traits, which summarize a given data set. \emph{Inter-battery} factor analysis extends this notion to multiple views of the data. In this paper we show how a nonlinear, nonparametric version of these models can be recovered through the Gaussian process latent variable model. This gives us a flexible formalism for multi-view learning where the latent variables can be used both for exploratory purposes and for learning representations that enable efficient inference for ambiguous estimation tasks. Learning is performed in a Bayesian manner through the formulation of a variational compression scheme which gives a rigorous lower bound on the log likelihood. Our Bayesian framework provides strong regularization during training, allowing the structure of the latent space to be determined efficiently and automatically.
We demonstrate this by producing the first (to our knowledge) published results of learning from dozens of views, even when data is scarce.
We further show experimental results on several different types of multi-view data sets and for different kinds of tasks, including exploratory data analysis, generation, ambiguity modelling through latent priors and classification.
\end{abstract}

\begin{keywords}
representation learning, factor analysis, Gaussian processes, inter-battery factor analysis
\end{keywords}

\maketitle

\section{Introduction}
Automatically learning representations from observed data is a central aspect of machine learning. In supervised learning we wish to retain the information in the input data which is relevant for performing the supervised task, while in the unsupervised scenario we seek to discover the underlying patterns and structures in the data to associate them with some meaning. In this paper the focus is on the unsupervised scenario which we will refer to as \emph{representation learning}. A large body of work on representation learning considers approaches which build upon factor analysis.
The general formulation of factor analysis is often attributed to \citet{Spearman:1904tx} from his work in experimental psychology and his theory of ``general intelligence'' as a \emph{factor} to explain different human characteristics. However, the underlying ideas have a much longer history, stretching all the way back to the philosophers of ancient Greece. For a historical account of the philosophical ideas underlying factor analysis we would like to point the reader to the excellent work of \citet{Mulaik:1987un}.

A fundamental aspect of factor analysis is that the solution is unidentifiable. This means that there are infinitely many solutions which cannot be differentiated by the model. Even though all solutions might be completely equivalent from a statistical view point, from an experimental view point it is often important to be able to select a particular one. This is because different solutions are often attributed with different ``meanings''. However, with many possible solutions it is not clear how to find a criterion justifying one explanation of the data over another. Disregarding this problem has been referred to as the {\it the inductivist fallacy} \citep{Chomsky:1980ud}. In other words, additional information capable of differentiating between the solutions needs to be included in order to completely solve the problem.

The most common approach to factor analysis is to associate the importance of each factor relatively to the amount of variance it explains in the data. This is known as principal component analysis (PCA) \citep{Hotelling:1933ki}\footnote{Authors often allocate principal component analysis to \citet{Pearson:1901ud}, but he was trying to resolve a different problem: that of a symmetric regression problem, and the model is different. Hotelling was inspired by factor analysis and his model is similar to those of Spearman. This may seem to be splitting hairs because \emph{algorithmically} these models are fitted in the same manner, but the difference emerge when the models are non-linearized, leading to different algorithms, so the interpretation of the model is important.}. Another approach is to choose a representation that minimizes distortions between the data and the factors, an approach known as classical multi-dimensional scaling \citep{Mardia:multivariate79,Cox:2008uv}. A more recent approach is to relate the observed data and the discovered factors via a generative mapping for which we additionally include a preference for its functional form. Gaussian process latent variable models \citep[GP-LVMs][]{Lawrence:2005vk} achieve this in a probabilistic manner by employing  Gaussian process priors. Specifically, the GP-LVM is a latent variable model where the distribution over the $\nN$, $\nD-$dimensional observed instances $\mathbf{Y}\in\Re^{\nN\times \nD}$ is parameterized using a $\nQ-$dimensional latent space $\mathbf{X}\in\Re^{\nN \times \nQ}$, that is, the model likelihood takes the form $p(\mathbf{Y}|\mathbf{X})$. The Gaussian process mapping from the variables\footnote{To keep the notation unclutterred we refer to a multivariate variable and its collection of instantiations using the same notation, \eg $\bfX$, since the context always provides disambiguation.} $\bfX$ to the variables $\bfY$ is analytically integrated out. GP-LVMs are traditionally used for dimensionality reduction by selecting $\nQ \ll \nD$. Therefore, the challenge is to find a low-dimensional latent variable $\mathbf{X}$ which parameterizes, or retains, all the variance in the observed data $\mathbf{Y}$. In other words, in the generative paradigm assumed by the GP-LVM the latent variables play the r\^ole of factors.
In this paper we follow this research direction and propose a model which solves the representation learning problem by building upon the GP-LVM back-bone. 

\subsection{Multi-view Learning}

It is often the case that our data come from several different observation streams (also called ``views'' or ``modalities'') which arise from the same situation or phenomenon. The representation learning modelling approach has then to be extended to the \emph{multi-view} scenario. In the generative paradigm, the task is then to find a latent representation which consolidates all the different views. The notion of supervised and unsupervised learning becomes ``blurred'' in the multi-view scenario, as we can consider that one of the views is providing supervision for the representation of the other \citep{Ham:2003vz}. 
Since all views jointly characterize the same phenomenon, we can consider that there is \emph{shared} information amongst them (or amongst subsets of them) by which they are all related. Further, we also expect that there might exist unique or \emph{private} variations in each view, \ie information which is specific to a single data stream. Therefore, a multi-view factor analysis approach aims at explaining the complex multi-view data through a set of simpler factors which are separated to represent the shared and private variations. Inter-battery factor analysis (IBFA) \citep{Tucker:1958ul} is such an approach.
This model, as many other factor analysis models, was first proposed in experimental psychology. Much later it was rediscovered in machine learning \citep{Klami:2006jt,Ek:2008up} within approaches which encapsulate a segmented (factorized) latent space.

To make the above intuitions clearer, let us consider two views $\mathbf{Y}^{(1)}\in\Re^{\nN \times \nD_1}$ and $\mathbf{Y}^{(2)}\in\Re^{\nN \times \nD_2}$ where there is an implicit correspondence (alignment) between each pair of instances belonging to the two views. The challenge is to learn a joint representation of the distribution over \emph{both} sets of data parameterized by a single latent variable, i.e. $p(\mathbf{Y}^{(1)},\mathbf{Y}^{(2)}|\mathbf{X})$. Similarly to how a single-view latent variable model aims at preserving all relevant data variance in the latent space $\bfX$, a factorized latent variable model aims at representing the variance from all views in a segmented latent space $\mathbf{X} = [\mathbf{X}^{(1)},\mathbf{X}^{(2)},\mathbf{X}^{(1,2)}]$, where $\mathbf{X}^{(1,2)}$ encodes the variations that exist in both views and $\mathbf{X}^{(1)}$ and $\mathbf{X}^{(2)}$ represent the variations that are unique to each respective view. 
Efficient segmentation of the available variation into shared and private latent variables is crucial. Compared to learning a separate model for each view (\ie absence of shared information modelling), a factorized model provides a more efficient representation by re-using the shared variations to represent the shared signal of the views. In the other end of the spectrum, if we build a ``factorized'' model which fails to account for private factors we will end up with an over-parameterized latent representation which will be forced to relate irrelevant variations to each view. This detrimental effect is discussed in the next chapter.

\begin{figure}[t]
  \centering
  \includegraphics[width=1\textwidth]{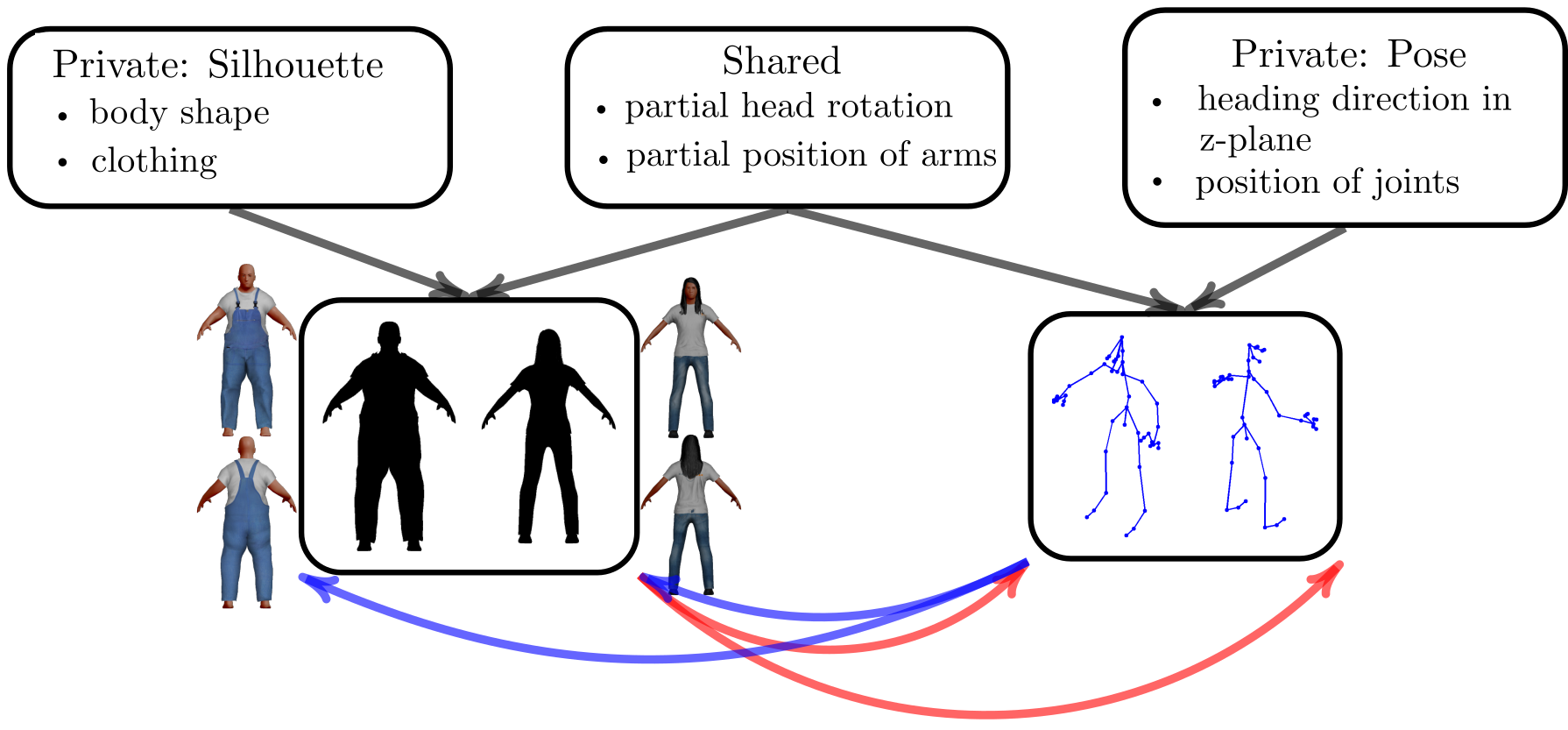}
  \vspace{2pt}
  \caption{
      The above figure shows the benefit of a factorized latent representation using the example of 3D human pose and silhouette images. The blue and red arrows show the multi-modal relationship that exists in both ``directions''. By following the direction of the \textcolor{blue}{blue arrows} it is not possible to determine information which is private to the silhouette, such as the shape of the body. Analogously, the silhouette for a person facing towards or away from the camera is the same, which means that the heading angle cannot be determined for this pose by following the \textcolor{red}{red arrows} direction. 
  }
  \label{fig:intro}
\end{figure}

The benefits of a factorized latent representation which separately models the shared and the private information from multi-view data is best motivated through an example. Consider the task of trying to infer the full three-dimensional pose of a human body from its two-dimensional silhouette image. This is a very challenging task as it is not possible to differentiate between all different poses by looking only at the silhouette; consider, for example, the case where the heading angle of a person walking is perpendicular to the image plane. Relating this to the factorized latent variable model, the pose parameters that are indeterminable from the silhouette view will be separately represented as private variables with respect to the pose view; the remaining determinable parameters will be contained in the shared portion. Such a factorization facilitates intuitive inference, as the model clearly tells us which part of the pose space we can determine from the silhouette and which part cannot be determined without additional information. Therefore, such a model can handle multi-modal estimation tasks when the input data can only partially determine the output, as illustrated in Figure \ref{fig:intro}.

In this paper we present a formulation of the inter-battery factor analysis model that allows for nonlinear and probabilistic mappings. By relaxing the assumption of linear mappings, our model manages to learn the complex relationships that exist in real, noisy data. Our approach inherits the attractive nonparametric properties of Gaussian processes by building upon a Gaussian process latent variable model. Specifically, we select separate generative mappings to relate the latent space $\bfX$ to each of the views, $\bfY^{(k)}$. These mappings are governed by Gaussian processes in which we introduce additional kernel parameters that, in combination with our inference procedure, allow the model to ``switch off'' latent dimensions which are deemed unnecessary by each view independently. This provides an automatic factorization of the latent space. Therefore, we show how inter-battery factor analysis naturally manifests itself as a Gaussian process latent variable model. We further develop a Bayesian inference scheme and demonstrate that it is crucial in eliminating information redundancy in the latent space and, hence, in achieving efficient, automatic latent space factorization. We build on our previous work \citep{Damianou:2012wu} and further we reinterpret the model as inter-battery factor analysis and consider extensions to more than two views.

The main contributions of our paper are the following:
\begin{itemize}[noitemsep]
  \item a probabilistic, nonlinear, nonparametric formulation of inter-battery factor analysis,
  \item a variational framework for approximate Bayesian learning,
  \item a framework for introducing priors that encourage specific latent space configurations,
  \item a multi-view latent consolidation model that naturally extends beyond two views; to our knowledge, we present the first work which shows results of a factorized generative model with truly large number of views.
\end{itemize}

In the next section we will describe the relevant related work, placing the proposed model in context. We will then provide an introduction to Gaussian processes in general and Gaussian process latent variable models in specific. \refsec{sec:mrd} represents the central part of the paper where we will describe and motivate the model we propose. We will then proceed in \refsec{sec:experiments} to show experimental results of the proposed model comparing the approach to competing methods. Finally, \refsec{sec:conclusions} concludes the paper and outlines directions of future work.

\section{Related Work}\label{sec:relatedwork}

A large portion of the multi-view learning literature has been motivated by canonical correlation analysis (CCA) \citep{Hotelling:1936wo}.\footnote{Interestingly, whilst Hotelling motivated principal components very clearly as a latent variable model, following in the factor analysis tradition, he motivated canonical correlates analysis through a form of normalized correlation between the data. It was left to \citet{Bach:2005wz} to deliver a probabilistic interpretation (also explored by \citet{Kalaitzis:2012wwa} from a latent variable perspective).} Assuming two data views, in CCA the goal is to find two linear transformations, one for each view of the data, which align directions with significant correlation between the views. In an application scenario these transformations are then used as a means of performing feature selection, where each view is projected onto the subspace that retains the maximum correlation. This model is often applied in a scenario where one wants to extract information about one view given the other. The same approach has been extended to learn transformations involving multiple views \citep{Rupnik:2010ti}. Traditionally, CCA assumes that the transformations are linear, something which constrains the applicability of the method. To that end, several nonlinear extensions have been suggested. One approach is to perform the alignment in an implicit feature space induced by a kernel function \citep{Kuss:2003wp}. Using CCA in an unsupervised manner to find a joint representation of two data sets assumes that the maximally correlated directions are also directions of maximal variance. However, correlation as a measure can be rather misleading, as it does not depend on the actual statistical variance that is represented by the aligned directions. This means that the aligned subspace after projection might only represent a small portion of the data variance. In many application scenarios, correlation between views is manifested not only in the signal but also in the noise; this renders CCA-based approaches problematic, because they are prone to learning representations that amplify the noise.

One approach for overcoming the limitations of CCA is to encode the views in such a manner that the representation encapsulates as much of the data variance as possible. This can be seen as a \emph{joint dimensionality reduction problem} which is associated with the task of discovering low-dimensional manifolds given the observed views. Several past approaches have followed this idea. Given two views which are in partial correspondence, \citet{Ham:2003vz}  proceed by assuming that the variations in each view can be parameterized by the same manifold. The authors applied three different single-view dimensionality reduction methods to learn this manifold. Similarly, \citet{Ham:2005vs} also exploit partial correspondences and learn a manifold based on preserving neighbourhood structures in the observed data. Rather than learning an explicit shared representation, \citet{Ham:2006ij} assume a separate low-dimensional manifold for each data view and seek to also learn a mapping between them. Each of the above three methods takes inspiration from the approaches previously developed for spectral dimensionality reduction from a single view \citep{Tenenbaum:2000jp,Yan:2007dz,Roweis:2000ey,Weinberger:2006vi,Lawrence:2010vp}. These methods build constraints from local statistics in the data by aiming to find a manifold that preserves the \emph{local} distances in the data set. However, computation of such distances is not always possible, for example when there is missing data.
 Further, noise in the data can result in ``topological instability'' \citep{Balasubramanian:2002iy}, for example when two noisy points appear close together when in reality they should be embedded far apart.

 By directly modelling the generative mapping this problem can be avoided. \citet{Bach:2005wz} showed how CCA can be derived through a probabilistic latent variable model, while \citet{Kalaitzis:2012wwa} showed that it can also arise through formulation of a form of residual component analysis. Both models interpret the canonical correlates as directions induced in a latent variable model. In common with other generative approaches, rather than trying to map from the data to the latent space, they directly map from the latent to the data space by building a probabilistic model of the data.

In practice, latent variable modelling can be challenging: without constraints on the generative mapping the problem is ill-posed, as there is an infinite number of possible combinations of mappings and representations that could have generated the observed data. This means that, to reach a solution, restrictions on the model need to be incorporated. For single view data, \citet{Tipping:1999uo} proposed a model based on linear mappings, while \citet{Bishop:1998fl} suggested a nonlinear version using a mixture approach. The work presented in this paper is formulated as a multi-view extension of the Gaussian process latent variable model \citep[GP-LVM][]{Lawrence:2005vk}, which assumes that the mappings constitute samples from a Gaussian process. The GP-LVM, graphically illustrated in Figure \ref{fig:sgplvms}(a), provides a probabilistic nonlinear generalization of single view factor analysis where the unidentifiabilty is circumvented by including a preference (\ie restriction) in the generative mapping. The approach was adapted to the multi-view case by \citet{Shon:2006wr}, who proposed a model that allows different generative mappings for each observation space from a single shared latent space, as is illustrated in Figure \ref{fig:sgplvms}(b).

Both the spectral and generative families of methods for multi-view modelling described above assume that all views share the same variations, as they leave no way of incorporating variance that is specific to one of the particular views. For many types of data this is a very crude assumption, as it means that all views share all generating parameters. To exemplify this problem let us revisit the human pose and silhouette example and consider the task of determining the pose-related parameters from the silhouette view. Clearly the same pose parameters can generate different silhouettes depending on the ``shape'' of the human.
Similarly, a specific silhouette could have been generated by different poses, \ie it is the pose-specific information which disambiguates the silhouette and not the shared information for pose/silhouette. Therefore, a shared latent space is needed to relate the two views and a private latent space is needed to disambiguate the prediction. A latent variable approach which does not consider private spaces will result in one of two undesirable scenarios: it will either completely avoid representing the non-shared variations or it will retain those variations but at the cost of reducing the quality of the estimate for the views that are not explained by them.

A first attempt to rectify this problem was suggested by \citet{Ek:2007uo} who modified the model proposed by \citet{Shon:2006wr} such that the latent space was represented as a parametric mapping of the variations in one of the views. This was achieved by the incorporation of a back-constraint \citep{Lawrence:2006wr} from one view to the latent space. This model is illustrated in Figure \ref{fig:sgplvms}(c) for the case of two observation views, $\bfY^{(1)}$ and $\bfY^{(2)}$. The back-constraint from view $\bfY^{(2)}$ implies a bijective relationship to which the generation of the remaining view, $\bfY^{(1)}$ has to adapt through discarding the variations which are not present in $\bfY^{(2)}$. The purpose of this was to force the latent space to ignore variations that did not exist in the non-constrained space. This approach should be considered as feature selection rather than consolidation of the views, as it directly uses one view as supervision to determine which variations are important in the other. Therefore, this approach could still not solve the problem of polluting the latent space with non-shared variations.

Another possibility to accommodate both shared and private variations within the same model is to learn a latent representation which factorizes the latent space into factors such that the shared and private variations are represented by separate latent variables. This is the previously discussed IBFA approach \citep{Tucker:1958ul}, which consists of a consolidated representation of the data through a structure which separates shared and private variations. This strategy has been adapted to a generative framework by the pioneering work of \citet{Klami:2006jt}. The ideas of factorized latent structure for multi-view learning have been adopted in several works \citep{Jia:2010tua,Leen:2008vc,Salzmann:2010vh} for latent variable models and also applied to topic models \citep{Jia:2011uy,Virtanen:2012ur,2013arXiv1301.3461Z}. The work that applied this idea through the GP-LVM comes from \citet{Ek:2008up} and is graphically illustrated in Figure \ref{fig:sgplvms}(d).

In practice, learning the factorization of $\bfX$ is extremely challenging. This has lead to several different heuristics aimed at encouraging a specific separation of the variations. \citet{Ek:2008up} proposed a spectral algorithm which, in combination with CCA, could factorize two views. In \citep{Salzmann:2010vh} a regularizer that encourages orthogonality between the subspaces was proposed, in order to reach a solution. Despite promising results, the method required a rather ad-hoc annealing scheme during learning. Further, orthogonality of sub-spaces as a constraint on a multi-view probabilistic model is an idea that lacks theoretical support. Indeed, the orthogonality of principal component analysis arises only as a constraint to resolve its rotational ambiguity, but when maximum likelihood solutions are sought, orthogonality cannot in general be shown to constitute a sensible constraint, even if it may provide a useful initialization. \citet{Jia:2010tua} used a sparse linear model which reduces the complexity during model learning. The results of this \emph{linear} method were better than the nonlinear method of \citet{Salzmann:2010vh}. This indicates that the latter model, even though it is representationally more powerful, is not always able to exploit this representational power due to difficulties in optimization. Recently a Bayesian formulation of the linear model was introduced \citep{Klami:2012tf}.

\newcommand{\myvertexstyle}{vertexrv}

\begin{figure}[t]
  \begin{center}
    \scalebox{0.39}{
    \begin{tikzpicture}[node distance=3.5cm]





    \begin{scope}[yshift=-3.5cm]
      \node(y0)[vertexobs] at (0,0) {\Huge $\mathbf{Y}$};
      \node(x0)[\myvertexstyle,above of=y0]{\Huge $\mathbf{X}$};
      \node(ty)[\myvertexstyle,left of=y0]{\Huge $\mathbf{\theta}$};
    \draw[edge](x0)--(y0);
    \draw[edge](ty)--(y0);
    \node [below of=y0,yshift=1cm] {\Huge (a)};

  \end{scope}

  \begin{scope}[xshift=4cm]
      \node(y0)[vertexobs] at (0,0) {\Huge $\mathbf{Y}^{(1)}$};
      \node(x0)[\myvertexstyle,above right of=y0]{\Huge $\mathbf{X}$};
      \node(y1)[vertexobs,below right of=x0]{\Huge $\mathbf{Y}^{(2)}$};
      \node(t1)[\myvertexstyle,below of=y0]{\Huge $\mathbf{\theta}^{(1)}$};
      \node(t2)[\myvertexstyle,below of=y1]{\Huge $\mathbf{\theta}^{(2)}$};

    \draw[edge](x0)--(y0);
    \draw[edge](x0)--(y1);
    \draw[edge](t1)--(y0);
    \draw[edge](t2)--(y1);

    \path (t1) -- node (b) {} (t2);
    \node [below of=b,yshift=1cm] {\Huge (b)};

  \end{scope}
  \begin{scope}[xshift=12.6cm]
      \node(y0)[vertexobs] at (0,0) {\Huge $\mathbf{Y}^{(1)}$};
      \node(x0)[\myvertexstyle,above right of=y0]{\Huge $\mathbf{X}$};
      \node(y1)[vertexobs,below right of=x0]{\Huge $\mathbf{Y}^{(2)}$};
      \node(t1)[\myvertexstyle,below of=y0]{\Huge $\mathbf{\theta}^{(1)}$};
      \node(t2)[\myvertexstyle,below of=y1]{\Huge $\mathbf{\theta}^{(2)}$};

    \draw[edge](x0)--(y0);
    \draw[edge](x0)--(y1);
    \draw[edge](t1)--(y0);
    \draw[edge](t2)--(y1);
    \draw[edge,dashed] (y1) to [bend right] (x0);

    \path (t1) -- node (b) {} (t2);
    \node [below of=b,yshift=1cm] {\Huge (c)};

  \end{scope}

  \begin{scope}[xshift=23cm]
      \node(y0)[vertexobs] at (0,0) {\Huge $\mathbf{Y}^{(1)}$};
      \node(x0)[\myvertexstyle,above right of=y0]{\Huge $\mathbf{X}^{(1,2)}$};
      \node(x1)[\myvertexstyle,above left of=y0]{\Huge $\mathbf{X}^{(1)}$};
      \node(y1)[vertexobs,below right of=x0]{\Huge $\mathbf{Y}^{(2)}$};
      \node(x2)[\myvertexstyle,above right of=y1]{\Huge $\mathbf{X}^{(2)}$};
      \node(t1)[\myvertexstyle,below of=y0]{\Huge $\mathbf{\theta}^{(1)}$};
      \node(t2)[\myvertexstyle,below of=y1]{\Huge $\mathbf{\theta}^{(2)}$};

    \draw[edge](x0)--(y0);
    \draw[edge](x0)--(y1);
    \draw[edge](t1)--(y0);
    \draw[edge](t2)--(y1);
    \draw[edge](x1)--(y0);
    \draw[edge](x2)--(y1);



    \path (t1) -- node (b) {} (t2);
    \node [below of=b,yshift=1cm] {\Huge (d)};

  \end{scope}
\end{tikzpicture}

    }
  \end{center}
  \caption{
      The development of the GP-LVM models from the initially proposed \citep{Lawrence:2005vk} (panel (a)) with a single observation and latent space through the first models adapted to a multi-view scenario with several observation spaces \citep{Shon:2006wr,Ek:2007uo} (panels (b),(c)) to the first proposed model with a factorized latent space \citep{Ek:2008up} (panel (d)). Shaded nodes represent observed variables; $\bftheta$ denotes the kernel parameters of the Gaussian process mappings; the dashed line represents a back-constraint.}
\label{fig:sgplvms}
\end{figure}
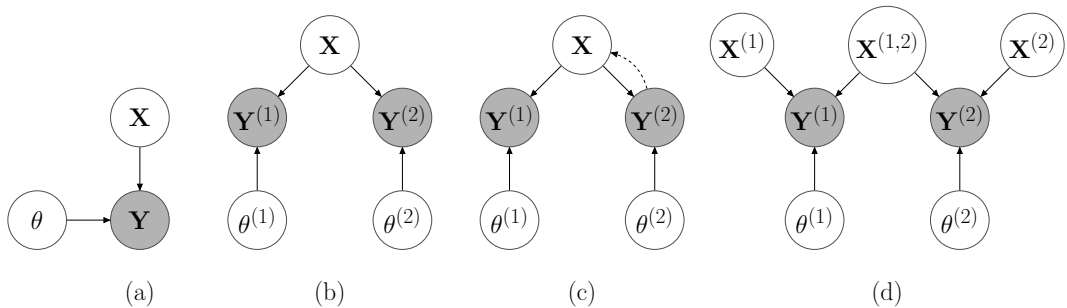

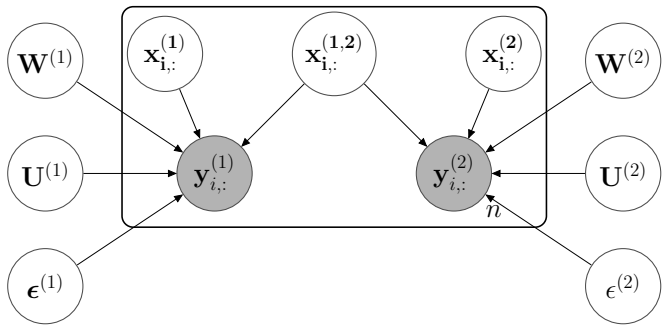
\begin{figure}
  \centering
  \scalebox{0.5}{

\tdplotsetmaincoords{90}{90}
\begin{tikzpicture}[node distance=4.5cm,tdplot_main_coords]
  \node(x)[vertexrv] {\huge $\mathbf{x^{(1,2)}_{i,:}}$};
  \node(x1)[vertexrv,left of=x] {\huge $\mathbf{x^{(1)}_{i,:}}$};
  \node(x2)[vertexrv,right of=x] {\huge $\mathbf{x^{(2)}_{i,:}}$};

  \node(y1)[vertexobs,below left of=x] {\huge $\mathbf{y}^{(1)}_{i,:}$};
  \node(y2)[vertexobs,below right of=x,label={[name=l]south east:\huge $\nN$}] {\huge ${\mathbf{y}^{(2)}_{i,:}}$};

  \node(u1)[vertexrv,left of =y1] {\huge ${\mathbf{U}^{(1)}}$};
  \node(u2)[vertexrv,right of =y2] {\huge ${\mathbf{U}^{(2)}}$};

  \node(w1)[vertexrv,above of =u1,yshift=-1.5cm] {\huge ${\mathbf{W}^{(1)}}$};
  \node(w2)[vertexrv,above of =u2,yshift=-1.5cm] {\huge ${\mathbf{W}^{(2)}}$};

  \node(e1)[vertexrv,below of =u1,yshift=1.5cm] {\huge ${\boldsymbol {\epsilon}^{(1)}}$};
  \node(e2)[vertexrv,below of =u2,yshift=1.5cm] {\huge ${{\epsilon}^{(2)}}$};

  \draw[edge](x) to (y1);
  \draw[edge](x) to (y2);
  \draw[edge](x1) to (y1);
  \draw[edge](x2) to (y2);
  \draw[edge](w1) to (y1);
  \draw[edge](w2) to (y2);
  \draw[edge](u1) to (y1);
  \draw[edge](u2) to (y2);
  \draw[edge](e1) to (y1);
  \draw[edge](e2) to (y2);

  \node(b1)[block][fit = (x)(x1)(x2)(y1)(y2)(l)]{};
\end{tikzpicture}

  }
  \caption{
      The graphical model for the inter-battery factor analysis model. The two observed sets of data, $\mathbf{Y}^{(1)}$ and $\mathbf{Y}^{(2)}$, are modelled using two latent variables $\mathbf{X}^{(1)}$ and $\mathbf{X}^{(2)}$ which describe the private variations, as well as a single latent variable $\mathbf{X}^{(1,2)}$ which describes the shared variations. The mappings from the latent to the observed data are linear and parameterized by $\mathbf{W}^{(1),(2)}$ and $\mathbf{U}^{(1),(2)}$ for the shared and the private spaces respectively.
  \label{fig:IBFA}
}
\end{figure}

The models with a segmented latent structure, such as the ones mentioned in the previous two paragraphs, are in this paper referred to as \emph{factorized} latent variable models, to adhere to the formalism and emphasize that they constitute factor analyzers. One way of seeing these models is as extensions of the IBFA approach \citep[see][]{Klami:2012tf}, which we will now introduce more formally. Consider the case where we seek to relate two views $\bfY^{(1)} = \{ \bfy^{(1)}_{\jn,:} \}_{\jn=1}^\nN$ and $\bfY^{(2)} = \{ \bfy^{(2)}_{\jn,:} \}_{\jn=1}^\nN$ to and through a latent space $\bfX = \{ \bfx_{\jn,:} \}_{\jn=1}^\nN$. To achieve this, the probabilistic IBFA approach considers a factorization of the latent variables (\ie columns of $\bfX$ or, equivalently, dimensions of each latent instance $\bfx_{\jn,:}$) together with the definition of the following model:
\begin{equation}
\begin{aligned}
\dataVector^{(1)}_{\dataIndex,:} &= \bfW^{(1)} \latentVector^{(1,2)}_{\dataIndex,:} + \bfU^{(1)} \latentVector^{(1)}_{\dataIndex,:} + \bfepsilon^{(1)}_{\dataIndex,:} \\
\dataVector^{(2)}_{\dataIndex,:} &= \bfW^{(2)} \latentVector^{(1,2)}_{\dataIndex,:} + \bfU^{(2)} \latentVector^{(2)}_{\dataIndex,:} + \bfepsilon^{(2)}_{\dataIndex,:} ,
\end{aligned}
\end{equation}
where the latent variables $\latentVector$ are assigned a standard normal distribution and the noise terms are also Gaussian with diagonal covariance matrices: $\bfepsilon_{i,:}^{(1)} \sim \gaussianSamp{\bfzero}{\bfSigma^{(1)}}$ (and similarly for $\bfepsilon_{i,:}^{(2)}$). The latent variables are factorized into those shared across views, $\latentVector^{(1,2)}_{\dataIndex,:}$, and those specific to each view, $\latentVector^{(1)}_{\dataIndex,:}, \latentVector^{(2)}_{\dataIndex,:}$. Separate factor loadings (namely $\bfW^{(k)}$ and $\bfU^{(k)}, k=1,2$) map the latent factors linearly to the observation space with the addition of noise. Notice that by setting $\bfU^{(1)}, \bfU^{(2)}$ to be matrices of zeros and allowing $\bfSigma^{(1)}, \bfSigma^{(2)}$ to be non-diagonal, we obtain a probabilistic version of CCA \citep{Bach:2005wz}, which does not make the factorization of the latent space explicit. Figure \ref{fig:IBFA} illustrates this model graphically.

In this paper we propose a probabilistic and nonlinear formulation of the IBFA model framed as a GP-LVM model. The proposed paper extends the shared multi-view approach proposed by \cite{Shon:2006wr} using the factorized structure described by \cite{Ek:2008vr}. The main contribution of the paper is a fully Bayesian treatment of the model which avoids reliance on heuristics, such as the regularizers proposed by \cite{Salzmann:2010vh} for learning the factorization. Not only does this allows us to learn the structure of the factorization but also the dimensionality of each of the subspaces, which is a free parameter of the maximum likelihood. In contrast to \citep{Klami:2012tf} our model is nonlinear and fully nonparametric. The nonparametric nature is directly inherited from the use of a Gaussian process prior over the mappings between the latent space and the data space. Before we continue, we explain the notation used throughout the paper and, subsequently, we provide a more detailed description of Gaussian processes and their use in data consolidation and dimensionality reduction.

\subsection{Notation}
Throughout this paper, matrices are represented using boldface capital letters, vectors using boldface, lowercase letters and scalars using regular type face, lowercase letters. The indexing is as follows: given $K$ views we will refer to the observed outputs in view $k$ as $\mathbf{Y}^{(k)}$, and this collection of variables will be referred to as the ``data view $k$''. The individual points in a matrix are stored by rows, \eg $\mathbf{Y}^{(k)} = [\mathbf{y}^{(k)}\inn{1},\ldots,\mathbf{y}^{(k)}\inn{\nN}]^{\textrm{T}}$. We will also denote dimensions (columns) of $\bfY^{(k)}$ by $\bfy^{(k)}\d$ whereas $y^{(k)}\nd$ denotes dimension $\jd$ of the $\jn$th point. When we refer to test points the notation for rows, columns and single elements of a matrix becomes $\bfy^{(k)}\nTs$, $\bfy^{(k)}\dTs$ and $y^{(k)}\ndTs$ respectively. Any matrix indexing will follow these conventions. $\mathbf{Y}^{\mathcal{A}}$ indicates a set of output data corresponding to a set of views $\mathcal{A}\subseteq(1,\ldots,K)$. As a short hand notation for referring to data from \emph{all} views we will use $\mathcal{Y}$ i.e. $\mathcal{Y}\equiv\mathbf{Y}^{(1,\ldots,K)} = \{ \bfY^{(1)}, \bfY^{(2)}, \ldots, \bfY^{(K)}\}$. 
Latent variables are also called inputs because of the dependencies in the graphical model. The indexing  
$\mathbf{X}^{(k)}$ represents the latent space which is \emph{private} for view $k$, that is, in the generative model the information in $\bfX^{(k)}$ is used to generate $\bfY^{(k)}$ and no other view. Similarly, $\mathbf{X}^{(k,l)}$ represents the latent space which is \emph{shared} between view $k$ and $l$.

\section{Gaussian Process Modeling}
\label{sec:gp}
A Gaussian process (GP) is a stochastic process where each finite set of realizations follows a joint Gaussian distribution. The process is parametrized by its mean $\mu(\bfx)$ and covariance function $k(\bfx,\bfx')$, where $\bfx, \bfx'$ denote instances taken from the input domain of the GP. A very common choice for the mean function is to constitute a constant vector of zeros, of appropriate size. The covariance function domain is an infinite set and is parametrized by $\bftheta$, which is called a \emph{hyperparameter set} with respect to the model.  Due to the cardinality of the input domain, a GP can be used as a flexible prior over functions \citep{Rasmussen:2005te} which allows for a fully Bayesian and analytic treatment over the space of functions.

More formally, consider the situation where we wish to model the relationship between two variables $\bfX$ and $\bfY$ as a function $f$, \ie $f:\bfX \rightarrow \bfY$.
Given corresponding multivariate instantiations $\bfX=[\bfx\inn{1},\ldots,\bfx\inn{\nN}]^\top$ and $\bfY=[\bfy\inn{1},\ldots,\bfy\inn{\nN}]^\top$ where $\bfx\n \in\Re^\nQ$ and $\bfy\n \in\Re^\nD$, we assume that each dimension of an instance $\bfy\n$ is generated by an independent function $f\d$ from input $\bff\n$ with the addition of Gaussian noise $\epsilon\nd \sim \gaussianSamp{0}{\beta^{-1}}$, according to:
\begin{equation}
\label{eq:generative}
y\nd = f\d(\bfx\n) + \epsilon\nd .
\end{equation} 
All mapping functions are assigned a Gaussian process prior with common hyperparameters, that is, $f\d \sim \mathcal{GP}(\bfzero,k(\bfx,\bfx'))$. The Gaussian process prior means that the set of all available function instantiations $\bff\d$ are distributed as 
\begin{equation}
\label{eq:gpPrior}
p(\bff\d | \bfX, \bftheta) = \mathcal{N}(\bff\d | \bfzero, \Kff),
\end{equation}
 where $\Kff = k(\mX,\mX)$ denotes the covariance matrix obtained after evaluating the covariance function $k$ on all available instances $\bfX$.
The choice for a particular covariance function can result from a model selection routine (\eg cross-validation), or can be seen as an assumption during model design. A popular covariance function which encodes the assumption of smoothness in the input space is the infinitely differentiable exponentiated quadratic (EQ) one, also known as RBF:
\begin{equation}
\label{eq:EQ}
k_\mathsmaller{{\text{EQ}}}(\vx\inn{i},\vx\inn{r}) = \sigma^2_\mathsmaller{\text{EQ}} \exp \left( -\frac{w}{2} \sum_{\jq=1}^\nQ (\sx\inq{i}{\jq} - \sx\inq{r}{\jq})^2 \right),
\end{equation}
where the kernel variance $\sigma^2_\mathsmaller{\text{EQ}}$ and the so-called squared inverse length-scale $w$ constitute the hyparparameter set $\bftheta$.

Further, the assumption about Gaussian noise in the generative model of equation \eqref{eq:generative} means that the outputs are distributed according to $$p(\bfy\d | \bff\d, \beta) = \mathcal{N}(\bfy\d | \bff\d, \bb \bfI).$$ Then, the marginal likelihood can be written in closed form as
\begin{equation}
  p(\bfy\d|\bfX,\bftheta) = \int_{\bff\d} p(\bfy\d | \bff\d) p(\bff\d | \bfX, \bftheta),
\label{eq:gp_marginallikelihood}
\end{equation}
where we dropped conditioning on $\beta$ from the expression, for clarity.
Both of the distributions appearing in the integral are Gaussian and, due to the self-conjugacy of the Gaussian distribution, the marginal likelihood is also a Gaussian.
 Traditionally, the parameters $\bftheta$ of the \cal{GP} and the noise precision, $\beta$, are learned together by maximizing the above marginal likelihood. 

\subsection{Gaussian Processes Latent Variable Models}
The Gaussian process latent variable model \citep[GP-LVM][]{Lawrence:2005vk} is an unsupervised learning framework for using GP priors for dimensionality reduction. 
Given a set of high dimensional data $\bfY \in \Re^{\nN \times \nD}$, the aim is to explain them through a set of low dimensional variables $\bfX \in \Re^{\nN \times \nQ}$.
The setting is similar to that of standard GP
regression but now only the outputs $\bfY$ are observed, whereas the inputs
$\bfX$ are considered to be \emph{latent}. Each dimension of the observations,
$\bfy\d$, is assumed to be generated by the same latent input variable $\bfX$
via a GP mapping, as in equation \eqref{eq:generative}. To better understand
how dimensionality reduction is achieved by the GP-LVM, we write here the full
generative model with all appropriate factorizations: 
\begin{align}
p(\bfY|\bfX,\bftheta) &= \int_\bfF p(\bfY|\bfF) p(\bfF|\bfX,\bftheta )  \nonumber \\ 
&= \int_\bfF \prod_{\jd=1}^{\nD} \prod_{\jn=1}^{\nN} p(y\nd | f\nd) \prod_{\jd=1}^{\nD} p(\bff\d|\bfX,\bftheta) \nonumber \\ 
&=\prod_{\jd=1}^{\nD} \gaussianDist{\bfy\d}{\bfzero}{\bfK_{ff}+\beta^{-1}\eye} \label{eq:gplvm_factorized} .
\end{align}

The above equation is exactly the same as for the GP regression case, but now the optimization procedure also includes the latent variables $\bfX$. By choosing the dimensionality of $\bfX$ to be much smaller than that of the observed data, the \cal{GP} provides sufficient regularization such that both the parameters $\bftheta$ and the latent locations $\bfX$ can be found through maximum likelihood. Due to the flexible nature of the GP-LVM a large range of extensions have been suggested. In particular, by adding priors and seeking a MAP solution to the latent variable $\bfX$ different structures of latent representations can be found such that are topologically constrained \citep{Urtasun:uw}, consistent with a dynamic assumption \citep{Wang:2008ia} or constrained by class information \citep{Urtasun:2007vx}, to name just a few.

\subsection{Shared Gaussian Process Latent Variable Models}
The focus of this paper is to model $\numViews$ different views $\bfY^{(k)}$ within the same model. \citet{Shon:2006wr} proposed a GP-LVM where two sets of observations, $\bfY^{(1)}$ and $\bfY^{(2)}$, are assumed to be generated from the same latent variable $\bfX$ by two sets of independent \cal{GP} mappings, each with a separate covariance function. 
To generalize this beyond two views we introduce further notation for the collections of instantiated random variables for all views, \ie we denote the observed views as $\calY = \{ \bfY^{(k)} \}_{k=1}^K$ and their associated noiseless versions as $\calF = \{ \bfF^{(k)} \}_{k=1}^K$. The latter depend on separate \cal{GP} mappings which are parameterized respectively by $\bfTheta = \{ \bftheta^{(k)} \}_{k=1}^K$. By denoting the $\jd$th dimension of all data points in the $k$th observation space as $\bfy\d^{(k)}$, the marginal distribution of equation \eqref{eq:gplvm_factorized} is now generalized to,
\begin{align}
p(\calY | \bfX, \bfTheta, \{ \beta^{(k)}\}_{k=1}^K) 
   &= \int_{\calF} p(\calY|\calF,\{ \beta^{(k)}\}_{k=1}^K) p(\calF|\bfX, \bfTheta) \label{eq:sgplvm} \\
   &= \int_{\calF} \prod_{k=1}^K p(\mY^{(k)}|\mF^{(k)},\beta^{(k)}) p(\mF^{(k)}|\bfX, \bftheta^{(k)}) \nonumber \\
   &= \prod_{k=1}^K \prod_{\jd=1}^\nD p(\bfy\d^{(k)}|\bfX, \bftheta^{(k)},\beta^{(k)}), \nonumber
\end{align}
where we exceptionally included all parameters in the expressions to clearly show the dependencies.

From the above explained generative model, referred to as shared GP-LVM, different variants emerge depending on the regularization and factorization imposed for the latent space. These variants were reviewed in Section \ref{sec:relatedwork} and summarized in Figure \ref{fig:sgplvms}. Having just described the shared GP-LVM in detail, we will now explain the issues that have caused major practical problems in past shared GP-LVM approaches. Specifically, the straightforward adoption of the GP-LVM to the multi-view scenario by the shared GP-LVM implies certain non-obvious assumptions. Being a generative model, the GP-LVM objective reflects how well \emph{all} the variations in all views are summarized (modelled) in the latent space. This means that if a variation is confined to only one view this might have a detrimental effect on the generation of the other views, as the latent space will be reluctant to encode this. The approach of \citet{Ek:2008up} has proven to be the more robust in terms of avoiding the negative effect of having private variations polluting the model. Recall that this model introduced a factorized latent space where the shared and private subspaces are completely segmented: a single shared space $\bfX^{(s)}$ represents the variations that co-existed in all $K$ views and one private space $\bfX^{(k)}$ associated with each view represents the variations unique to that specific view. The learning of the factorized latent space model is performed in the same manner as for the standard GP-LVM model, \ie by maximum likelihood through gradient based methods. The non-convexity of the objective function requires a good initialization of the latent locations for such approach to reach a good solution. For a model with one single latent space the suggested approach in \cite{Lawrence:2005vk} was to initialize the latent space with the solution to an analogous spectral nonlinear algorithm such as isomap \citep{Tenenbaum:2000jp}, locally linear embeddings \citep{Roweis:2000ey} or maximum variance unfolding \citep{Weinberger:2004hg}. Alternatively, a linear generative approach such as probabilistic principal component analysis \citep{Tipping:1999uo} can be used. However, for the aforementioned factorized model no such analogous approach exists and, instead, an algorithm or heuristic has to be developed for initialization. \citep{Ek:2008up} used CCA to initialize the shared representation and a constrained version of PCA, referred to as non-consolidating component analysis (NCCA), to initialize the private space. However, as stated in the introduction, the objective of CCA is to maximize correlation, which is quite different objective from that of a shared GP-LVM which aims to reconstruct the data. Another approach is to alter the model by adding regularizers \citep{Salzmann:2010vh}. However, for that method to avoid local minima a rather involved annealing scheme was required which made the model difficult to train. 

The fundamental problem of all past shared GP-LVM approaches was that the dimensions of the latent space were divided into either shared or private; this hard separation rendered challenging the ``transfer'' of variations (information) between the different spaces during optimization. Furthermore, due to intractability issues the previous approaches were unable to communicate uncertainty throughout all nodes of the model, something which resulted in poorly regularized models.

In this paper we aim at expressing the shared GP-LVM as a IBFA model and, importantly, embedding it in our derived fully Bayesian framework. Our model then enjoys the benefits of the popular IBFA formulation while using powerful probabilistic and nonlinear mappings with simultaneous regularization.
In particular, the Bayesian formalism solves many of the regularization problems mentioned in the previous paragraphs in two ways: firstly, it allows for a relaxation of the structural factorization of the model from a hard and discrete representation, where each latent variable is exclusively associated with either a private or a shared space, to a smooth and continuous one. Secondly, the Bayesian formulation that we will present allows for robust model training without having to rely on ad-hoc regularizers; instead, a principled variational learning scheme is used for automatic and simultaneous estimation of both, the dimensionality and the structure of the latent representation. Notice that this Bayesian framework also provides better handling of the uncertainty through an approximation to the full posterior of the latent points given the data, something which is desirable for preventing overfitting or further extending the model in more complicated scenarios, for example deep architectures \citep{Damianou:deepgp13,damianou:thesis15}. Further, as we will show, it allows for including priors on the latent space such as dynamic models \citep{Damianou:2011vb}.

\section{Bayesian Nonparametric Nonlinear IBFA}\label{sec:mrd}
\begin{figure}[t]
  \begin{minipage}{0.49\textwidth}
    \begin{center}
      \scalebox{0.45}{

\tdplotsetmaincoords{90}{90}
\begin{tikzpicture}[node distance=3.5cm,tdplot_main_coords]
  \node(x)[vertexrv] {\Huge $\mathbf{X}$};

  \node(f1)[vertexrv,below left of =x] {\Huge $\bff^{(1)}_{\jn,:}$};
  \node(f2)[vertexrv,below right of =x] {\Huge $\bff^{(2)}_{\jn,:}$};

  \node(y1)[vertexobs,below of =f1] {\Huge $\mathbf{y}^{(1)}_{\jn,:}$};
  \node(y2)[vertexobs,below of =f2, label={[name=f2l]south east:\Huge $\nN$}] {\Huge ${\mathbf{y}^{(2)}_{\jn,:}}$};

  \node(t1)[vertexrv,left of =f1] {\Huge ${\bftheta^{(1)}}$};
  \node(t2)[vertexrv,right of =f2] {\Huge ${\bftheta^{(2)}}$};

  \node(s1)[vertexrv,left of =y1] {\Huge ${\beta^{(1)}}$};
  \node(s2)[vertexrv,right of =y2] {\Huge ${\beta^{(2)}}$};

  \node(w1)[vertexrv,above left of = f1] {\Huge $\mathbf{w}^{(1)}$};
  \node(w2)[vertexrv,above right of = f2] {\Huge $\mathbf{w}^{(2)}$};

  \node(tx)[vertexrv, above of =x]{\Huge $\bftheta_x$};

  \draw[edge](x) to (f1);
  \draw[edge](t1) to (f1);
  \draw[edge](f1) to (y1);
  \draw[edge](s1) to (y1);
  \draw[edge](tx) to (x);

  \draw[edge](x) to (f2);
  \draw[edge](t2) to (f2);
  \draw[edge](f2) to (y2);
  \draw[edge](s2) to (y2);

  \draw[edge](w1) to (f1);
  \draw[edge](w2) to (f2);

  \node(b1)[block][fit = (f1)(y1)(f2)(y2)(f2l)]{};

\end{tikzpicture}

      }
    \end{center}
  \end{minipage}
  \begin{minipage}{0.49\textwidth}
    \begin{center}
      \scalebox{0.45}{

\begin{tikzpicture}[node distance=3.5cm]
  \begin{scope}[xshift=5cm]
      \node(x0)[vertexrv] at (0,0){\huge $\mathbf{X}$};
      \node(t)[vertexrv,above of=x0]{\huge $\mathbf{\bftheta}_x$};

      \foreach \x in {1,2,3}
               {
                 \begin{scope}[rotate=120-(\x-1)*60,shift={(0,-3)}]
                   \node(y\x)[vertexobs] at (0,0) {\huge $\mathbf{Y}^{(\x)}$};
                   \node(t\x)[vertexrv] at (1.5,-3) {\huge $\mathbf{\bftheta}^{(\x)}$};
                   \node(w\x)[vertexrv] at (-1.5,-3) {\huge $\mathbf{w}^{(\x)}$};
                 \end{scope}

                 \draw[edge](x0)--(y\x);
                 \draw[edge](t\x)--(y\x);
                 \draw[edge](w\x)--(y\x);
               }
               \begin{scope}[rotate=-60,shift={(0,-3)}]
                 \node(yi)[vertexobs] at (0,0) {\huge $\mathbf{Y}^{(k)}$};
                 \node(ti)[vertexrv] at (1.5,-3) {\huge $\mathbf{\bftheta}^{(k)}$};
                 \node(wi)[vertexrv] at (-1.5,-3) {\huge $\mathbf{w}^{(k)}$};

                 \draw[edge](x0)--(yi);
                 \draw[edge](ti)--(yi);
                 \draw[edge](wi)--(yi);

               \end{scope}
               \begin{scope}[rotate=-120,shift={(0,-3)}]
                 \node(yN)[vertexobs] at (0,0) {\huge $\mathbf{Y}^{(\numViews)}$};
                 \node(tN)[vertexrv] at (1.5,-3) {\huge $\mathbf{\bftheta}^{(\numViews)}$};
                 \node(wN)[vertexrv] at (-1.5,-3) {\huge $\mathbf{w}^{(\numViews)}$};
               \end{scope}

               \draw[edge](x0)--(yN);
               \draw[edge](tN)--(yN);
               \draw[edge](wN)--(yN);

      \draw[edge](t)--(x0);


  \end{scope}
\end{tikzpicture}

      }
    \end{center}
  \end{minipage}
  \caption{
      Manifold relevance determination (MRD).
      Left: the graphical model for two views, depicting the dependencies of all variables before integrating out the latent mapping. 
       Each observation space $\bfY^{(k)}\in\mathbb{R}^{\nD_k}$ depends on a common latent space $\bfX\in\mathbb{R}^{\nQ}$. A weight vector $\bfw^{(k)}~\in~\mathbb{R}^{\nQ}$  associates the relative relevance of each dimension in the latent space for generating the view. $\bftheta^{(k)}$ denotes a set of additional kernel parameters that control the prior over the generative mapping. Right: extension to $\numViews$ views, shown after marginalizing the latent mappings and dropping dependence on the noise precision $\beta^{(k)}$. Despite the apparent similarity with the graphical model of \fig \ref{fig:sgplvms}(b) which depicts a fully shared latent space, here the role of $\mX$ is completely different; $\mX$ is marginalized out and, together with the additional weight parameters, operates in a Bayesian factorized model.
         }
  \label{fig:mrd}
\end{figure}
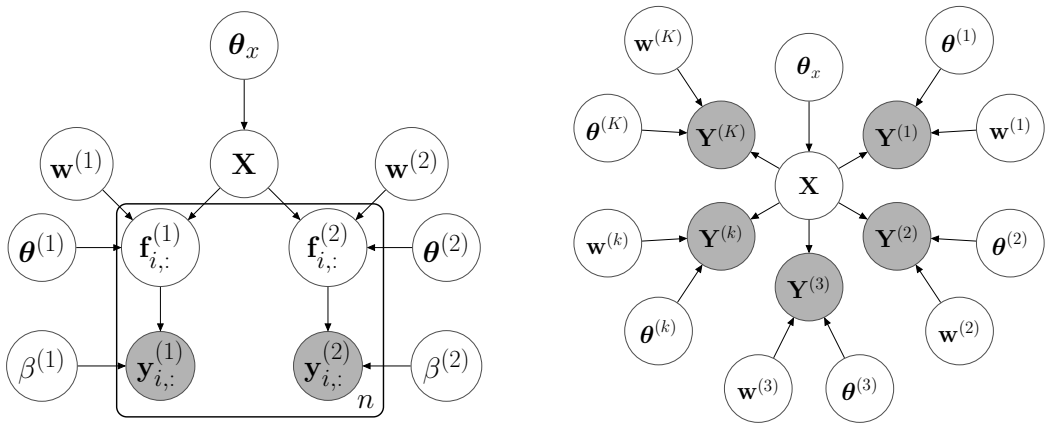

A central motivating idea behind the model we propose is to relax the ``hard'' (discrete) factorization into shared and private latent subspaces and instead use learn a ``soft'' (continuous) factorization. This is achieved by allowing the different views to choose the importance, if any, of each latent dimension from a \emph{single} latent space, while employing a continuous scale to measure the level of importance. This selection is done jointly with learning a posterior distribution over the latent space and as we will show this will let the shared and private dimensions to naturally emerge during optimization. We refer to this idea as \emph{manifold relevance determination (MRD)}.

A subtle but important difference wth past models is that, in MRD, there is only a single latent variable $\bfX$, from where we can still extract a private latent space, \eg $\bfX^{(k)}$, or a shared space, \eg $\bfX^{(k,l)}$, by chosing different sets of dimensions from $\bfX$. Contrast this with shared GP-LVM approaches which explicitly used different random variables to represent the private and shared space, thereby assuming a static structure a priori. Using MRD, the model is allowed to (and indeed often does) completely allocate a latent dimension to private or shared spaces, but may also choose to endow a shared latent dimension with more or less relevance for a particular view. Importantly, this factorization is learned from data by maximizing a variational lower bound on the model evidence, rather than through construction of bespoken regularizers to achieve a similar effect. MRD can be seen as a natural generalization of the traditional approach to manifold learning; we still assume the existence of a low-dimensional representation encoding the underlying phenomenon, but the variance contained in an observation space does not necessarily need to be governed by the full manifold or by a submanifold that has been selected in an ad-hoc way.

\subsection{Latent Factorization as Feature Selection}
MRD circumvents the challenges associated with the shared GP-LVM by relaxing the discrete factorization of the latent space to obtain a continuous one. To achieve this, we rephrase the factorized latent variable modeling in terms of a feature selection problem. Two ingredients are necessary for solving this feature selection problem: firstly, a means of incorporating continuous scale parameters factorizing the latent space and, secondly, a Bayesian training framework by which the scale parameters are determined, exploiting the Bayesian shrinkage principle. We elaborate on these two ingredients below.

To start with, we adopt the automatic relevance determination (ARD) idea \citep{Neal:1996ex} which is a popular way of performing automatic feature selection \citep{Neal:1996ex} in Bayesian settings. In the Gaussian process framework, ARD can be implemented by introducing a weight parameter $w_\jq$ which is associated by way of multiplication with the corresponding dimension $\xq$ of the input space. The weight $w_\jq$ is then learned such that it scales $\xq$ according to the importance of the $\jq-$th dimension for predicting the output. The introduction of the ARD vector $\bfw$ to the exponentiated quadratic (EQ) covariance function of equation \eqref{eq:EQ}, leads to the expression of the EQ-ARD covariance function:
\begin{equation}
\label{eq:ARD}
k_\mathsmaller{\text{EQ}}(\vx\inn{i},\vx\inn{r}) = \sigma^2_\mathsmaller{\text{EQ}} \exp \left( -\frac{1}{2} \sum_{\jq=1}^\nQ w_\jq (\sx\inq{i}{\jq} - \sx\inq{r}{\jq})^2 \right).
\end{equation}
In this equation the weight $w_\jq$ encodes the relative relevance of dimension $q$ in determining the co-variance between the $i$th and the $r$th data point. A similar effect is achieved by introducing ARD to some other covariance function, such as the linear one. From the perspective of a shared GP-LVM, the above observation motivates the idea of using a \emph{single} latent space while considering an independent ARD covariance function for each view. Using a different weight set $\bfw^{(k)}$ for each view $k$ will allow the model to automatically infer the responsibility of each latent dimension for generating points in each view, as can be seen in Figure~\ref{fig:mrd}. We collectively denote the set of additional kernel hyperparameters as $\bfW=\{\bfw^{(k)}\}_{k=1}^K$, and refer to them as ``weights'' (rather than squared inverse length-scales) to highlight their distinct role in our model.

While incorporating the ARD idea in the shared GP-LVM setting might seem straight-forward, it is important to note that the addition of $\bfW$ alone cannot have the same shrinkage/feature selection effect as observed in standard GPs. This is because, in the shared GP-LVM case, the inputs to the covariance function are latent, and treating both $\bfW$ and $\bfX$ as parameters can lead to severe overfitting, as has been demonstrated by \citet{Damianou:variational15}. Therefore, in contrast to the traditional shared GP-LVM approaches, in MRD we aim for a fully Bayesian training framework where the latent space is treated as a distribution, so that the effects of Bayesian shrinkage can be realized \citep[see e.g.][]{Tipping:relevance00,Bishop:bayesPCA98}. Unfortunately, this task is intractable in our case because our nonlinear IBFA approach needs to employ nonlinear covariance functions, through which distributions on the latent space cannot be propagated analytically. Standard variational approaches also fail to cope with this intractability. In this work, we solve this issue by adopting recent work on variational propagation of uncertainty in Gaussian processes \citep{Titsias:2010tb,Damianou:variational15}. The resulting framework is explained in the next section.

\subsection{Bayesian Training}
The traditional maximum likelihood training of \cal{GP}s is challenging, due to the existence of multiple local minima. As described in the previous section, learning the factorized model poses an even greater challenge and incorporating additional parameters by introducing ARD kernels is only going to make training harder. To solve the latent factorization learning and ARD parameter determination at the same time we wish to use the data evidence as an optimization objective within a Bayesian framework. Specifically, the evidence is obtained after placing a prior distribution on the latent space, $\bfX$, and marginalizing it out. Then, equation \eqref{eq:sgplvm} is transformed into:
\begin{equation}
    \label{eq:evidence}
  p(\calY | \bfW,\bfTheta, \bftheta_x) =
  \int_\calF \left( p(\calY | \calF)
  \int_{\bfX} p(\calF | \bfX, \bfW, \bfTheta)
  p(\bfX | \bftheta_x) \right), 
\end{equation}
where the expressions with calligraphic notation for all views are expanded into a product as in equation \eqref{eq:sgplvm}. 
A simple choice for the latent space prior is a standard normal, $p(\bfX) \sim \mathcal{N}(\bfzero, \bfI)$, but in Section \ref{sec:dynamics} we will investigate more structured choices. We are also allowed to define further priors on the parameters $\bfW$, $\bfTheta$ and $\bftheta_x$. For the remainder of this paper we will, for clarity, drop the dependency on these parameters from our expressions. The Bayesian training procedure allows for an automatic Occam's razor during which the dimensions of each weight vector $\bfw^{(k)}$ are switched off if needed; this would be impossible if $\bfX$ was not marginalized, since the likelihood would typically increase by allowing a larger latent space.
However, the above integral is intractable, since the factors in $p(\calF|\bfX)=\prod_{k=1}^K p(\bfF^{(k)}|\bfX)$ contain $\bfX$ nonlinearly. Even standard variational approaches fail. Such methods attempt to find a variational distribution 
$\calQ\dep{\calF,\bfX}$
which best approximates the true posterior of the integrands through minimizing the KL divergence,
\begin{align}
 & \KL{\calQ\dep{\calF, \bfX}}{p(\calF, \bfX | \calY)} =
  		\log p(\calY) - \calL\dep{\calQ(\calF, \bfX)}, \label{eq:KLfx} \\
& \text{where} \; \; \; \calL\dep{\calQ(\calF, \bfX)} =
	\int_{\calF, \bfX} \calQ\dep{\calF, \bfX}
	\log \frac{p(\calY|\calF)p(\calF|\bfX)p(\bfX)}{\calQ\dep{\calF, \bfX}} \label{eq:Lfx} .
\end{align}
By rearranging the terms in equation \eqref{eq:KLfx} we see that minimizing the KL-divergence between the true and the approximate posterior is equivalent to maximizing the lower bound $\calL\dep{\calQ(\calF, \bfX)}$ on the model evidence.
Indeed, since the KL term is non-negative, we can drop it to form the inequality
\begin{equation}
\label{eq:boundInequality}
\log p(\calY) \ge \calL\dep{\calQ(\calF, \bfX)}
\end{equation}
revealing an alternative optimization objective: maximizing the functional $\calL$ with respect to the variational distribution $\calQ$ and the (hyper)parameters is equivalent to maximizing the model evidence with respect to the model (hyper)parameters\footnote{Recall that the model (hyper)parameters appear in the model evidence of equation \eqref{eq:evidence} and similarly in equation \eqref{eq:Lfx} for $\calL$, but were dropped for clarity from the expressions.}. Having made this observation, we will henceforth further simplify our notation by writing $\calL(\calF,\bfX)$ instead of $\calL(\calQ(\calF,\bfX))$.

 However, the above standard variational approach is problematic, since it still requires integration of $\bfX$ through the challenging term $p(\calF | \bfX)$ which still appears when we expand the numerator of $\calL\dep{\calF, \bfX}$. To circumvent this problem we follow \cite{Titsias:2009vf,Titsias:2010tb} and apply the ``data augmentation'' principle, where we augment the probability space with auxiliary pseudo-inputs $\calZ=\{\bfZ^{(k)}\}_{k=1}^K$ corresponding to output values $\calU=\{\bfU^{(k)}\}_{k=1}^K$ drawn from the same priors as $\calF$, that is, $p(\bfU^{(k)} | \bfZ^{(k)})$ is a distribution with the same form as its corresponding $p(\bfF^{(k)} | \bfX)$. Here we consider the general case where a separate set of inducing inputs is used per modality, but another approach would be to use the same $\bfZ$ for all $k$. 
 The probability of the GP prior shown in equation \eqref{eq:gpPrior} now takes the following augmented expression for each view and dimension:
 
 \newcommand{\kv}{^{\mathsmaller{(k)}}}

 \begin{align*}
p(\vf\d\kv, \vu\kv\d | \mX, \bfZ\kv) 
	&= p(\vf\kv\d | \vu\kv\d , \mX, \bfZ\kv ) p(\vu\kv\d | \bfZ\kv ) \\
	&= \gaussianDist{\vf\kv\d}{\Kfu\kv(\Kuu\kv)^{-1}\vu\kv\d}{\Kff\kv-\Kfu\kv(\Kuu\kv)^{-1}\Kuf\kv}
	   \gaussianDist{\vu\kv\d}{\bfzero}{\Kuu\kv},  
 \end{align*}

\noindent where $\Kuu\kv = k\kv(\bfZ\kv, \bfZ\kv)$ denotes the covariance matrix obtained by evaluating the covariance function of view $k$ on the auxiliary inputs, $\Kuf\kv=k\kv(\bfZ\kv, \mX)$ and $\Kfu\kv = k\kv(\mX,\bfZ\kv)$. As can be seen, the auxiliary variables are used to \emph{compress} the latent function signal into a representation which relies on a low-rank covariance matrix \citep{Quinonero:unifying05}. 

After expanding the probability space with auxiliary variables, equations \eqref{eq:KLfx} and \eqref{eq:Lfx} now become
\begin{align}
& \KL{ \calQ\dep{\calF, \bfX, \calU}}{p(\calF, \bfX, \calU | \calY, \calZ)}  =  
\log p(\calY) - \calL\dep{\calF, \mX, \calU, \calZ} \label{eq:posteriorApprox} \\
& \text{where} \; \;  
\calL\dep{\calF, \mX, \calU, \calZ} = \int_{\calF, \bfX, \calU} \calQ\dep{\calF, \bfX, \calU}
	\log \frac{p(\calY|\calF) p(\calF|\calU, \bfX, \calZ) p(\calU|\calZ) p(\bfX)}{\calQ\dep{\calF, \bfX, \calU}},  \label{eq:klend}
\end{align}
where the numerator inside the logarithm shows the augmented joint distribution. The above integral is still intractable, since $p(\calF|\calU, \bfX, \calZ)$ again contains $\mX$ nonlinearly. However, we are now able to remove this term from the log. after first compressing it in a way that its contribution is manifested through the set of auxiliary variables. To achieve this in a principled way, we follow a special mean-field methodology by defining a variational distribution which factorizes as,
\begin{align}
\calQ\dep{\calF, \bfX, \calU} 
  &= p(\calF | \calU, \bfX, \calZ) q(\calU) q(\bfX) \label{eq:varDistr} \\
  &= q(\bfX) \prod_{k} \underbrace{\prod_{\jd=1}^{\nD_k} p(\fd^{(k)} | \ud^{(k)}, \bfX, \bfZ^{(k)}) q(\ud^{(k)})}_{\calQ^{(k)}} .\nonumber 
\end{align}
 While the forms of each individual factor will be defined later on, for the moment we notice that the above factorization causes the collection of challenging terms $p(\calF|\calU,\bfX, \calZ)$ to cancel out inside the logarithm of $\calL\dep{\calF, \mX, \calU, \calZ}$ in equation \eqref{eq:klend}, leaving us with a tractable expression for the bound. By keeping a variational distribution for $q(\calU)$, the framework encourages compression of the latent function signal into the auxiliary variables \citep{Titsias:2009vf,Hensman:nested14}. After replacing equation \eqref{eq:varDistr} into the bound \eqref{eq:klend} and cancelling the aforementioned terms inside the log, we expand the resulting expression by separating the integrals, so that the variational bound becomes:
\begin{align}
 \calL\dep{\calF, \mX, \calU, \calZ} & = \mathbb{E}_{\calQ}[\log p(\calY|\calF)] - \KL{q(\calU)}{p(\calU | \calZ)} - \KL{q(\bfX)}{p(\bfX)},
\label{eq:boundEnd}
\end{align}
where $\mathbb{E}_{\mathcal{Q}}[p(\cdot)]$ is the expectation of $p(\cdot)$ under $\calQ=\calQ\dep{\calF, \bfX, \calU}$. As per equation \eqref{eq:boundInequality}, the expression \eqref{eq:boundEnd} constitutes our final objective to be maximized. The obtained form also reveals the availability of the posterior marginal $q(\bfX)$, constituting an approximation to the true latent space posterior $p(\bfX | \calY)$. This approximate posterior appears in the term $-\KL{q(\mX)}{p(\mX)}$ of our variational objective. This term acts as a Bayesian regularizer, penalizing unnecessarily complex posteriors. 

We have so far left unspecified the forms of the variational distributions appearing in $\calQ$, in equation \eqref{eq:varDistr}. We choose these forms so that the variational objective \eqref{eq:boundEnd} remains tractable. Specifically, we choose $q(\bfX)$ to be a factorized distribution:
\begin{equation}
\label{eq:qX}
q(\bfX) = \prod_{\jn=1}^\nN\mathcal{N}(\xn| \bfmu\n,\mathbf{S}\n),
\end{equation}
where the parameters of each Gaussian are to be learned through the variational optimization framework. Notice that the dimensions of each $\bfx\n$ are uncorrelated, so that $\mathbf{S}\n$ is a diagonal matrix.
As for $q(\calU) = q(\{ \bfU^{(k)} \}_{k=1}^\numViews)$, one option which maintains tractability of equation \eqref{eq:boundEnd} is to allow each $q(\bfU^{(k)})$ to be a Gaussian distribution the parameters of which need to be learned. Another option is to follow \citep{Titsias:2010tb} and replace each $q(\bfU^{(k)})$ with its optimal form, found by differentiating the objective \eqref{eq:boundEnd} with respect to $q(\bfU^{(k)})$, setting the expression to zero and solving for $q(\bfU^{(k)})$. The latter approach is taken in our work.
Specifically, the optimal distribution for each $q(\bfU^{(k)})$ depends on all the terms which interact with $q(\bfU^{(k)})$ in the objective, that is, $\mathbb{E}_{\calQ^{(k)}}[p(\mY^{(k)} | \mF^{(k)})]$, $p(\mU^{(k)} | \bfZ^{(k)})$ and $q(\mX)$. 
Further details for this derivation are given in Appendix \ref{appendix:bound}. The result of the above procedure, referred to as optimally ``collapsing'' the variational distribution $q(\calU)$, is a variational lower bound which does no longer depend on $\calU$ and is tighter than the previous bound:
\begin{equation}
\label{eq:boundCollapsed}
\calL\dep{\calF, \mX, \calZ} = - \KL{q(\bfX)}{p(\bfX)} + \sum_{k} \calL^{(k)} \ge \calL\dep{\calF, \calU,\mX, \calZ},
\end{equation}

\noindent where the term $\mathcal{L}^{(k)}$ is the collapsed version of the terms 
$$
\mathbb{E}_{\calQ}[\log p(\mY^{(k)}|\bfF^{(k)})] - \KL{q(\bfU^{(k)})}{p(\bfU^{(k)} | \bfZ^{(k)})}
$$
appearing when we expand equation \eqref{eq:boundEnd}.

To summarize, the optimization procedure uses equation \eqref{eq:boundCollapsed} as an objective and it uses a gradient based method to jointly optimize the following parameters:
\begin{itemize}[noitemsep]
\item \emph{Variational parameters}: the $2 (\nN \times \nQ)$ parameters $\{ \bfmu\n, \bfS\n \}_{\jn=1}^\nN$ of $q(\bfX)$ 
; the $\nM \times \nD_k$ matrix of inducing inputs $\bfZ^{(k)}$ for each view $k$.
\item \emph{Model parameters}: the Gaussian noise variances $\{ \beta^{(k)}\}_{k=1}^K$
\item \emph{Hyperparameters}: the kernel parameters $\bfTheta$ and the kernel relevance weights $\bfW$.
\end{itemize}

\subsection{Factor Constraints Through Priors\label{sec:dynamics}}

In the previous section we showed that the MRD framework can approximately integrate out the latent space $\bfX$ and maximize the logarithm of the evidence $p(\calY)$. In addition to the previously discussed benefits of MRD as an inference engine for nonlinear, nonparametric IBFA, notice that MRD also allows principled incorporation of additional priors over the latent space. Such priors express our preference for specific properties of IBFA's factors and provide disambiguation at test time (when different factors compete to explain test data). The ability to incorporate priors in a principled way stems from the Bayesian nature of MRD, as can be seen in equation \eqref{eq:evidence}. As an important example of this, we will now describe how a latent prior that respects the temporal dynamics of the data can be incorporated into the MRD model. 

Many types of data have an inherent dynamical structure. When learning a latent representation of such data there are several benefits to encourage the representation to respect this dynamic. An example is when we wish to synthesize novel data by either inter- or extrapolating a sequence. For the standard GP-LVM model, an autoregressive (Markovian) prior was suggested by \cite{Wang:2008ia}, while \cite{Damianou:2011vb} proposed a regressive (temporal) prior for the fully Bayesian model. The autoregressive structure proposed by \citeauthor{Wang:2008ia} was introduced to the shared GP-LVM by \cite{Ek:2008up,Ek:2007uo}, where the dynamics were exploited to disambiguate sequences of human motion in order to perform human pose estimation. In the experimental section of the paper we will reproduce the experiments performed by \citeauthor{Ek:2007uo}, this time using the MRD approach and the regressive dynamics framework.

Here we follow \citet{Damianou:2011vb} and use a temporal, Gaussian process prior to model the dynamics. Using a nonparametric prior constitutes a flexible solution, in line with the Bayesian formulation of MRD. Given the work of \citet{Damianou:2011vb}, it is straightforward to include such prior in the MRD framework. Nevertheless, we will re-iterate the corresponding derivations here as this is instructive about how other kinds of latent space priors can be developed.

We start by reformulating the latent space as $\nQ$ independent latent \emph{functions} $\{ \bfx\q \}_{\jq=1}^\nQ$. To encode the sequential structure  of each $\nN-$dimensional latent function, we introduce correlation through the $\nN-$dimensional vector of time-stamps $\bft = [t_1, \ldots ,t_\nN]^\top, t_\jn \in \Re$, which we assume are given together with our outputs $\calY$. Each element $t_\jn$ represents the time at which the $\jn-$th collection of corresponding view-instances $\{ \bfy\n^{(k)}\}_{k=1}^K$ was observed. This kind of latent coupling reflects the temporal correlation between instances of each view $\bfY^{(k)}$, and we assume that the same dynamics are respected by all views. Then, we have:

\begin{align}
  x\q(t)\sim\mathcal{GP}(0,k_x(t',t'')),
\end{align}
where $k_x$ is the temporal covariance function.
The joint probability density of the model which is augmented with inducing points and time-stamps takes the form,
\begin{align}
  p(\calY,\calF,\calU,\bfX|\calZ,\mathbf{t}) = p(\calY|\calF)p(\calF|\calU,\bfX)p(\calU|\calZ )p(\bfX|\mathbf{t}),
\end{align}
where $p(\bfX|\bft)$ is a product of $\nQ$ independent Gaussian distributions (due to the GP priors employed for the latent space): 
\begin{equation*}
p(\bfX|\bft)=\prod_{\jq=1}^\nQ \gaussianDist{\bfx\q}{\bfzero}{\bfK_t}, \; \; \; \text{where:} \; \ \; \bfK_t = k_x(\bft, \bft),
\end{equation*}
 \ie $\bfK_t$ is a covariance matrix constructed through $k_x$ with training time-stamps as inputs. 
The derivations for the variational framework of this temporal model follow as before. Specifically, from equation \eqref{eq:boundEnd} we see that $p(\bfX)$ only appears in the last KL term; the rest of the terms do not get affected by the dynamics directly, but only indirectly through $q(\bfX)$. Therefore, the variational lower bound for the dynamical MRD is:
\begin{align}
 \calL\dep{\calF, \mX, \calU, \calZ} & = \mathbb{E}_{\calQ}[\log p(\calY|\calF)] - \KL{q(\calU)}{p(\calU | \calZ)} - \KL{q(\bfX)}{p(\bfX|\bft)}.
\label{eq:boundEndKL}
\end{align}
Although the above form is very similar to the non-dynamical variational bound of equation \eqref{eq:boundEnd}, here the framework employs a point-wise latent space coupling which is reflected in the approximate posterior by choosing a coupled form for $q(\mX)$. To maintain tractability we choose $q(\bfX)$ to be a Gaussian distribution, as before. However, in contrast to equation \eqref{eq:qX}, we now factorize the variational distribution according to dimensions, following \citet{Damianou:2011vb}. In this way, the latent points remain a posteriori coupled through the dynamics, that is 
$$
q(\bfX) = \prod_{\jq=1}^\nQ\mathcal{N}(\xq | \bfmu\q,\mathbf{S}\q),
$$
\noindent where $\bfS\q$ is a full, $\nN \times \nN$ covariance matrix (in contrast to the diagonal matrix assumed in equation \eqref{eq:qX}). Notice that we now have $\nQ$, $\nN \times \nN$ parameters to learn for all $\{\bfS\q\}_{\jq=1}^\nQ$, however reparameterization tricks to reduce this number exist \citep{Opper:2009vf,Damianou:2011vb}.

With this dynamical approach, we are also allowed to learn the structure of multiple statistically independent sequences which, nevertheless, share the same dynamical structure (\eg multiple strands of different people walking). Each sequence is represented by a collection of multi-view instances and corresponding time-stamps. Sequence learning through MRD is achieved by learning a common latent space for all latent timeseries while, at the same time, ignoring correlations between latent points which correspond to outputs of different sequences. To add this functionality to our framework, we just need to define the temporal covariance matrix $\bfK_t$ to have a block-diagonal structure by setting $k_x(t,t') = 0$ if $t$ and $t'$ belong to different sequences. In the experimental section of the paper we will evaluate our model in this setting.

\subsection{Latent Space Post-hoc Analysis\label{sec:latentPostHoc}}
As will be discussed in the next section, the training and inference within MRD can be realized within the framework of ``soft'' latent space factorization. However, in certain cases we might need an explicit ``hard'' segmentation for posterior tasks. For example, consider the case where we wish to use the shared latent space discovered from very high-dimensional views as features for a subsequent classification task. In this case, we will need to firstly decide on which latent dimensions we consider to be shared and which private. Another example is the scenario where we simply need to perform high-level reasoning about correlations discovered between the different views. Therefore, we wish to split the common latent space $\bfX$ into subspaces $[\bfX^{\calA, \calB}, \bfX^\calA, \bfX^\calB, \bfX^\mathsmaller{{\backslash\{\calA,\calB\}}}]$, where $\bfX^{\calA,\calB}$ denotes the subspace shared for view-sets\footnote{Latent space segmentation for multiple view-sets follows trivially in the same way.} $\calA$ and $\calB$; $\bfX^{\calA}$ denotes the subspace which is private for $\calA$ (and analogously for $\bfX^{\calB}$); $\bfX^\mathsmaller{{\backslash\{\calA,\calB\}}}$ denotes a subspace which is irrelevant for both $\calA$ and $\calB$. To achieve this segmentation, we compare the relative value of the ARD weights for each view and per dimension of $\bfX$. To have a comparison criterion which is consistent across views, we can first normalize all weight sets $\bfw^{(k)} = [w_1^{(k)}, \cdots, w_\nQ^{(k)}], \forall k \in \{ \calA, \calB \}$ to be in the same range. Without loss of generality let us assume that for each $k$ we create a weight set $\tilde{\bfw}^{(k)}$ which is a normalized version of $\bfw^{(k)}$ such that the maximum element per weight-vector is $1$. Then, dimension $\jq$ is deemed as shared between views $r,l$ if $\tilde{w}\q^{(r)}$ and $\tilde{w}\q^{(l)}$ are larger than a small threshold value $\varepsilon$, since $\tilde{w}\q^{(r)}$ is related to the amount of variance in view $r$ that is explained by latent dimension $\jq$.  
We will explain this concept mathematically below, but first let us introduce the notation $\bfX_{:,\calJ}, \calJ \subseteq \{1,2, \cdots, \nQ\}$ to mean a collection of columns (dimensions) of $\bfX$, that is, a subspace of the common latent space. Then, the segmentation of the latent space is defined as follows:

  \begin{align}
  \begin{split}
  \label{eq:segmentation}
    \forall & \jq\in\{1,\ldots,\nQ\}, \ a\in\mathcal{A}, \ b\in\mathcal{B}: \\
    \mathbf{X}^{\calA,\calB} &= 
      \mathbf{X}_{:,\calJ}, \; \;  \text{for} \ \; \calJ = \{ \jq | \tilde{w}^{(a)}_\jq \geq \varepsilon, \ \tilde{w}^{(b)}_\jq \geq \varepsilon \}, \\
  \mathbf{X}^{\mathcal{A}} &= 
      \mathbf{X}_{:,\calJ}, \; \;  \text{for} \ \; \calJ = \{ \jq | \tilde{w}^{(a)}_\jq \geq \varepsilon, \ \tilde{w}^{(b)}_\jq < \varepsilon \},  \\
  \mathbf{X}^{\mathcal{B}} &= 
      \mathbf{X}_{:,\calJ},\; \;  \text{for} \ \; \calJ = \{ \jq | \tilde{w}^{(a)}_\jq < \varepsilon,    \ \tilde{w}^{(b)}_\jq \geq \varepsilon \},  \\
  \mathbf{X}^\mathsmaller{{\backslash\{\calA,\calB\}}} &= 
      \mathbf{X}_{:,\calJ},\; \;  \text{for} \ \; \calJ = \{ \jq |  \tilde{w}^{(a)}_\jq < \varepsilon,   \ \tilde{w}^{(b)}_\jq < \varepsilon \}.
   \end{split}   
\end{align}

It is worth emphasizing that, in practice, we found that during optimization the model is capable of performing the above truncation automatically. Specifically, the strong Bayesian regularization drives weights for unnecessary (per view) dimensions to values which are practically zero, considering machine precision limits, giving a clear separation with regards to relevant dimensions. Therefore, we can safely set the threshold $\varepsilon$ to a small number without the need to represent it as a parameter. In our experiments we use $\varepsilon = 10^{-3}$.

\subsection{Predictions}

A central motivation behind the IBFA model is that it provides a natural latent structure for efficient and intuitive inference, especially in scenarios where the estimation task is ambiguous. 
We will now proceed to describe how we can infer outputs in a subset of views, denoted by $\calA$, given information about outputs in a different subset of views, denoted by $\calB$. The MRD framework renders this task easy because, even if the views live in different, incomparable spaces, they are linked through a common latent space. View-specific weights automatically define latent subspaces by which subsets of views are related.

 Our task at test time is to generate a new (or infer a training) set of outputs $\mY^\mathcal{\viewZi}\Ts$ given a set of (potentially partially) observed test points $\mY^\mathcal{\viewYi}\Ts$. The inference proceeds by predicting (the distribution over) the set of latent points
$\mX\Ts \in \mathbb{R}^{\nN\Ts \times \nQ}$ which is most likely to have generated
$\mY^\mathcal{\viewYi}\Ts$. For this, we use the approximate posterior
$q(\mX,\mX\Ts)  \approx p(\mX,\mX\Ts|\mY^\mathcal{\viewYi},\mY^\mathcal{\viewYi}\Ts)$. This approximate posterior is found by optimizing a variational lower bound on the marginal likelihood $p(\mY^\mathcal{\viewYi},\mY^\mathcal{\viewYi}\Ts)$. This bound has analogous form with the training objective function of equation \eqref{eq:boundEnd}. That is, to find $q(\mX,\mX\Ts)$ we use equation \eqref{eq:boundEnd} but now the data is the augmented set $(\mY^\mathcal{\viewYi},\mY^\mathcal{\viewYi}\Ts)$ and the latent space is $(\mX,\mX\Ts)$. We assume that $q(\mX,\mX\Ts)=q(\mX)q(\mX\Ts)$ and since $q(\mX)$ is already obtained from the training phase, the expression for the augmented variational bound can be decomposed to statistics that have been computed at training time (and are held fixed during test time) and to statistics which are inferred during test time \citep{Titsias:2010tb,Damianou:variational15}, see Appendix \ref{appendix:latentPosterior} for details. It is also worth noting that the test posterior is fully factorized, \ie $q(\mX\Ts) = \prod_{\jn=1}^{\nN_*} q(\vx\nTs)$, so that the test computations can be made in parallel. After finding $q(\mX,\mX\Ts)$, we can then find a distribution of the outputs $\mY^\mathcal{\viewZi}\Ts$ by taking the expectation of the likelihood $p(\mY^\mathcal{\viewZi} | \mX)$ under the marginal $q(\mX\Ts)$:
\begin{align}
p(\mY^{\calB}\Ts) 
&\approx \int_{\mF^{\calB}\Ts, \mU, \mX\Ts} p(\mY^{\calB}\Ts|\mF^{\calB}\Ts) p(\mF^{\calB}\Ts|\mU,\mX\Ts) q(\mU) q(\mX\Ts)  \label{eq:predictive1} \\
&= \int_{\mX\Ts} p(\mY^{\calB}\Ts | \mX\Ts) q(\mX\Ts)  . \label{eq:predictive2} 
\end{align}

The above expectation takes the form of Gaussian process prediction with uncertain inputs and is intractable. However, the predictive distribution's moments can be computed analytically following the methodology which is outlined in detail by \cite{Girard:uncertain01,Titsias:2010tb,Damianou:variational15}. The resulting expressions are given in Appendix \ref{appendix:topdown}. Notice that to compute this expectation we used the posterior $q(\mX\Ts)$ optimized for the test data of the observed views, $\mY^{\calA}$, while we re-used the posterior $q(\mU)$ estimated during training time. This is because, in contrast to the latent input variables, the auxiliary variables are used as \emph{global} variables.

The above described way of finding the posterior $q(\mX\Ts)$ based on a subset of views $\mY^\mathcal{\viewYi}\Ts$ makes full use of the ``soft'' latent space factorization which is unique to MRD: the posterior $q(\mX\Ts) \approx p(\mX\Ts | \mY^\mathcal{\viewYi}\Ts)$ is found, based on which the missing instances for view-set $\calB$ are predicted without having to decide on a ``hard'' latent space segmentation. Indeed, the relevance weights $\bfw^\calA$ and $\bfw^\calB$ are involved in the above inference and weight all outcomes automatically and consistently. 

The inference procedure described above means that the latent dimensions that are (emerging as) private for $\mY^\mathcal{\viewZi}\Ts$ are taken directly from the prior, since the views in $\mY^\mathcal{\viewYi}$ can tell us nothing about the private information in $\mY^\mathcal{\viewZi}$. Although this works well in practice, another approach is to ``force'' the private latent space $\mX^\mathcal{\viewZi}$ to take values that are closer to those found in the training set. For example, when the modelled data is images and the predictive task is generation of new images, we might prefer to sacrifice predictive accuracy for obtaining sharper images. For this scenario, a simple heuristic is suggested: after optimizing $q(\mX\Ts)$ based on $\mY^\mathcal{\viewYi}\Ts$ as was described above, we perform a nearest neighbour search to find the training latent points $\tilde{\mX}$ which are closest to the mean of $q(\mX\Ts)$ in the projection to the dimensions that are shared between the views in $\mY^\mathcal{\viewYi}$ and in $\mY^\mathcal{\viewZi}$. For every test marginal $q(\vx_*)$, we then use the nearest training point $\tilde{\vx}$ to fill (replace) the dimensions of the mean for $q(\vx_*)$ which have been informed only through the prior, \ie the dimensions corresponding to $\mX^\calB$. This results in a hybrid test distribution $q(\tilde{\mX}_*)$ which is predicted from view $\calA$ through posterior inference but also takes training information from view $\calB$ through the heuristic.
 $\mY^\mathcal{\viewZi}\Ts$ can then be predicted as explained in the previous paragraph, \ie by computing the expectation of the likelihood $p(\mY^\mathcal{\viewZi} | \mX)$ under $q(\tilde{\mX}\Ts)$. This procedure, summarized in Algorithm \ref{algorithm:MRD_inference}, is used for our experiments. 

Notice that if $\mY^\mathcal{\viewYi}\Ts = \mY^\mathcal{\viewYi}$, then the aforementioned nearest neighbour search for finding $\tilde{\mX}$ in essence constitutes a means of finding correspondences between the two sets of views through a simpler latent space. This case is demonstrated in Section \ref{sec:MRD_experimentYaleFaces}.

\begin{algorithm*}[t]
\caption{Inference algorithm in MRD, assuming two sets of views $\viewY$ and $\viewZ$}\label{algorithm:MRD_inference}
\begin{algorithmic}[1]
\SSTATE \emph{Given}: MRD model trained on views $(\viewY,\viewZ)$ to obtain $q(\bfX) \sim \mathcal{N}(\bfM, \bfSigma)$.
\SSTATE Define the split $\bfM=(\bfM^{\viewYi},\bfM^{\mathcal{AB}},\bfM^{\viewZi})$ following Section \ref{sec:latentPostHoc}.
\SSTATE \emph{Given}: A test point $\viewyTs$.
\SSTATE \label{state:MRD_q} Optimize $q(\mX, \vx\Ts) \approx p(\mX, \vx\Ts|\viewY,\viewyTs)$.
\SSTATE Get the marginal $q(\vx\Ts)$ with mean $\bfmu\Ts = (\bfmu^{\viewYi}\Ts,\bfmu^{\viewYi\viewZi}\Ts,\bfmu^{\viewZi}\Ts)$ and variances $\bfs_*$.
\SSTATE \label{state:MRD_tilde_xk}Find $\mathcal{K}$ points $\tilde{\bfmu}_{(\kappa)}$ from a $\mathcal{K}-$nearest neigbour search between $\bfmu^{\viewYi\viewZi}\Ts$ and $\bfM^{\viewYi\viewZi}$.

\FOR{$\kappa = 1, \cdots,\mathcal{K}$}
    \STATE \label{state:MRD_tilde_xk_star} Generate $\tilde{\bfmu}_{(\kappa),*}$ by combining the dimensions of $\tilde{\bfmu}_{(\kappa)}$ and $\bfmu\Ts$ according to sets $\calA$ and $\calB$, \ie set $\tilde{\bfmu}_{(\kappa),*} = (\bfmu^{\viewYi}\Ts,\bfmu^{\viewYi\viewZi}\Ts,\tilde{\bfmu}^{\viewZi}_{(\kappa)})$. {\small {(Another heuristic is to explicitly consider the relevance weights in this step by ``blending'' the information coming from $\tilde{\bfmu}$ and $\bfmu_*$ according to $\vw^\calA$ and $\vw^\calB$ respectively. Then, $\tilde{\bfmu}^{\viewZi}_{(\kappa)}$ is replaced by a weighted average of $\bfmu^{\viewZi}\Ts$ and $\tilde{\bfmu}^{\viewZi}_{(\kappa)}$).}}
    \STATE Construct $q(\tilde{\vx}_{(\kappa),*}) \sim \mathcal{N}(\tilde{\bfmu}_{(\kappa),*},\bfs_*)$.
    \STATE Generate $\viewz_{(\kappa),*}$ from $\int_{\tilde{\vx}_{(\kappa),*}} p(\viewz_{(\kappa),*} | \tilde{\vx}_{(\kappa),*}) q(\tilde{\vx}_{(\kappa),*})$ using equation \eqref{eq:predictive2}.
\ENDFOR
\SSTATE \textbf{end for}
\SSTATE Most likely predictions are obtained for $\kappa=1$. We only use $\kappa>1$ if we are interested in finding correspondences in the training set (see Section \ref{sec:MRD_experimentYaleFaces}).
\SSTATE In the dynamical MRD version, all test points $\viewYTs$ are considered together, so that the variational distribution $q(\mX, \mX\Ts)$ of state \ref{state:MRD_q} will form a timeseries.
\end{algorithmic}
\end{algorithm*}

\section{Experiments}
\label{sec:experiments}
In this section we will show the experimental evaluation of the model proposed in this paper. We will apply the model to several different types of data, with different number of views and associated with different tasks.

\subsection{Toy Data}
\begin{figure}[t]
  \begin{center}
    \subfloat[Eigenspectrum of observed data]{\includegraphics[width=0.39\textwidth]{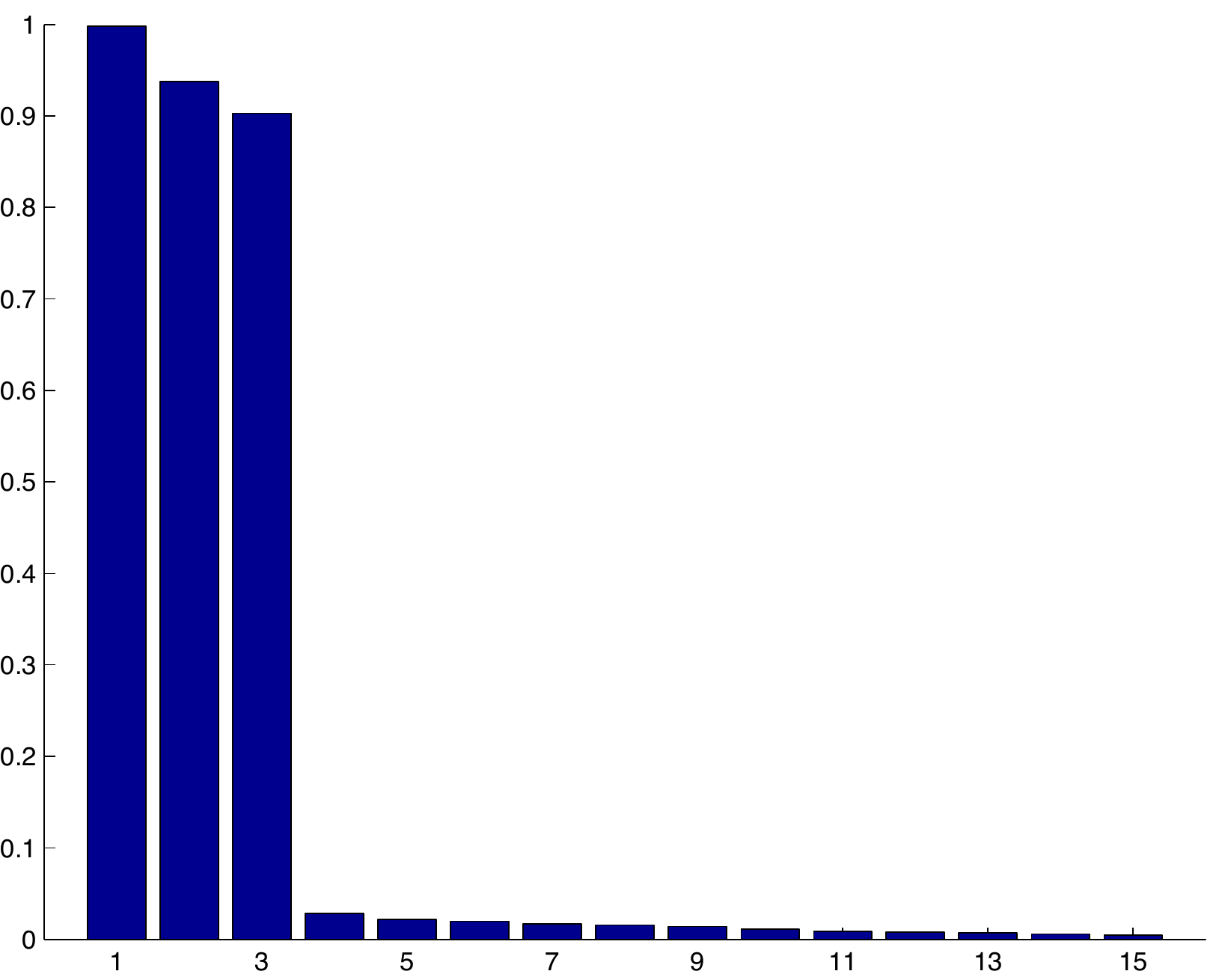}
      \label{fig:toy_pca}
    }
    \hspace{40pt}
    \subfloat[Learned ARD scales]{\includegraphics[width=0.36\textwidth]{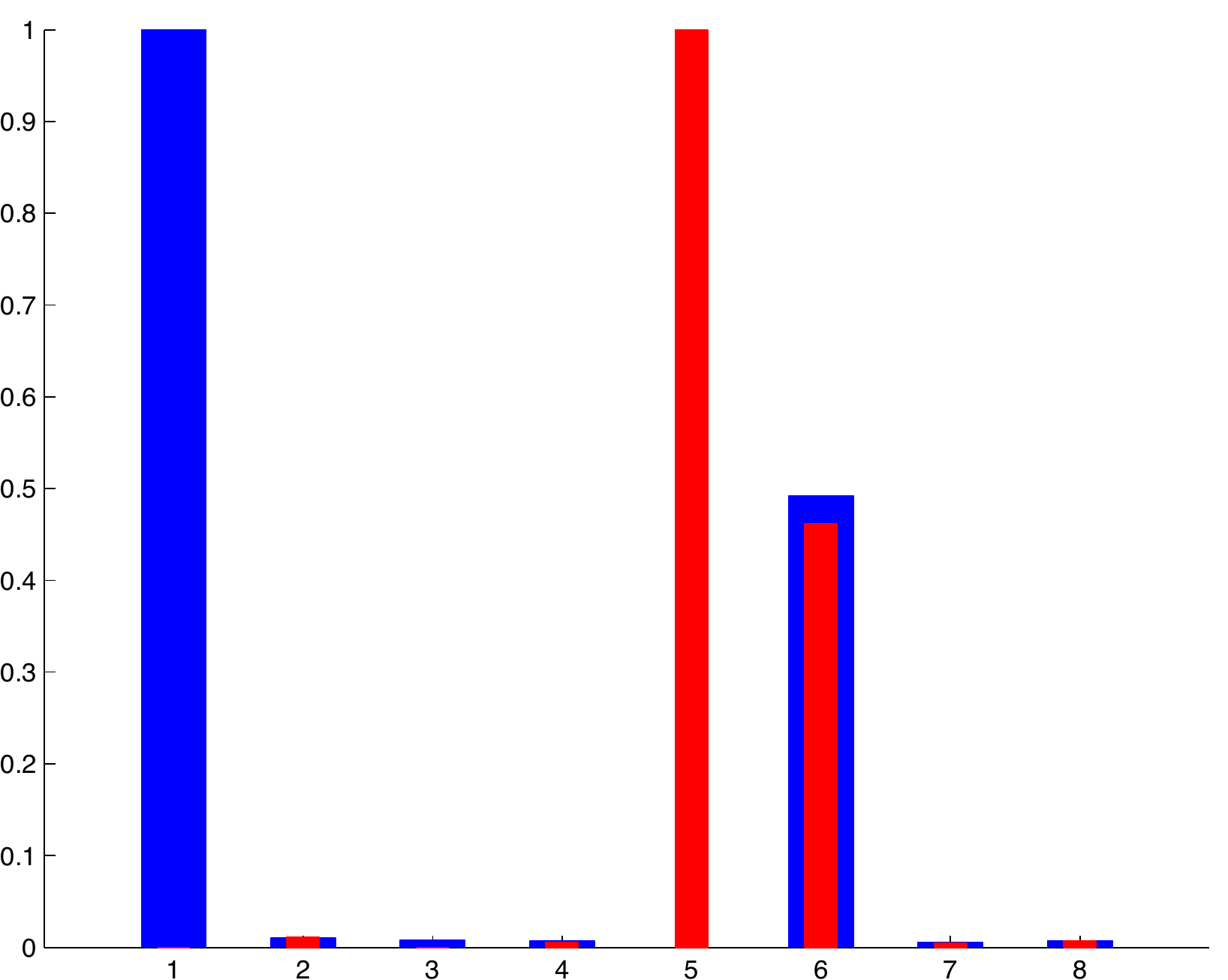}
      \label{fig:toy_scales}
    }
    \\
    \subfloat[Shared signal]{\includegraphics[width=0.317\textwidth]{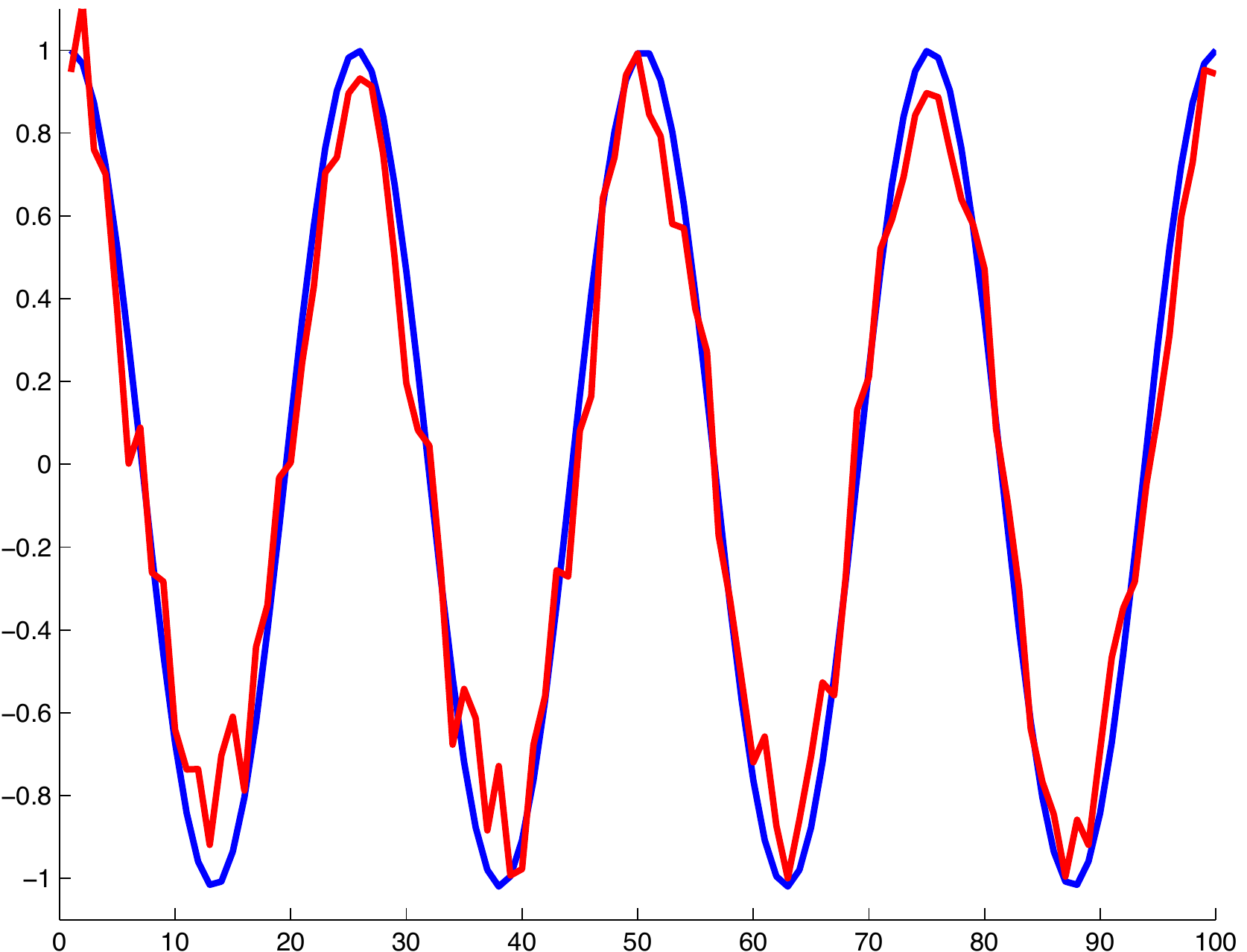}
      \label{fig:toy_shared}
    }
    \subfloat[Private signal for view $\viewY$]{\includegraphics[width=0.317\textwidth]{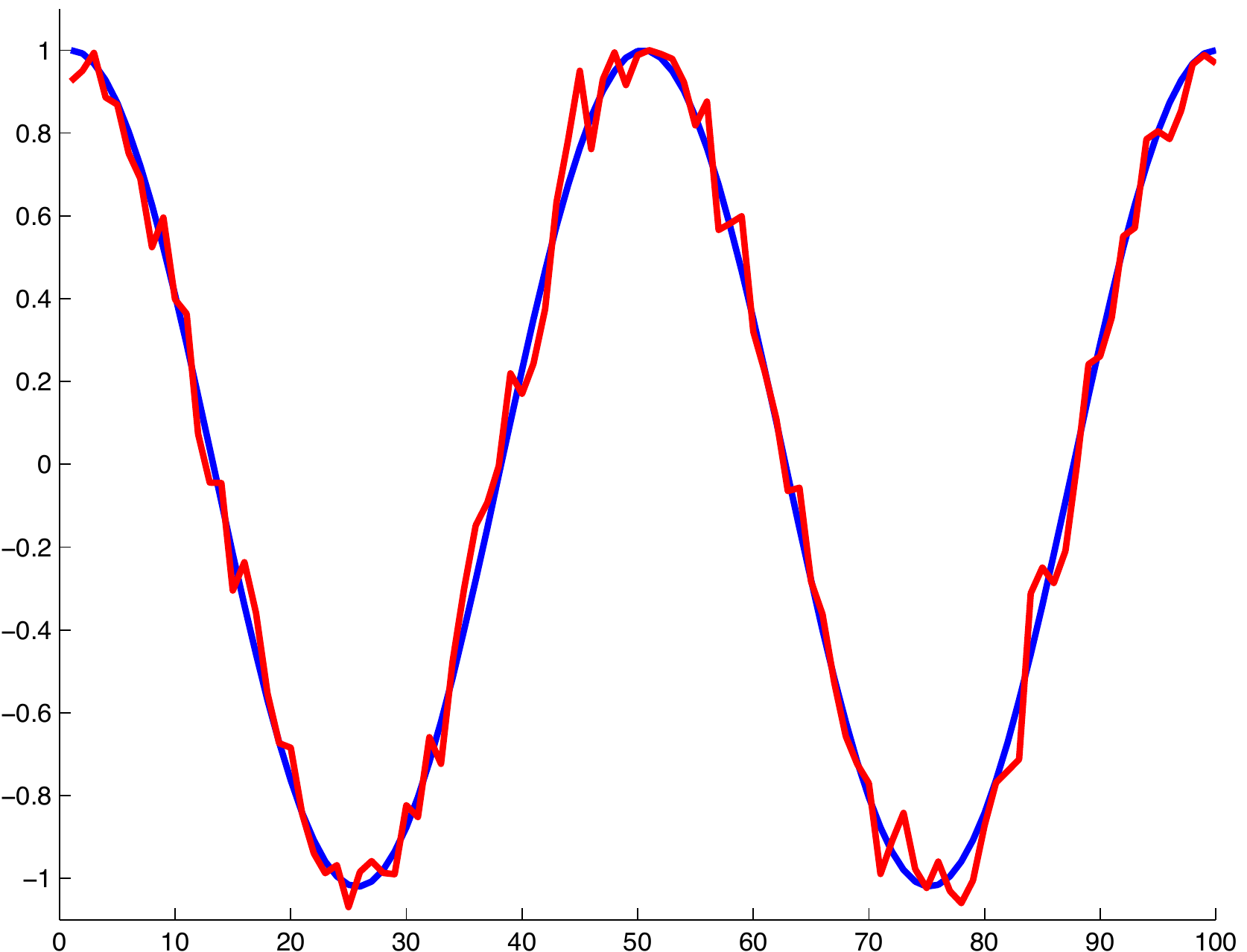}
      \label{fig:toy_private1}
    }
    \subfloat[Private signal for view $\viewY$]{\includegraphics[width=0.317\textwidth]{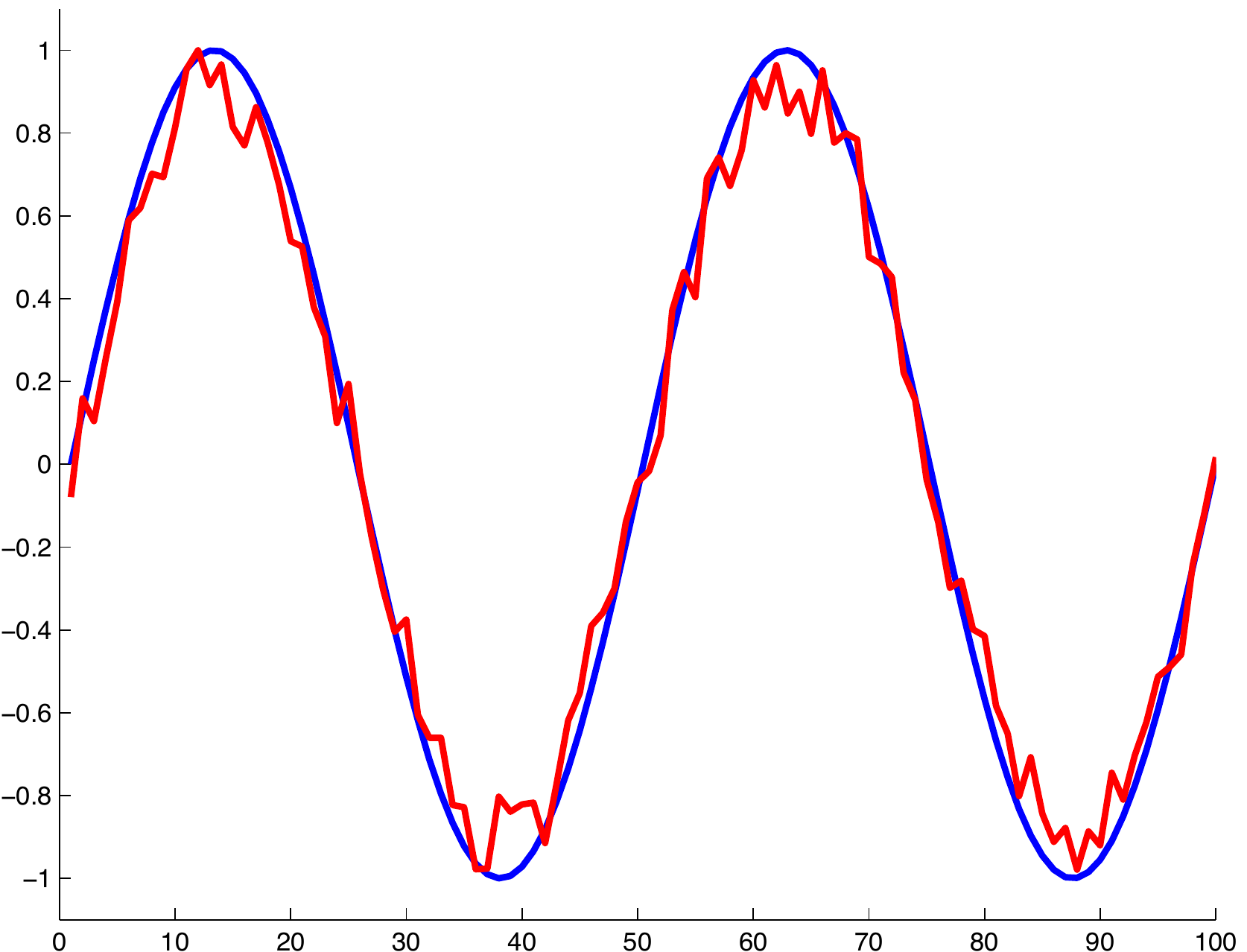}
      \label{fig:toy_private2}
    }
  \end{center}
  \caption[MRD applied to toy data.]{MRD on a toy data set. Initialized with $8$ latent dimensions the model switched off all dimensions except for three (panel \subref{fig:toy_scales}): one private for each observation space and one shared, which corresponds well to the eigenspectrum of the data that clearly shows three variables (panel \subref{fig:toy_pca}). The numbers on the $y-$axis are normalized so that the maximum is 1. The second row depicts the recovered latent signals in red and the generating signals in blue. For easier interpretation, they have been post-processed to remove the scale degree of freedom such that the learned latent space matches the generating signals. Note that no temporal information was given (\ie all points were treated as independent). When the temporal information was given through a prior $p(\mX|\bft)$, the recovered signals were smooth and matched almost exactly the true ones.}
\label{fig:mrdtoy}
\end{figure}

As a first experiment we will use an intuitive toy example similar to the one proposed by \citet{Salzmann:2010vh}. We generate three separate signals: a cosine and a sine, which will be our private signal generators, and a squared cosine as shared signal. We then independently draw three separate random matrices which map the two private signals to $10$ dimensions and the shared signal to $5$ dimensions. The two sets of observations $\viewY$ and $\viewZ$ are then formed by concatenating each respective $10-$dimensional private signal with the $5-$dimensional shared one, also adding isotropic Gaussian noise. Therefore, $\viewy\n,\viewz\n \in \Re^{15}$. Using a GP prior with a linear covariance function, the model should be able to learn a latent representation of the two data sets by recovering the three generating signals, a sine and a cosine as private and the squared cosine as shared. In \fig \ref{fig:mrdtoy} the result of the experiment is shown. The model learns the intrinsic dimensionality of the data and, additionally, is able to recover the factorization used for generating the data. We also experimented with adding a temporal prior on the latent space. In this case we encapsulate the prior knowledge that the recovered signals should be smooth. In this case the recovered signals almost exactly match the true ones and, therefore, we have not included this plot. We will therefore now proceed to apply the model to more challenging data where the generating parameters and their structure are truly unobserved.

\begin{figure}[t]
   \begin{center}
     \includegraphics[width=0.4\columnwidth]{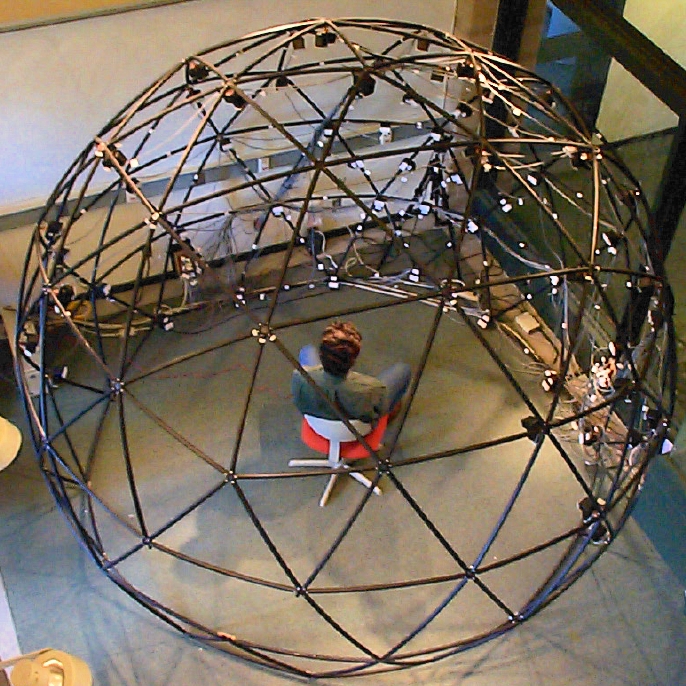}
   \end{center}
   \caption[``Yale Faces'' data generation procedure.]{The mechanism used to generate the Yale Faces data set \citep{Georghiades:2001tc}.}
   \label{fig:lightdome}
 \end{figure}

\subsection{Yale Faces \label{sec:MRD_experimentYaleFaces}}
The Yale face database B \citep{Georghiades:2001tc} is a collection of images depicting different individuals in different poses under controlled lighting conditions. The data set contains $10$ individuals in $9$ different poses each lighted from $64$ different directions. The different lighting directions are positions on a half sphere as can be seen in \fig \ref{fig:lightdome}. The images for a specific pose are captured in rapid procession such that the variations in the image for a specific person and pose should mainly be associated with the light direction. This makes the data interesting from a dimensionality reduction point of view, as the representation is very high-dimensional, $192\times168=32256$ pixels, while the generating parameters, \ie the lighting directions and pose parameters, are very low dimensional.
There are several different ways of using this data in the MRD framework, depending on which correspondence aspect of the data is used to align the different views. We chose to use all illuminations for a single pose. We generate two separate data sets, $\viewY$ and $\viewZ$, by splitting the images into two sets such that the two views contain three different subjects. The order of the data was such that the lighting direction of each $\yn$ matched that of $\viewz\n$ while the subject identity was random, such that no correspondence was induced between different faces. As such, the model should learn a latent structure factorized into lighting-related parameters (a point on a half-sphere) and subject-related parameters, where the first are shared and the latter private to each observation space.

    \begin{figure}[t]
    \begin{center}
    \subfloat[The scale vector $\vw^\viewYi$]{
    \includegraphics[width=0.33\textwidth]{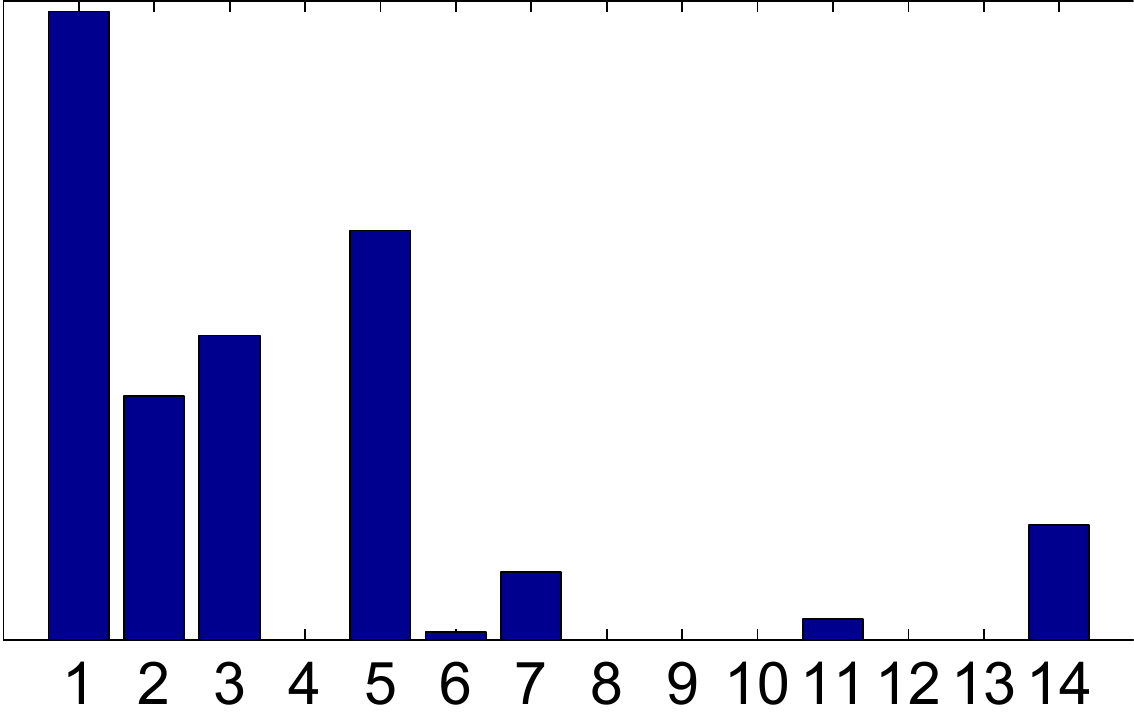}
        \label{fig:yale6SetsScales1}
    }
    \hspace{20pt}
    \subfloat[The scale vector $\vw^\viewZi$]{
        \includegraphics[width=0.33\textwidth]{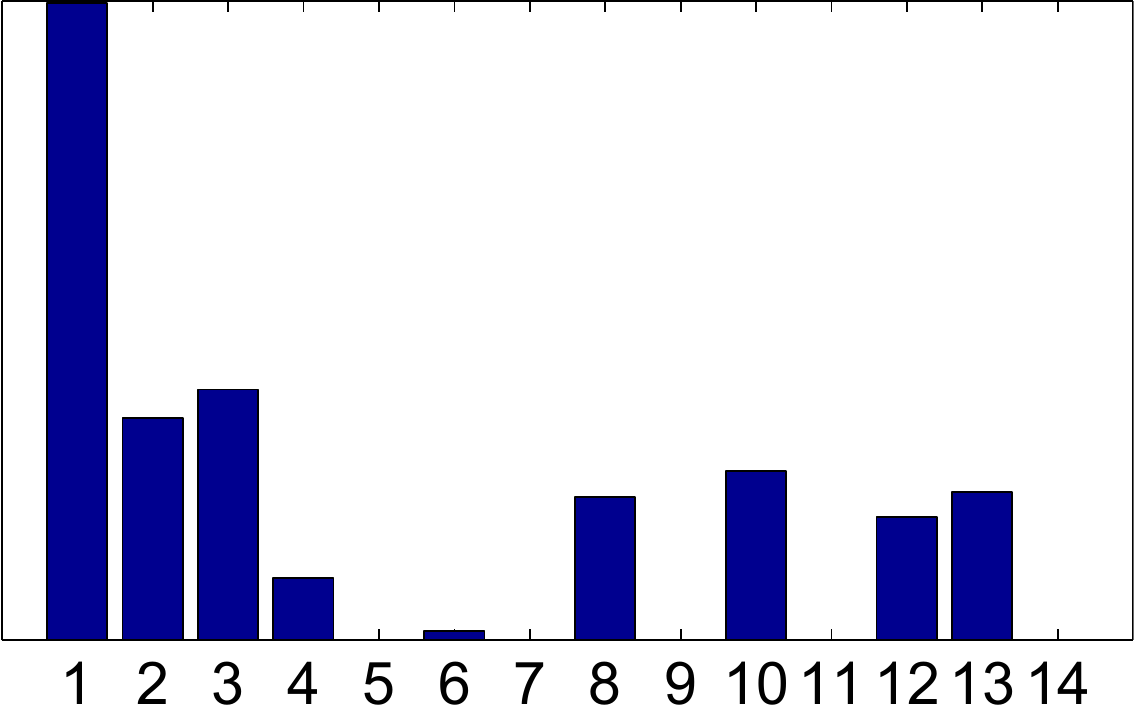}
        \label{fig:yale6SetsScales2}
    }
    \end{center}
    \caption[MRD: Relevance weights for the Yale faces experiment.]{
    The relevance  weights for the faces data. Despite allowing for soft sharing, the first 3 dimensions are switched on with approximately the same weight for both views of the data. Most of the remaining dimensions are used to explain private variance.
    }
    \label{fig:yale6SetsScales}
    \end{figure}
\begin{figure}[t]
    \begin{center}
    \subfloat[]{
    \includegraphics[width=0.32\textwidth]{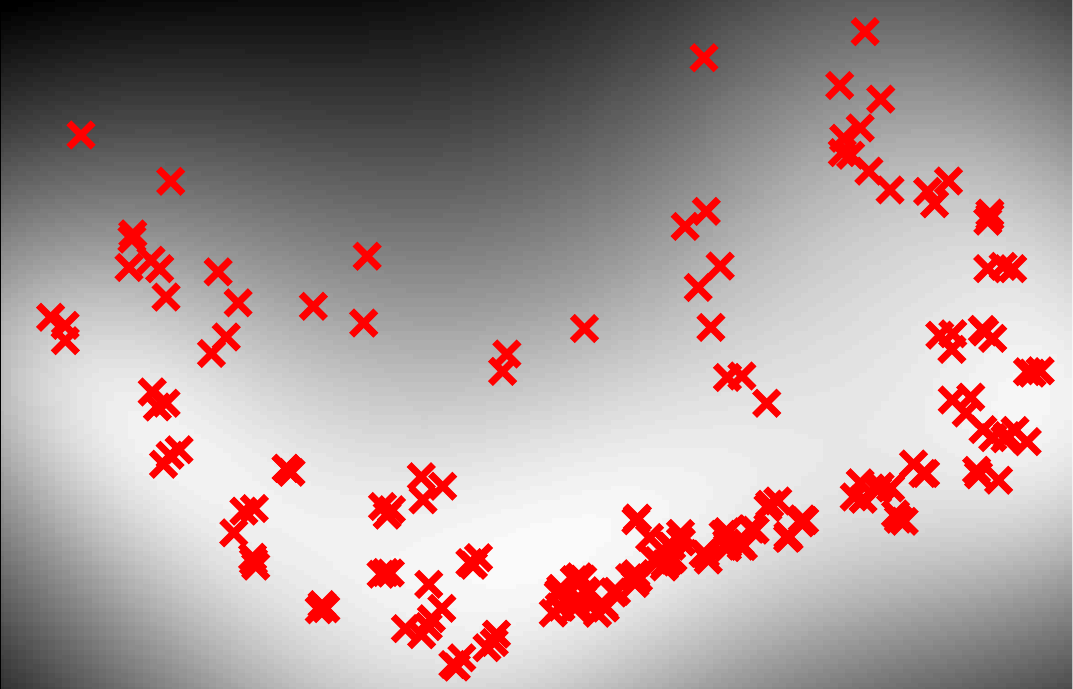}
        \label{fig:yale6SetsX12}
    }
    \subfloat[]{
        \includegraphics[width=0.32\textwidth]{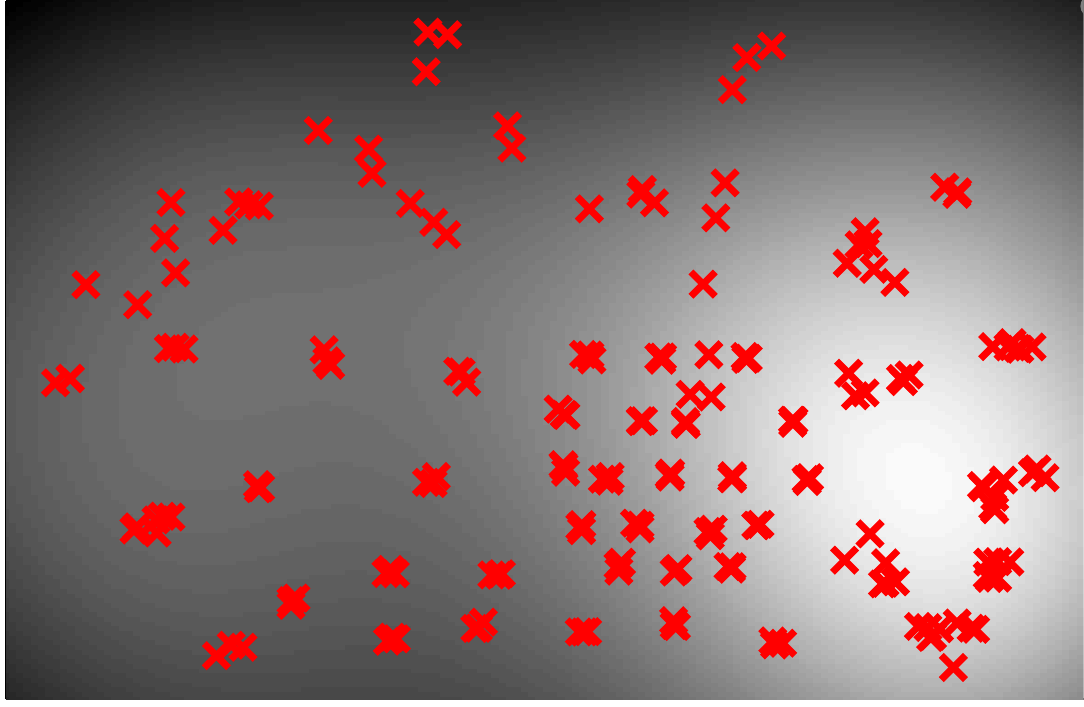}
        \label{fig:yale6SetsX13}
    }
    \subfloat[]{
        \includegraphics[width=0.32\textwidth]{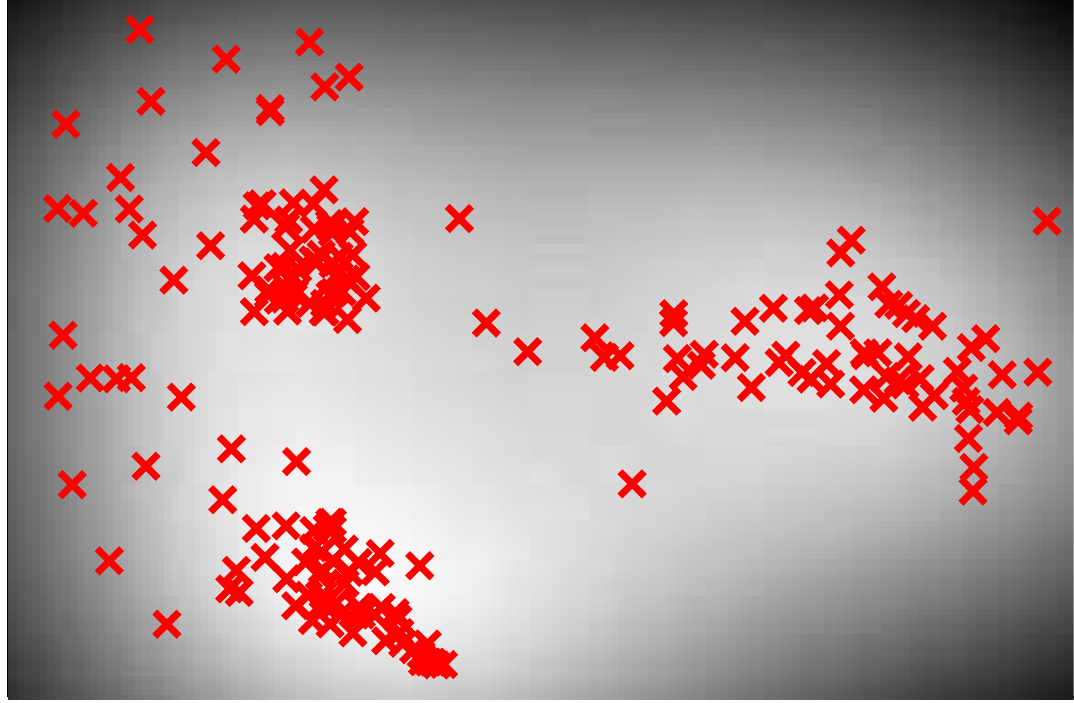}
        \label{fig:yale6SetsX5_14}
    }
    \end{center}
    \caption[MRD: Latent projections for the Yale faces experiment.]{
    Projection of the shared latent space into dimensions $\{1,2\}$ and $\{1,3\}$ (figures
    \protect\subref{fig:yale6SetsX12} and \protect\subref{fig:yale6SetsX13}) and projection of the $\viewY-$private dimensions $\{5,14\}$
    (figure \protect\subref{fig:yale6SetsX5_14}).
    Red x's represent (projected) locations of latent points that correspond to the training data. The greyscale intensities of the background are proportional to the predicted variance of the GP mapping, if the corresponding locations were given as inputs.
    MRD successfully factorized the latent space so that the latent points in Figure
    \protect\subref{fig:yale6SetsX5_14} form three clusters, each responsible for modelling one of the three faces in $\viewY$.
    }
    \label{fig:yale6SetsLatentSpace}
    \end{figure}

The optimized relevance weights $\{\vw^\viewYi, \vw^\viewZi\}$ are visualized as bar graphs in \fig \ref{fig:yale6SetsScales}. The latent space is clearly factorized into a shared part, consisting of dimensions indexed\footnote{Dimension 6 also encodes shared information, but of almost negligible amount ($\vw^\viewYi_6, \vw^\viewZi_6 \approx 0$).} as $1$,$2$ and $3$, two private and an irrelevant part (dimension $9$). The two data views correspond to approximately equal weights for the shared latent dimensions. Projections onto these dimensions are visualized in Figures \ref{fig:yale6SetsLatentSpace}\subref{fig:yale6SetsX12} and \ref{fig:yale6SetsLatentSpace}\subref{fig:yale6SetsX13}. Even though not immediately obvious from these two-dimensional projections, interaction with the shared latent space reveals that it actually has the structure of a half sphere, recovering the shape of the space defined by the fixed locations of the light source shown in \fig \ref{fig:lightdome}.

\begin{figure}[t]
    \begin{center}
    \subfloat[]{
    \includegraphics[width=0.3\textwidth]{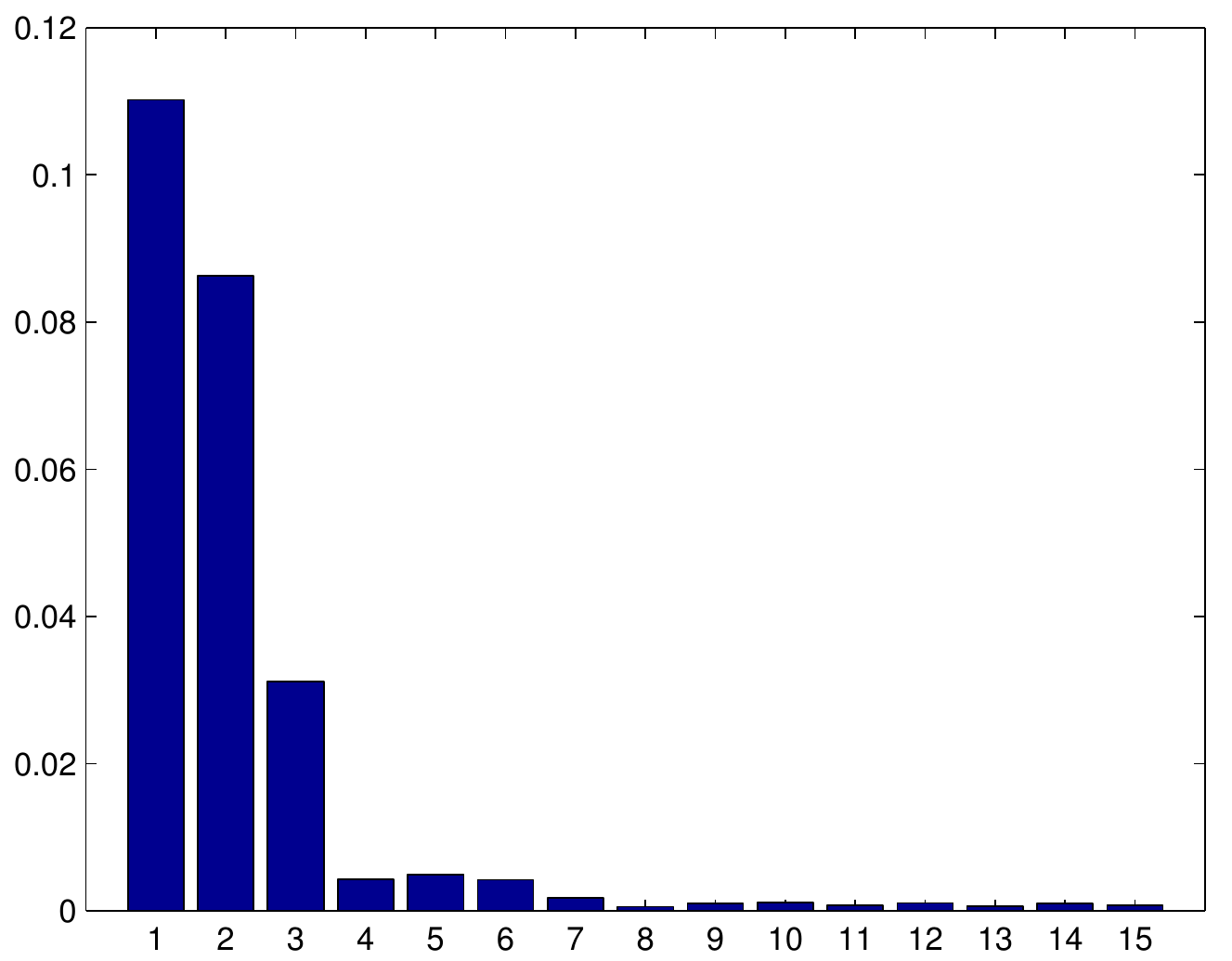}
        \label{fig:yaleOneFaceScales}
    }
    \subfloat[]{
        \includegraphics[width=0.31\textwidth]{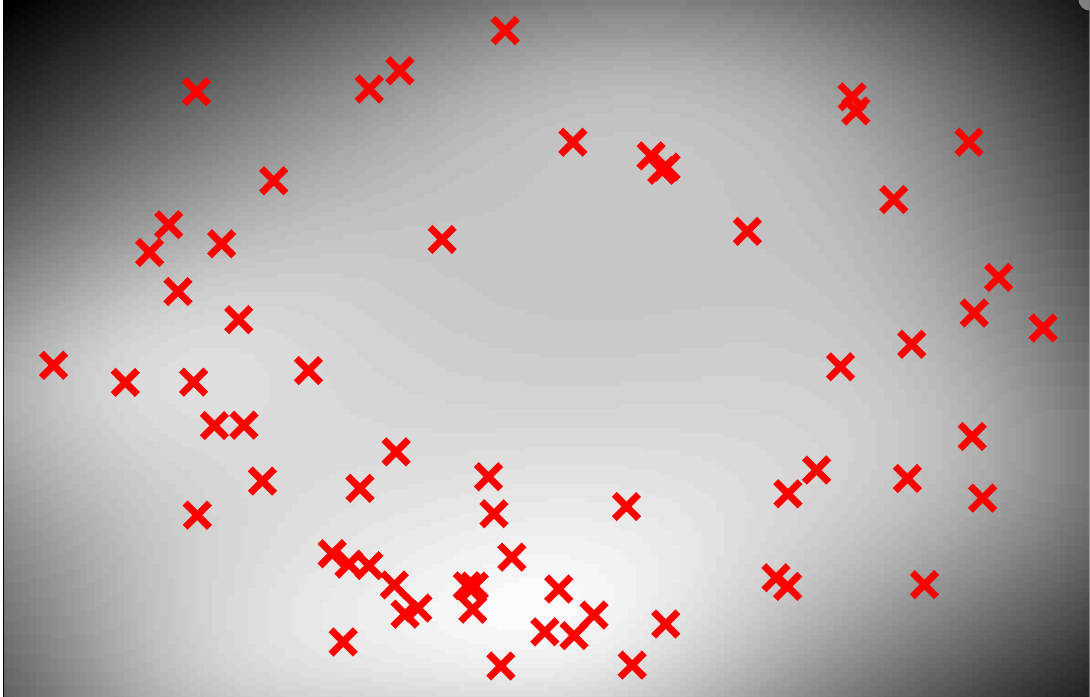}
        \label{fig:yaleOneFaceX21}
    }
    \subfloat[]{
        \includegraphics[width=0.31\textwidth]{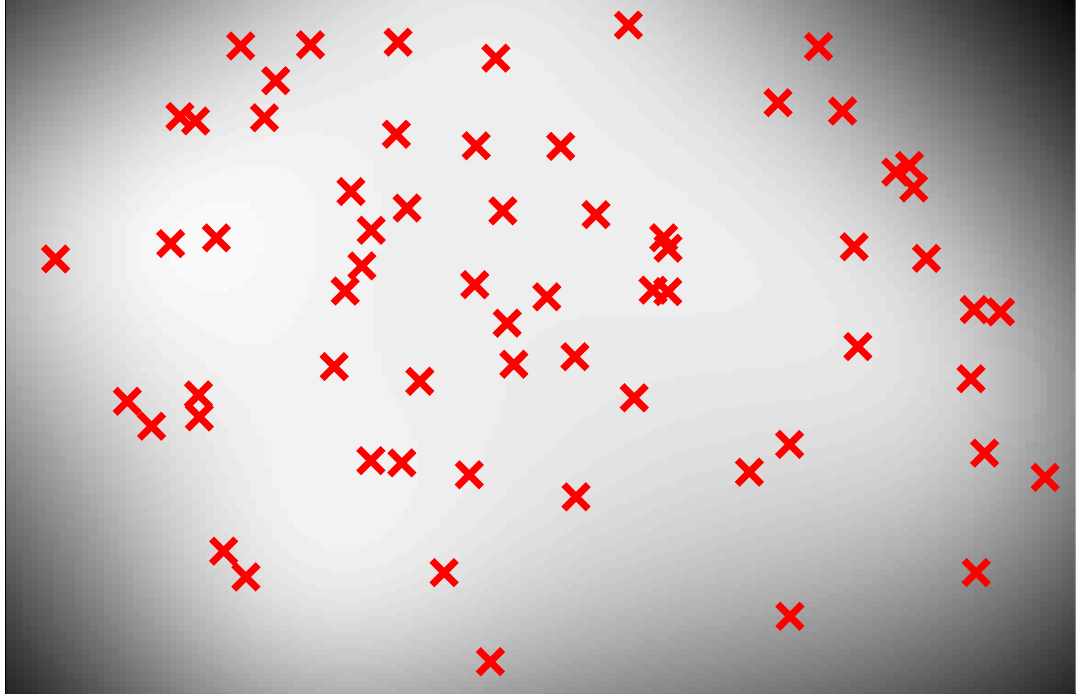}
        \label{fig:yaleOneFaceX23}
    }
    \end{center}
    \vspace{-4pt}
    \caption[Variational GP-LVM applied to one of the views of the Yale faces data: ARD weights and latent projections.]{
    Latent space learned by the  Bayesian GP-LVM for a single face view. The weight set $\bfw$ associated with the learned latent space is shown in \protect\subref{fig:yaleOneFaceScales}. In figures \protect\subref{fig:yaleOneFaceX21} and \protect\subref{fig:yaleOneFaceX23} we plotted pairs of the $3$ dominant latent dimensions against each other. Dimensions $4,5$ and $6$ have a very small but not negligible weight and represent other minor differences between pictures of the same face, as the subjects often blink, smile etc.
    }
    \label{fig:yaleOneFace1}
    \end{figure}

By projecting the latent space onto the dimensions corresponding to the private spaces,  we essentially factor out the variations generated by the light direction. As can be seen in \fig \ref{fig:yale6SetsLatentSpace}\subref{fig:yale6SetsX5_14}, the model then separately represents the face characteristics of each of the subjects. This indicates that the shared space successfully encodes the information about the position of the light source and not the face characteristics. This indication is enhanced by the results found when we performed dimensionality reduction with the Bayesian GP-LVM for pictures corresponding to all illumination conditions of a single face (\ie a data set with one modality). Specifically, the latent space discovered by the Bayesian GP-LVM and the shared subspace discovered by MRD have the same dimensionality and similar structure, as can be seen in \fig \ref{fig:yaleOneFace1}.
As for the private manifolds discovered by MRD, these correspond to subspaces for disambiguating between faces of the same view. Indeed, plotting the largest two dimensions of the first latent private subspace against each other in Figure \ref{fig:yale6SetsLatentSpace}\subref{fig:yale6SetsX5_14} reveals three clusters, corresponding to the three different faces within the data set. Similarly to the Bayesian GP-LVM applied to a single face, here the private dimensions with very small weight model slight changes across faces of the same subject (blinking etc).

\begin{figure}[t]
\begin{center}
\newcommand{\tmpYaleFigSpace}{0.095}
\subfloat{ \includegraphics[width=\tmpYaleFigSpace\textwidth]{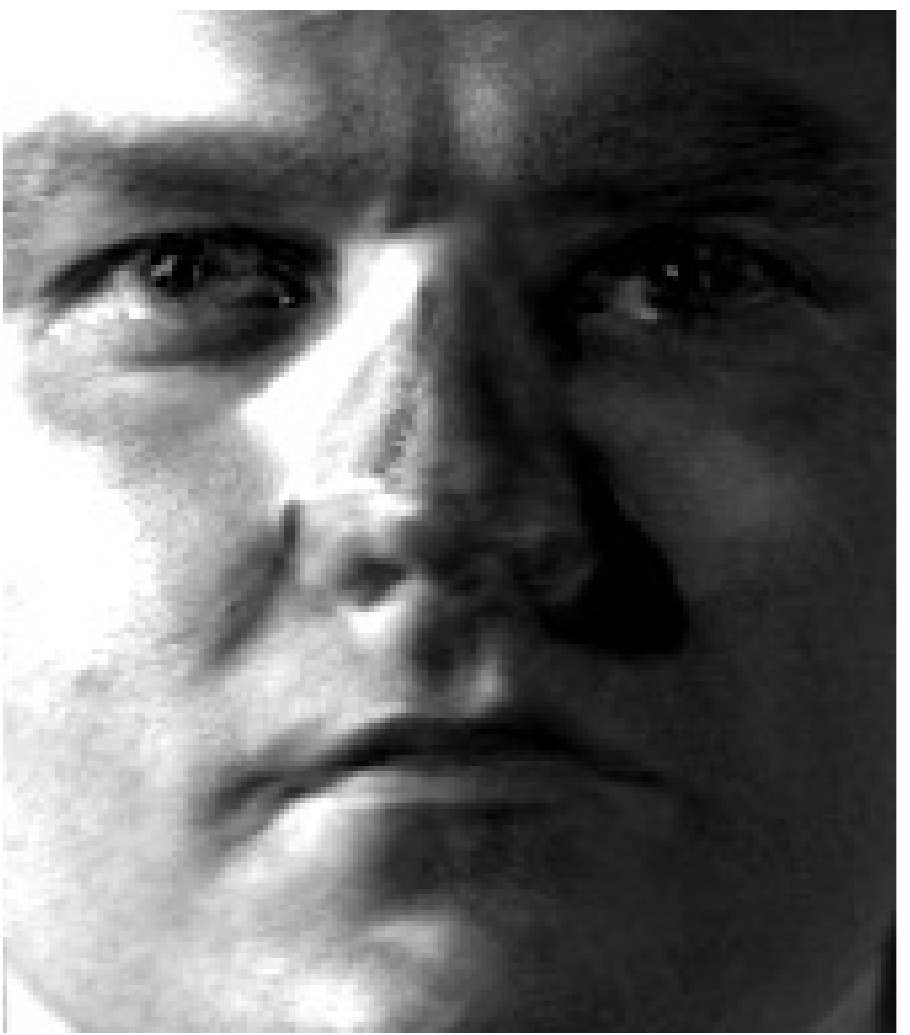} }
\subfloat{ \includegraphics[width=\tmpYaleFigSpace\textwidth]{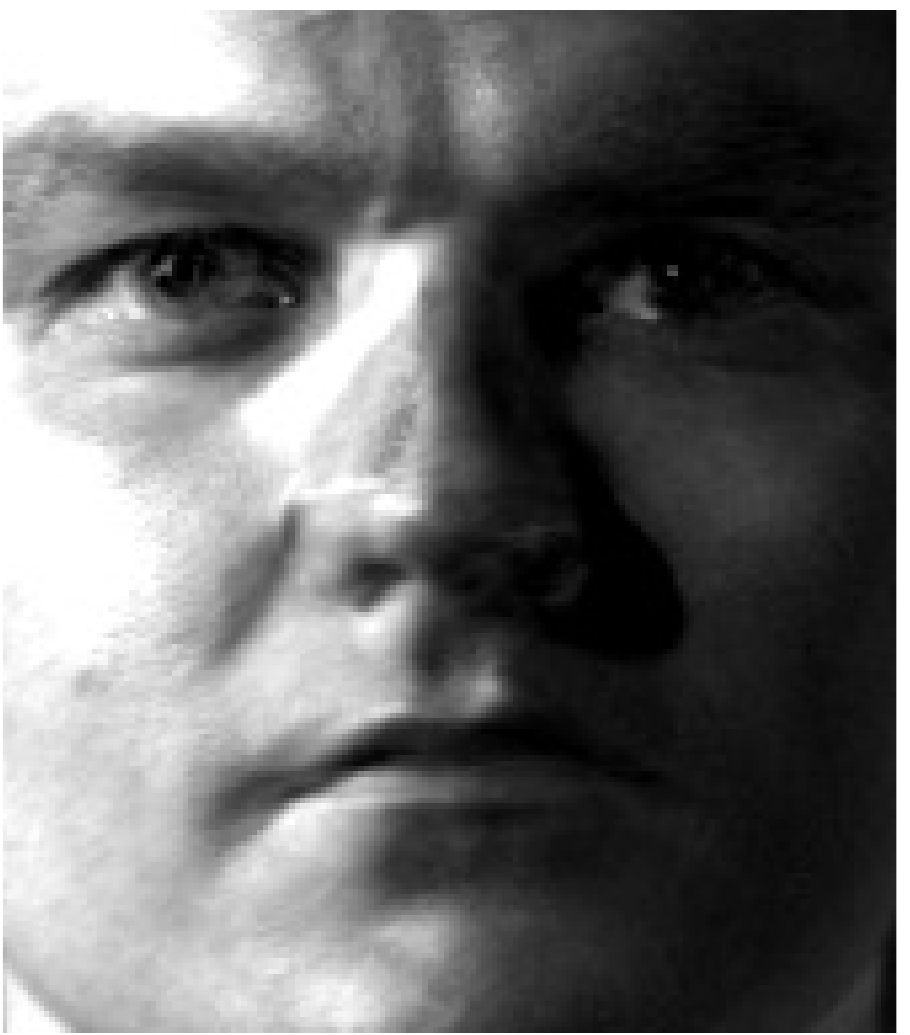} }
\subfloat{ \includegraphics[width=\tmpYaleFigSpace\textwidth]{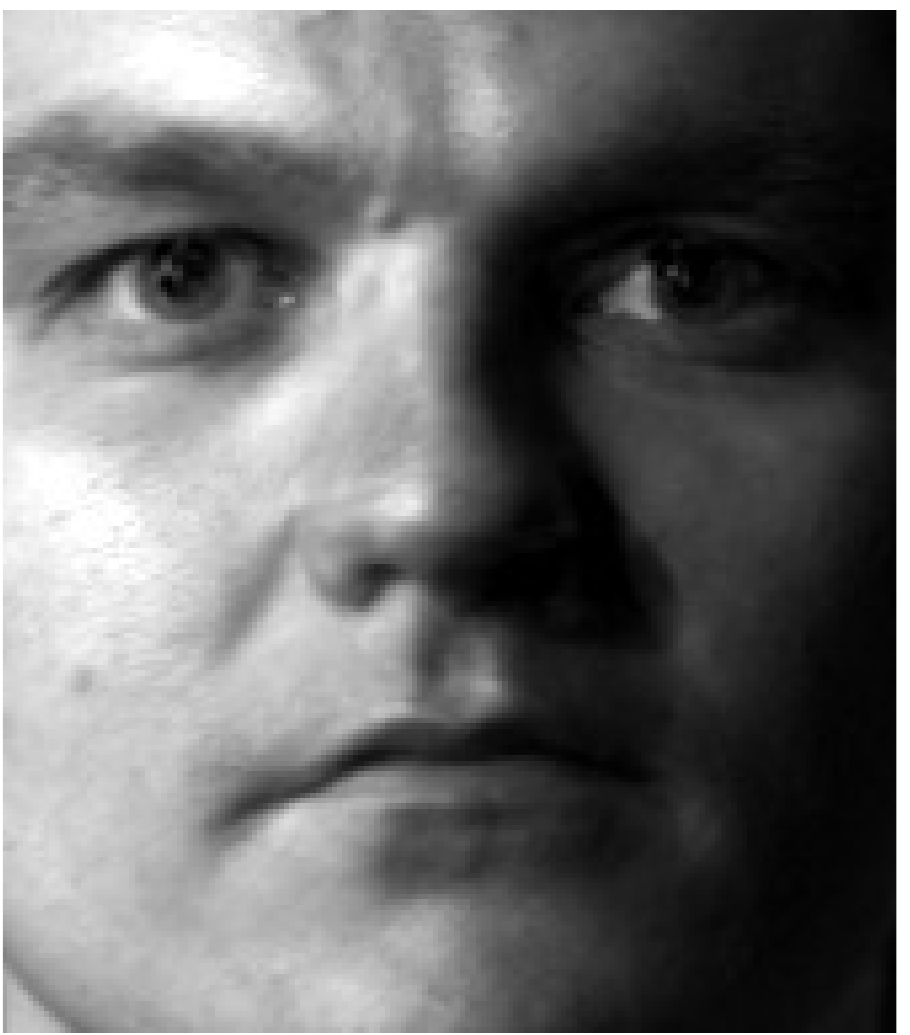} }
\subfloat{ \includegraphics[width=\tmpYaleFigSpace\textwidth]{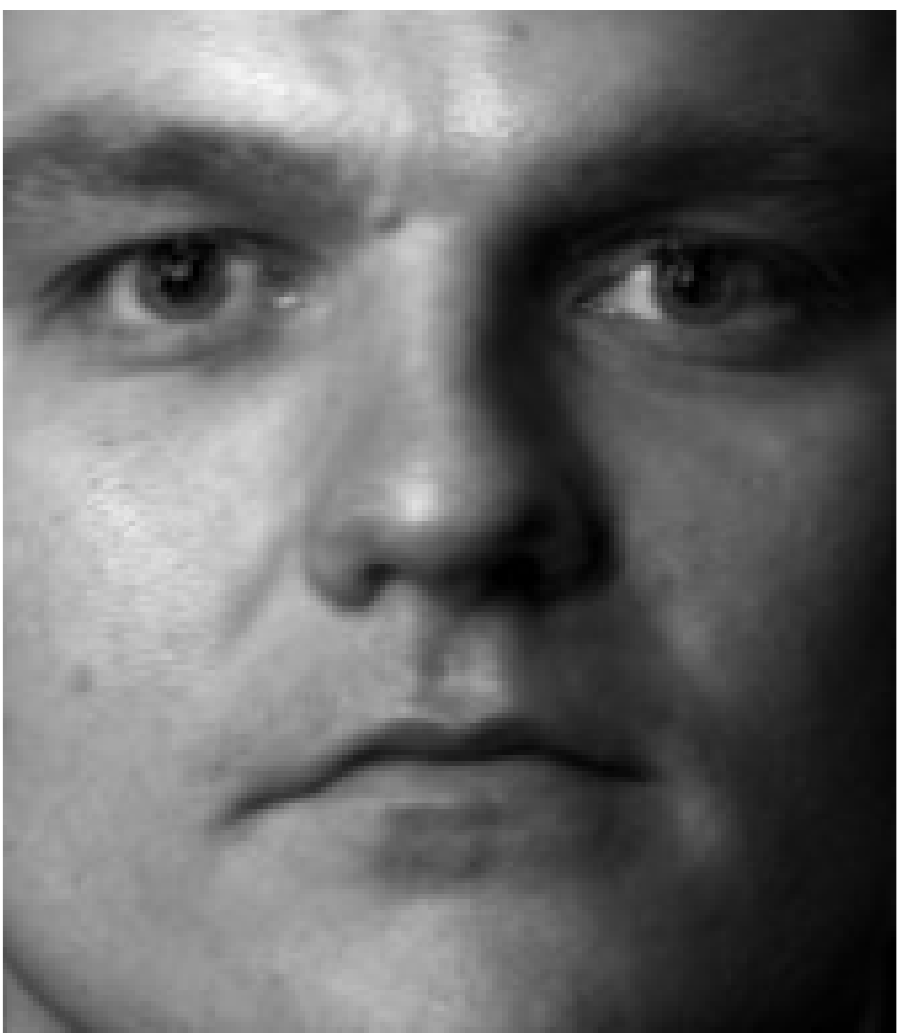} }
\subfloat{ \includegraphics[width=\tmpYaleFigSpace\textwidth]{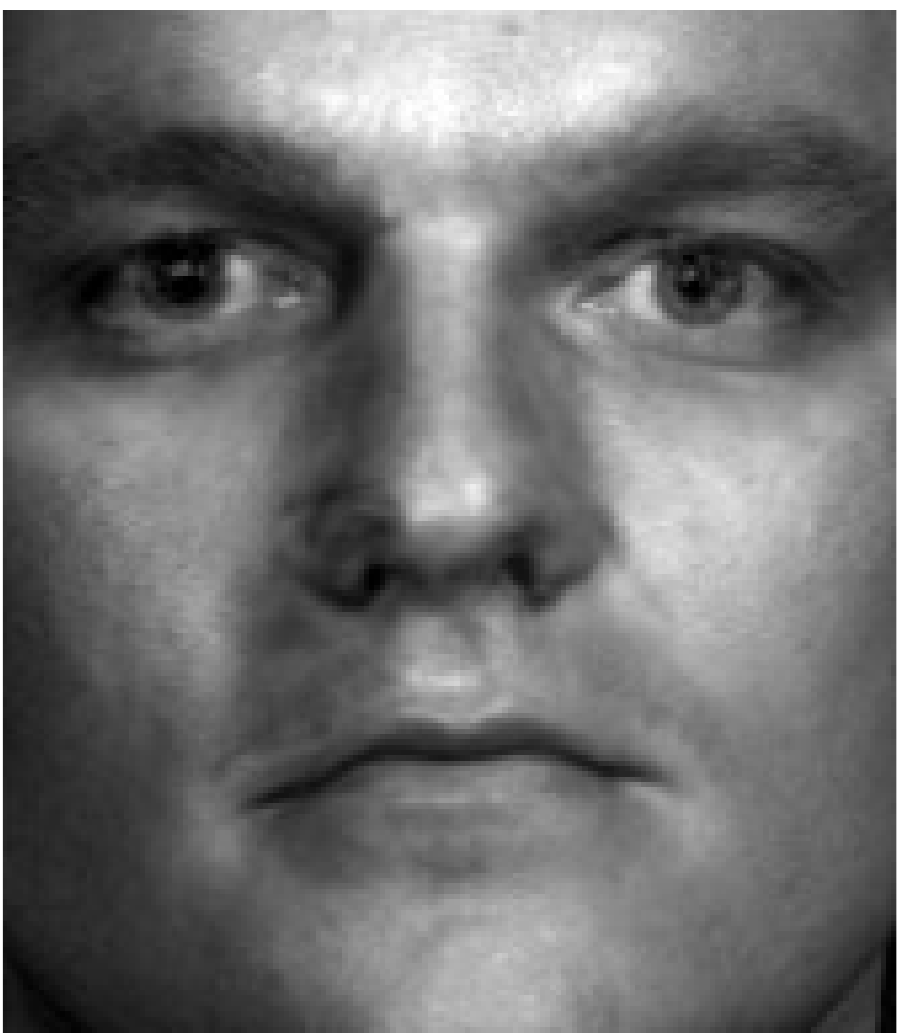} }
\subfloat{ \includegraphics[width=\tmpYaleFigSpace\textwidth]{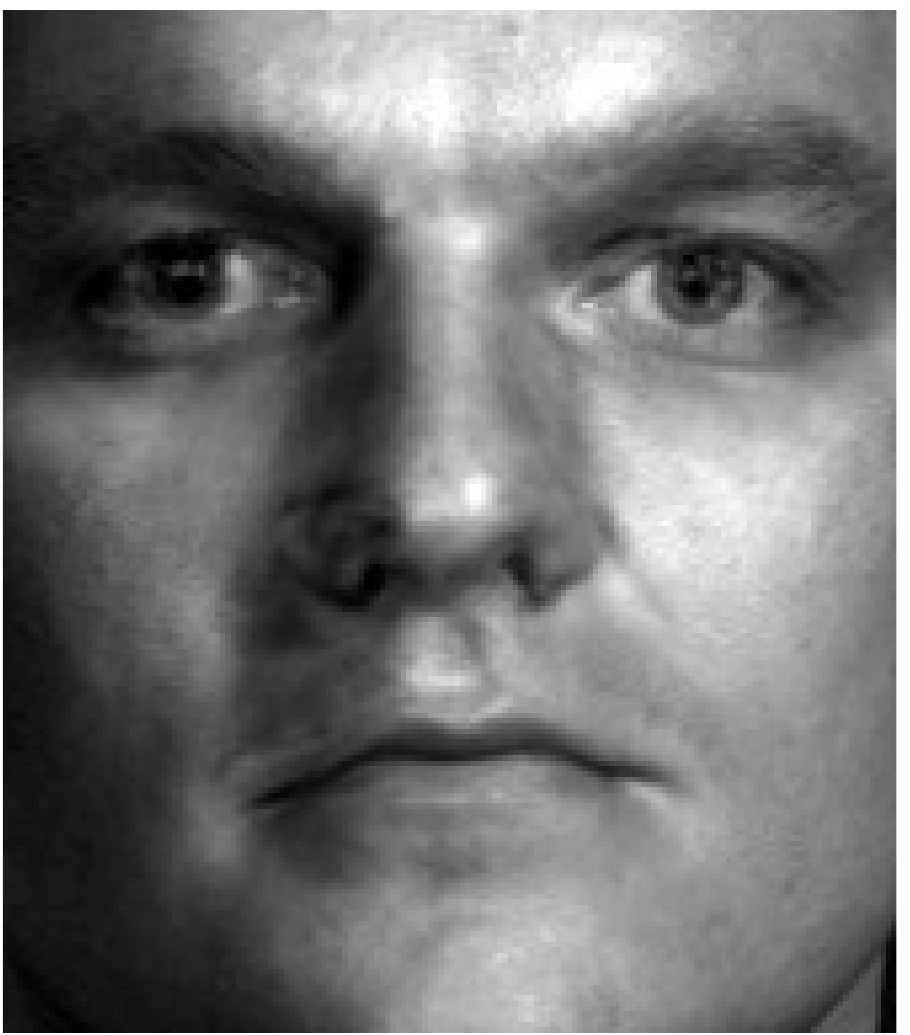} }
\subfloat{ \includegraphics[width=\tmpYaleFigSpace\textwidth]{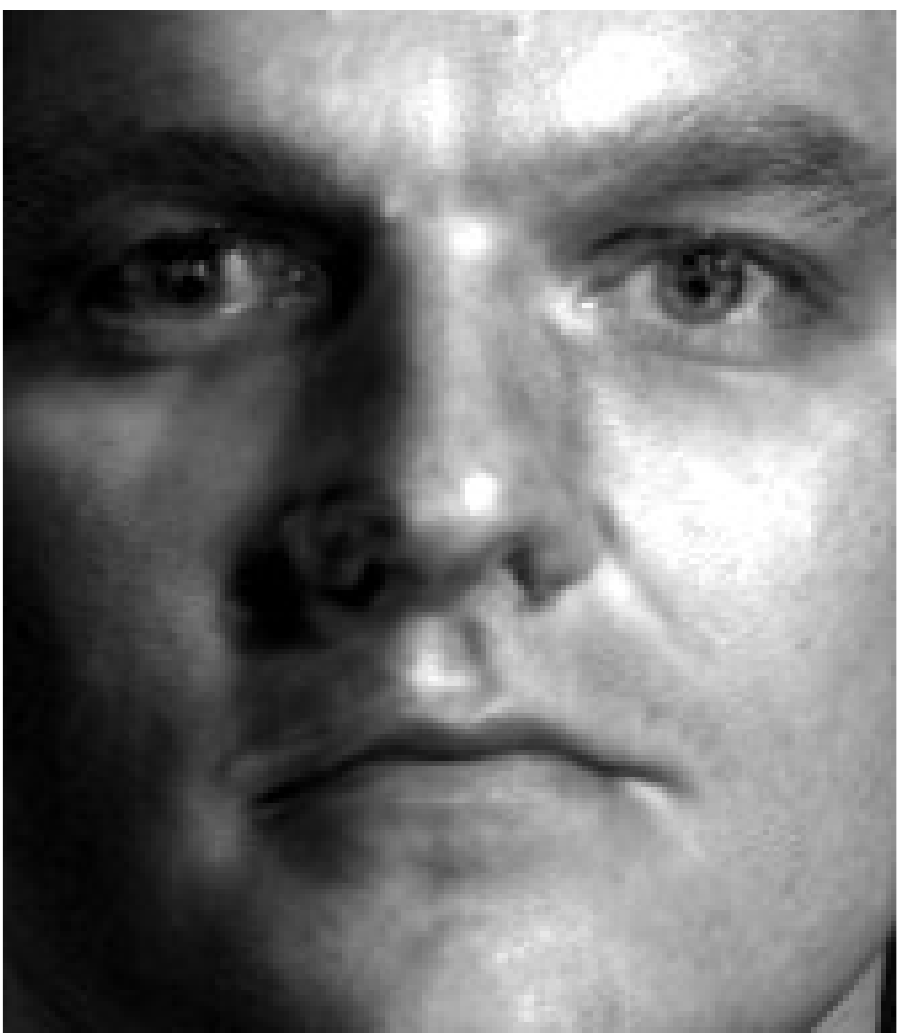} }
\subfloat{ \includegraphics[width=\tmpYaleFigSpace\textwidth]{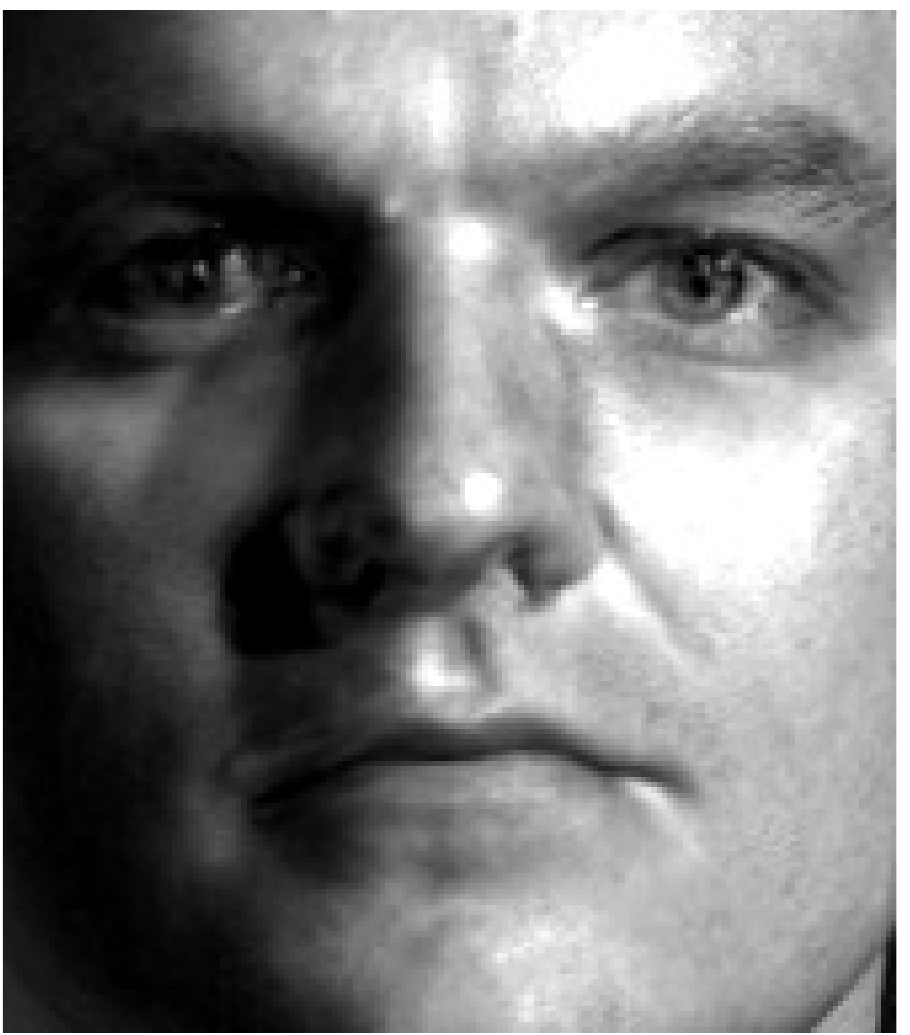} }
\subfloat{ \includegraphics[width=\tmpYaleFigSpace\textwidth]{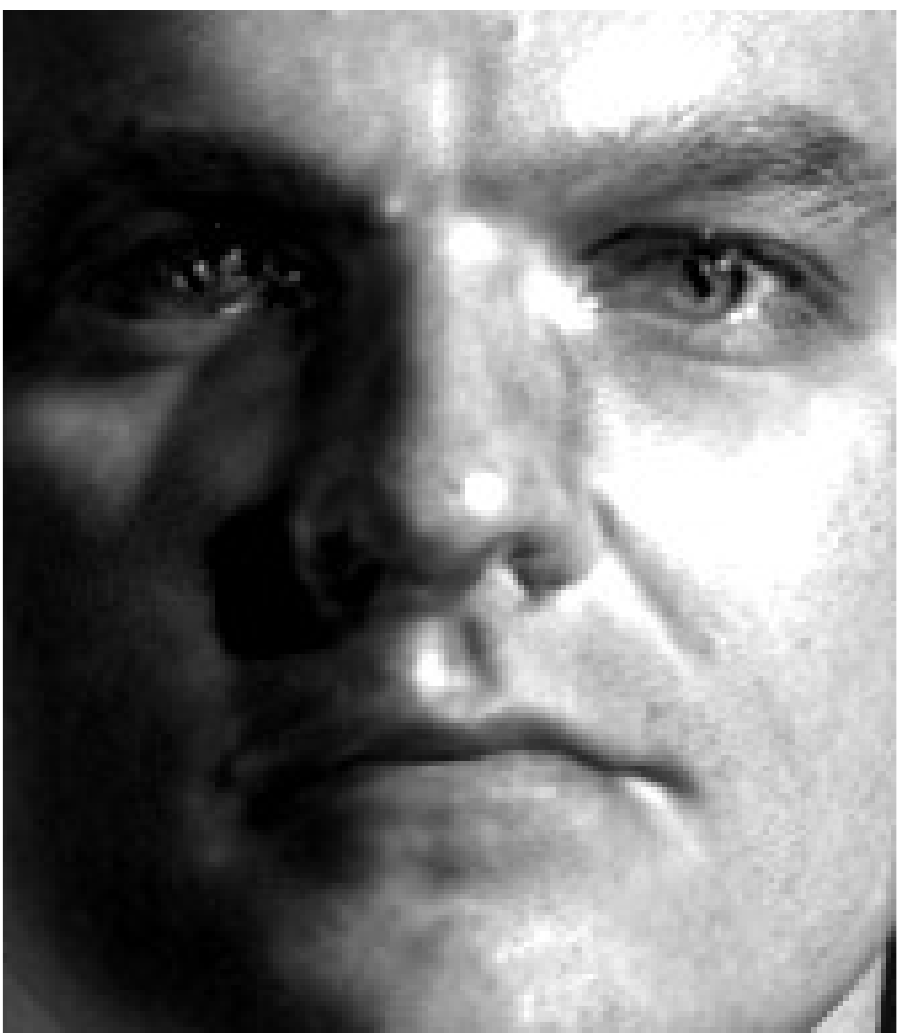} }
\vspace{-4pt}
\newline
\subfloat{ \includegraphics[width=\tmpYaleFigSpace\textwidth]{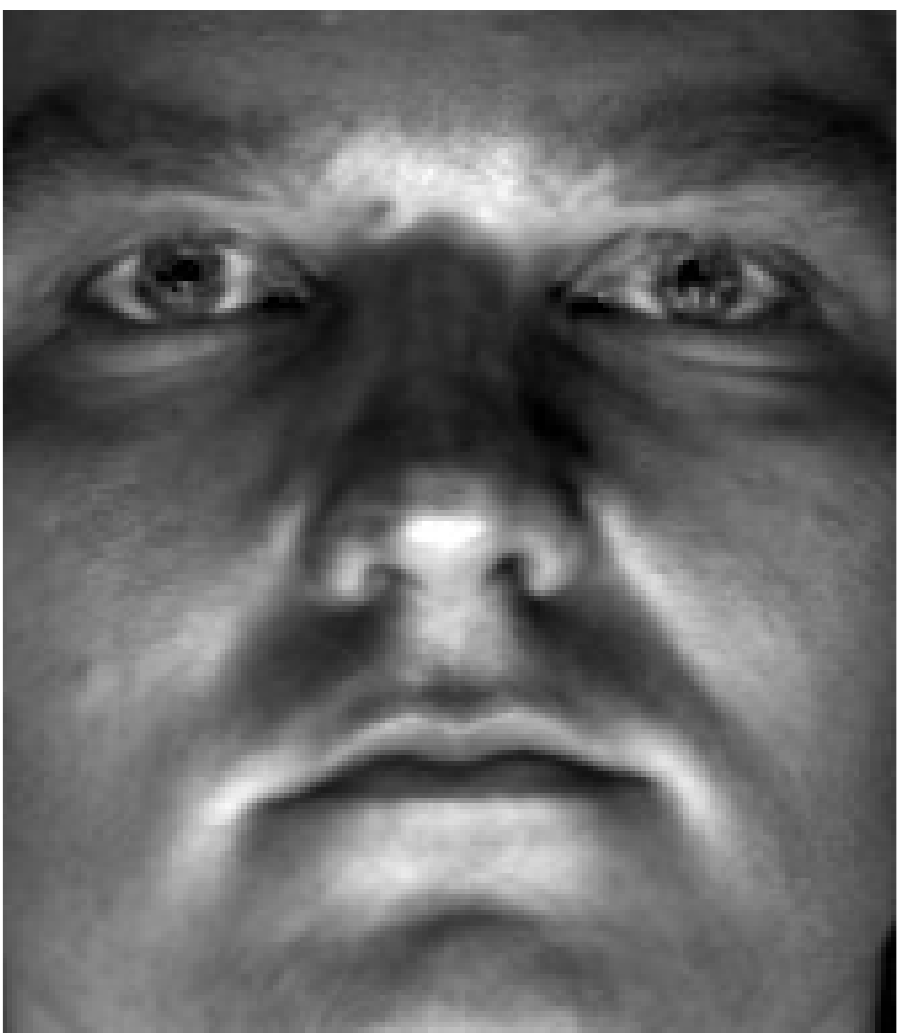} }
\subfloat{ \includegraphics[width=\tmpYaleFigSpace\textwidth]{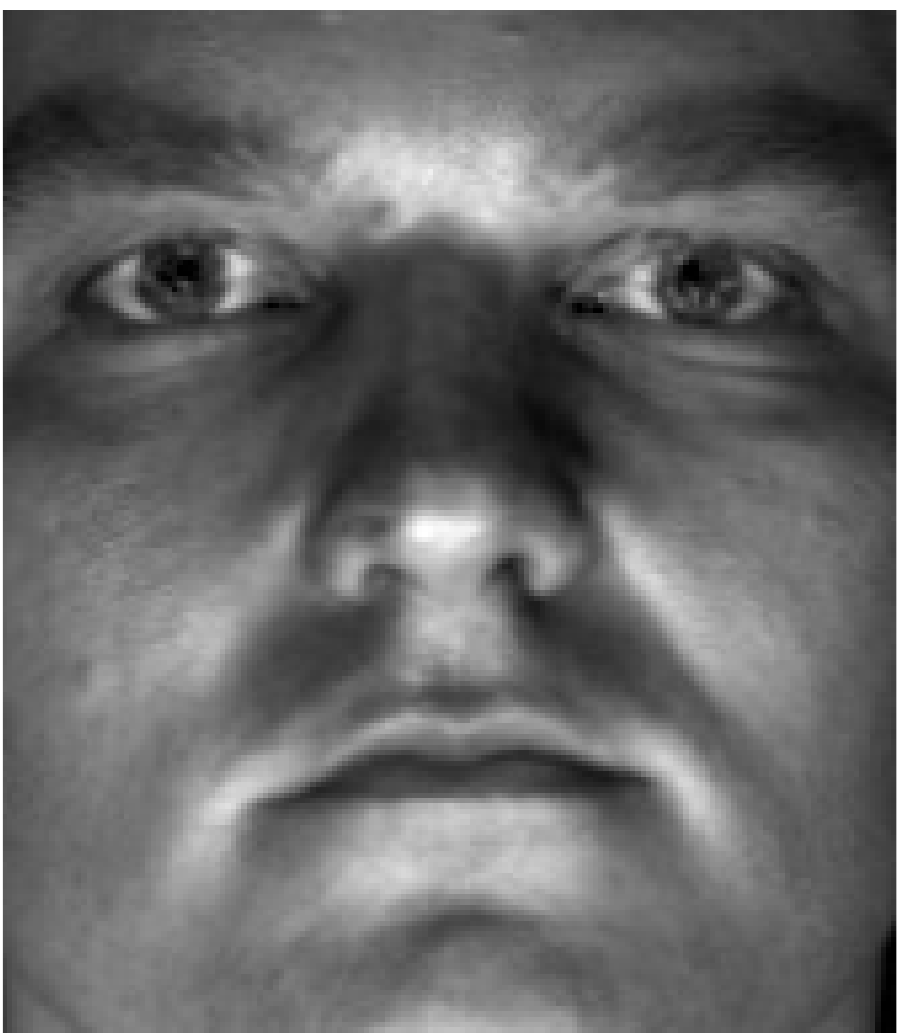} }
\subfloat{ \includegraphics[width=\tmpYaleFigSpace\textwidth]{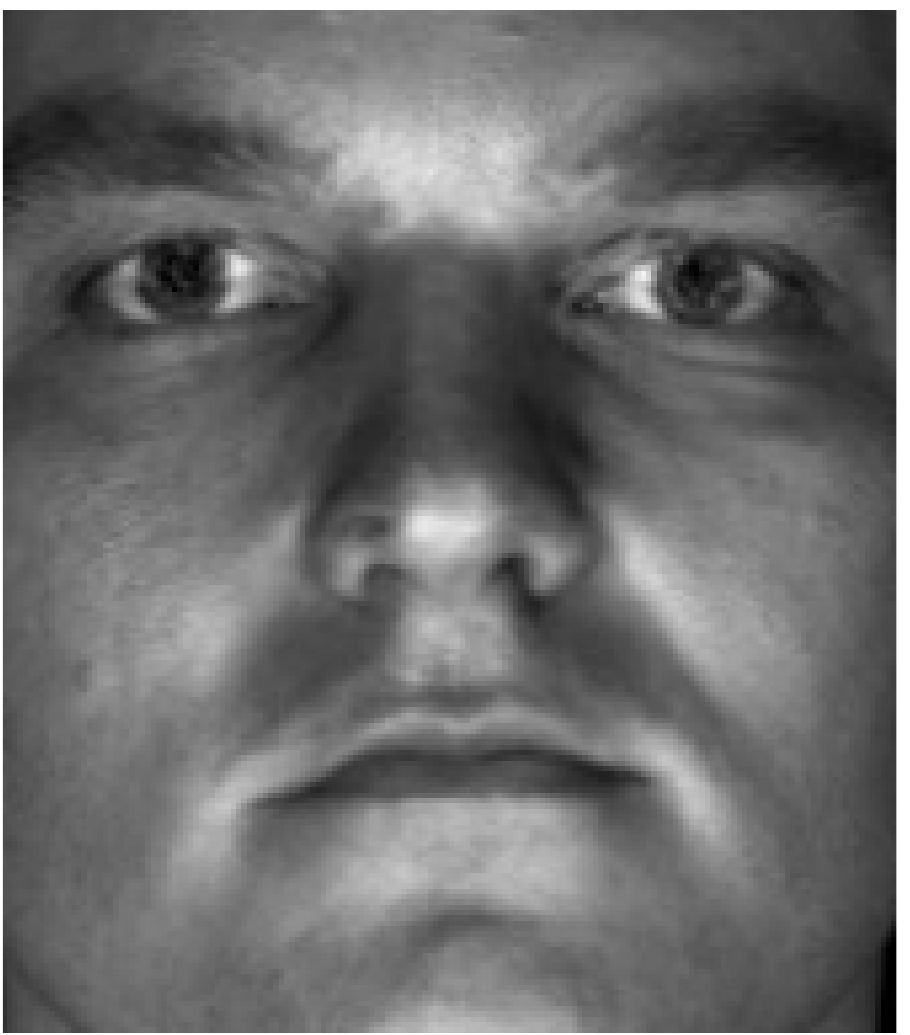} }
\subfloat{ \includegraphics[width=\tmpYaleFigSpace\textwidth]{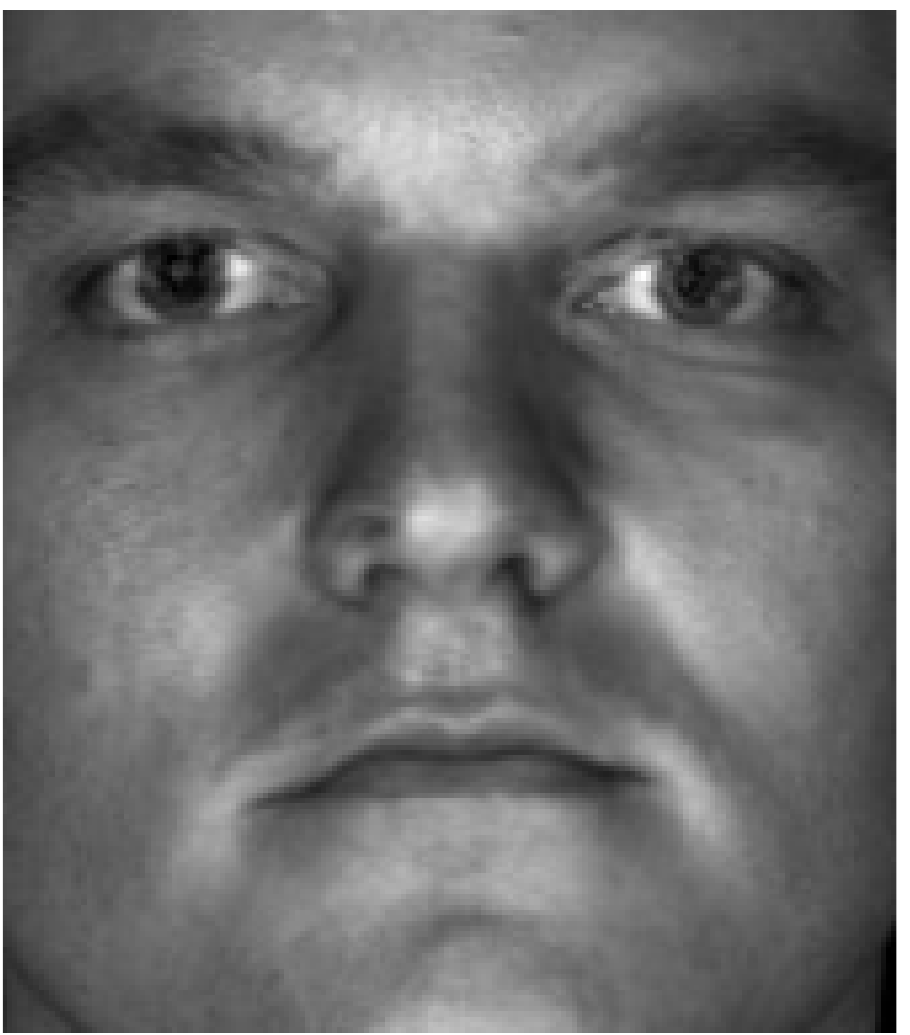} }
\subfloat{ \includegraphics[width=\tmpYaleFigSpace\textwidth]{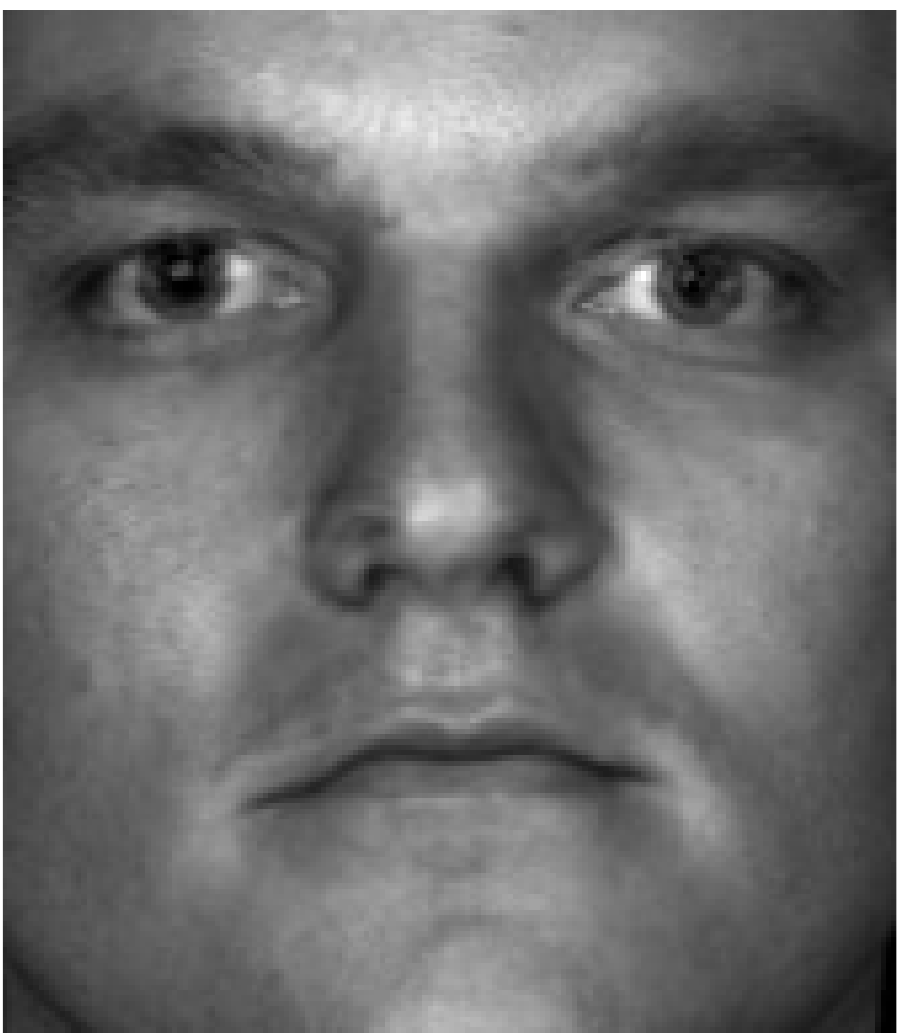} }
\subfloat{ \includegraphics[width=\tmpYaleFigSpace\textwidth]{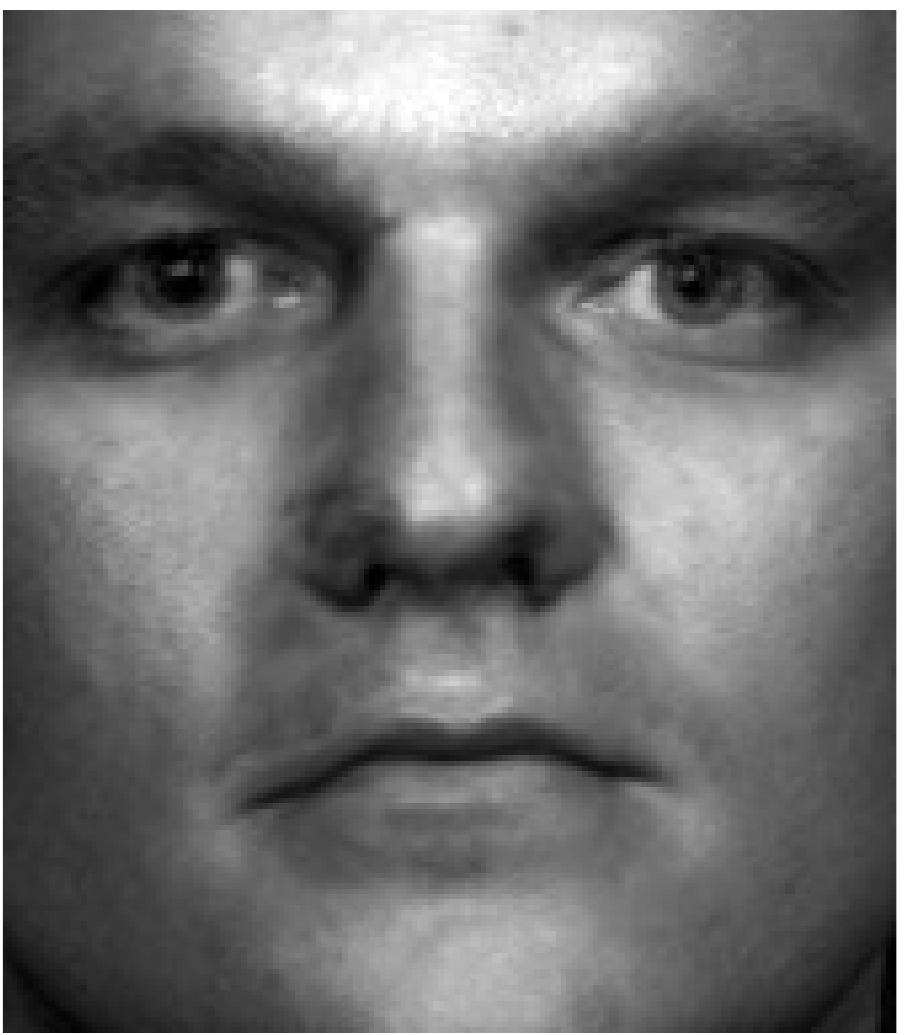} }
\subfloat{ \includegraphics[width=\tmpYaleFigSpace\textwidth]{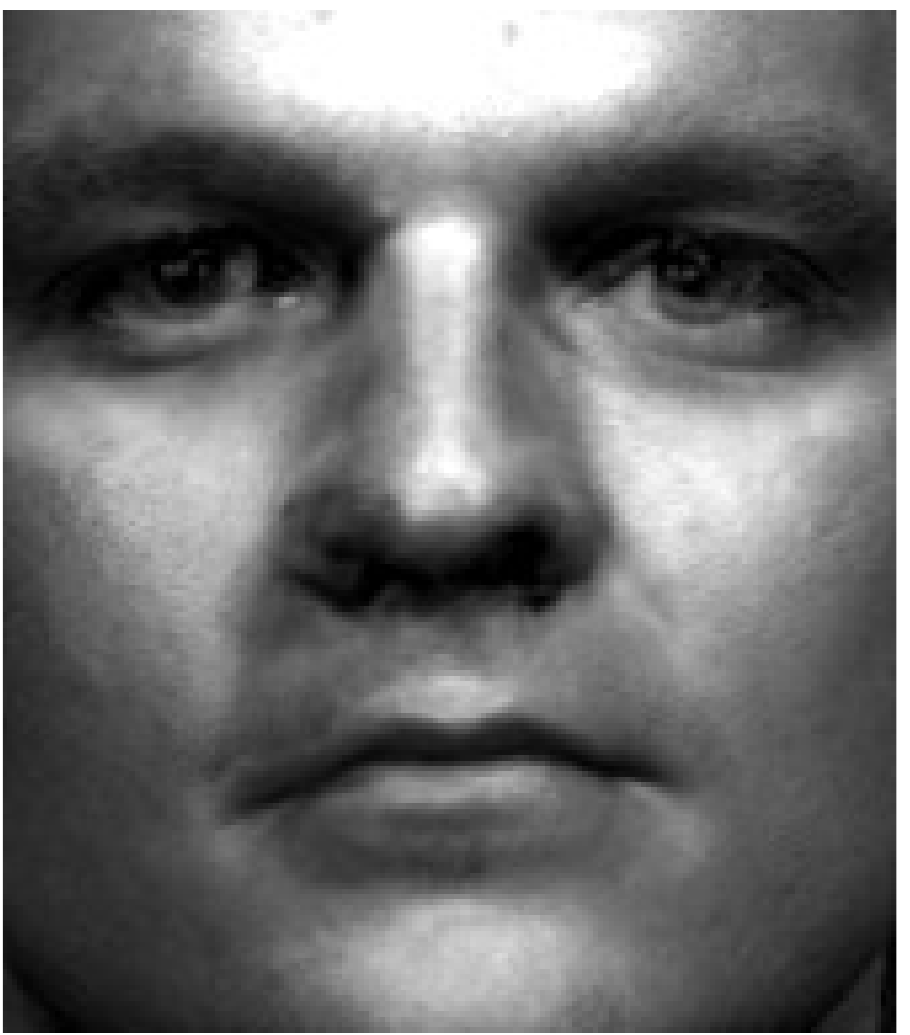} }
\subfloat{ \includegraphics[width=\tmpYaleFigSpace\textwidth]{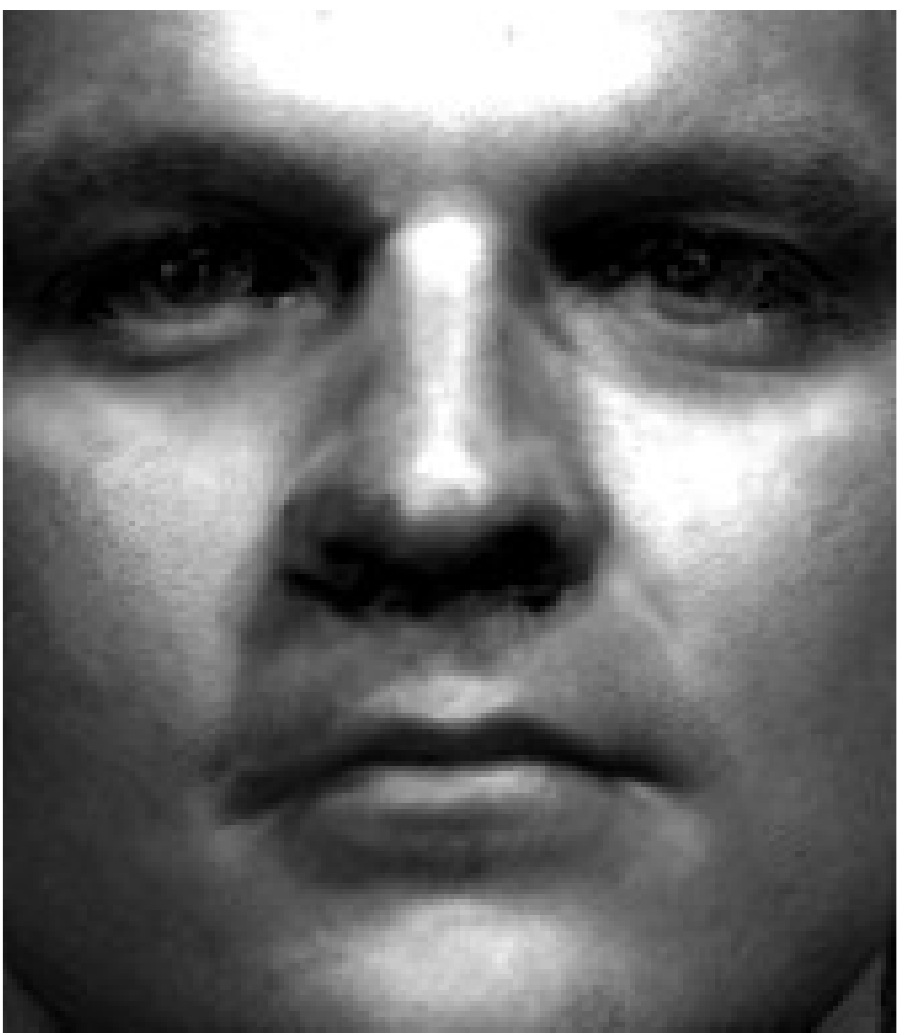} }
\subfloat{ \includegraphics[width=\tmpYaleFigSpace\textwidth]{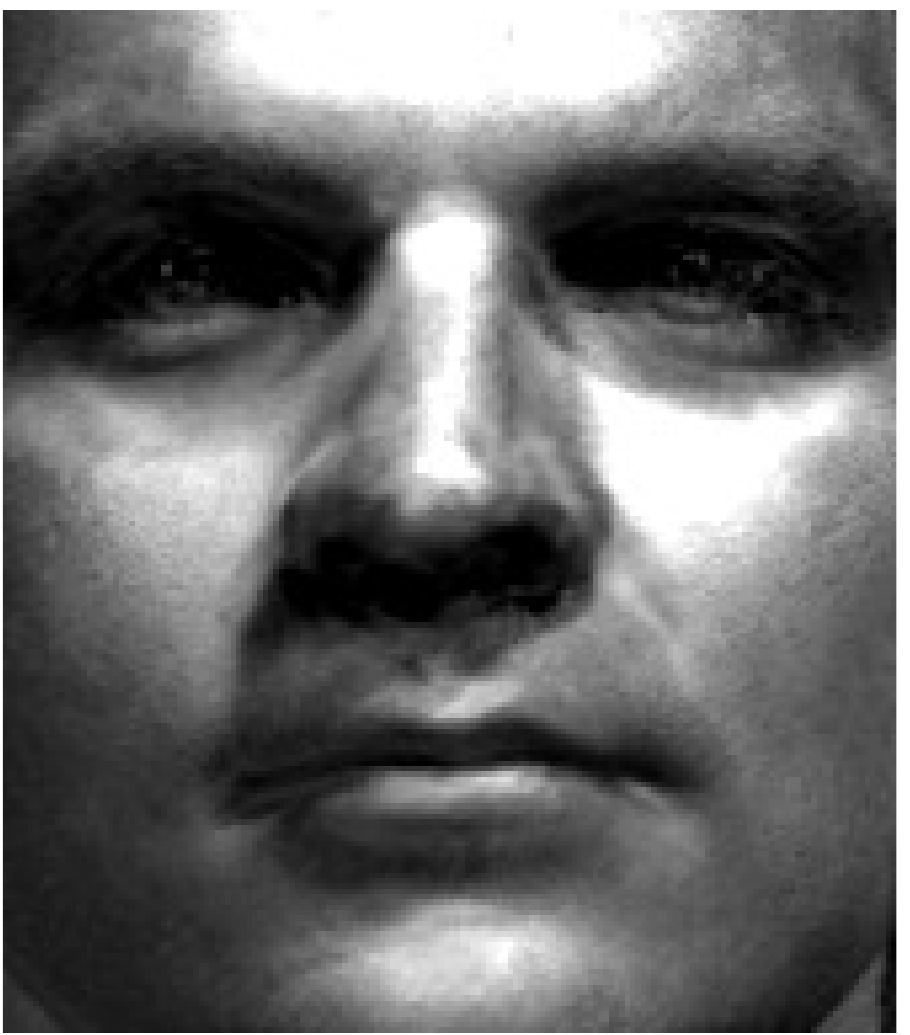} }
\vspace{-4pt}
\newline
\subfloat{ \includegraphics[width=0.1\textwidth]{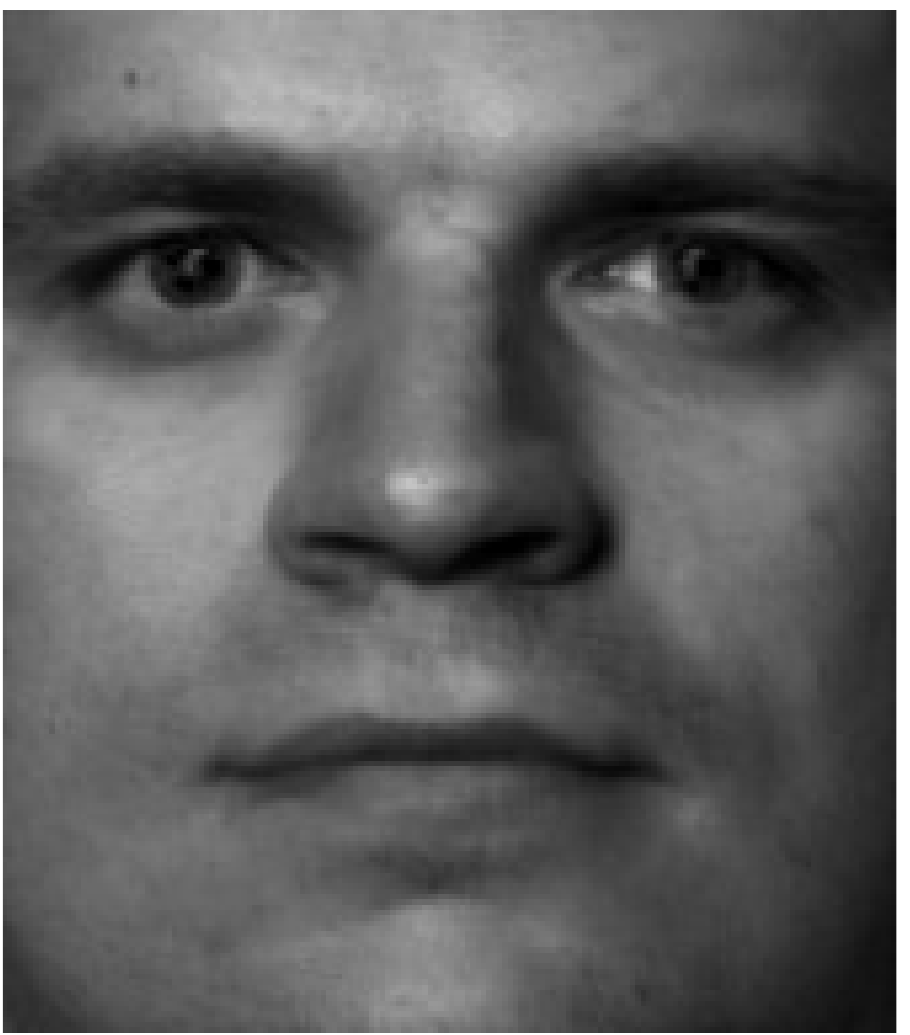} }
\subfloat{ \includegraphics[width=0.1\textwidth]{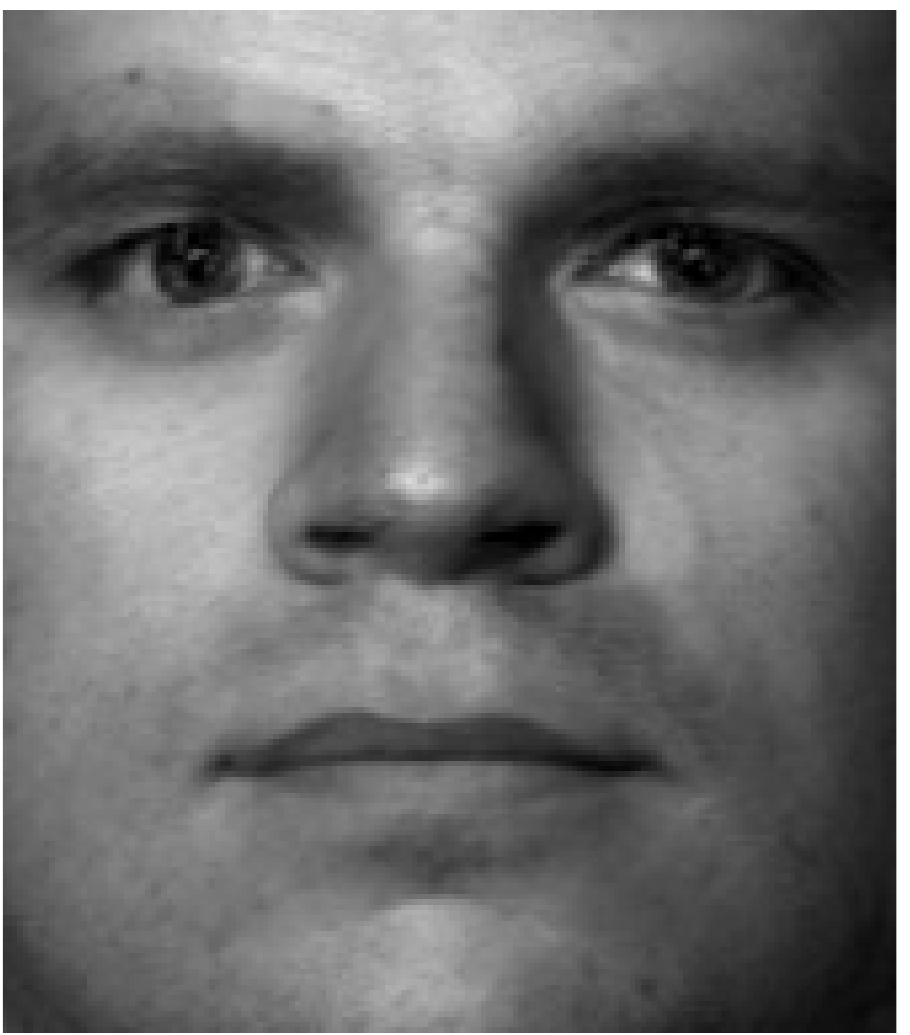} }
\subfloat{ \includegraphics[width=0.1\textwidth]{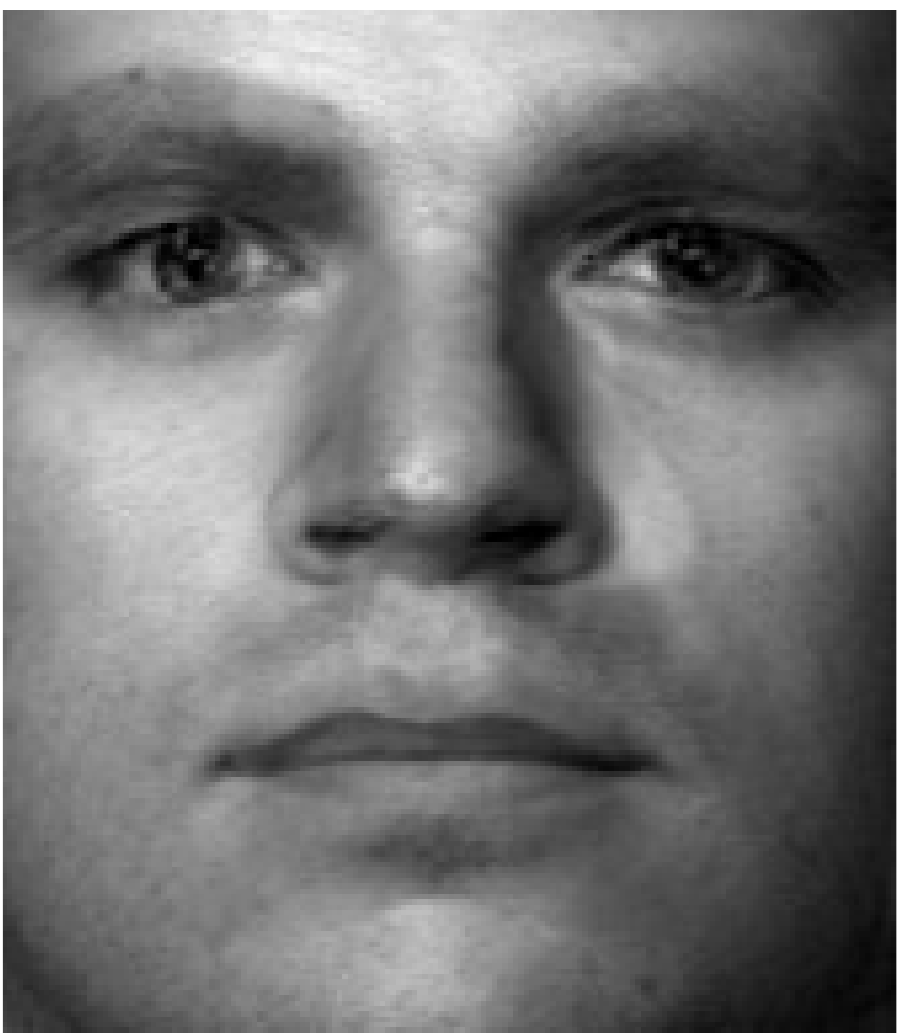} }
\subfloat{ \includegraphics[width=0.1\textwidth]{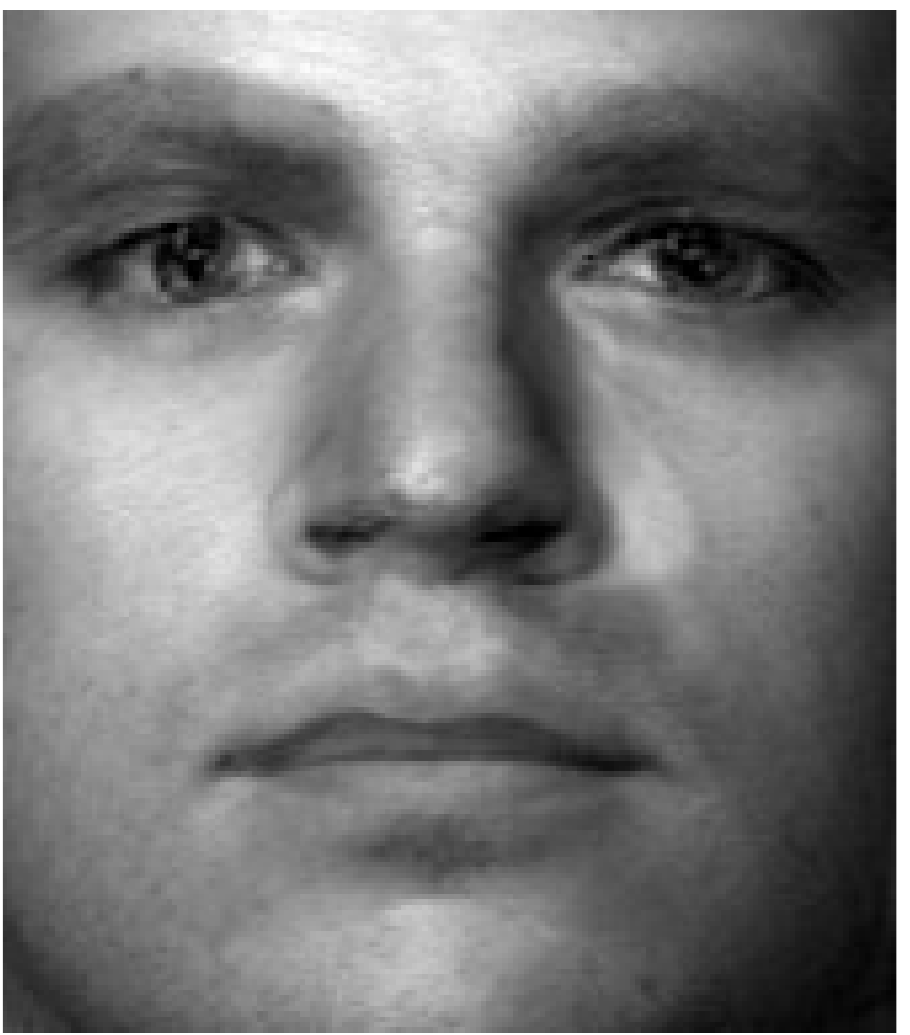} }
\subfloat{ \includegraphics[width=0.1\textwidth]{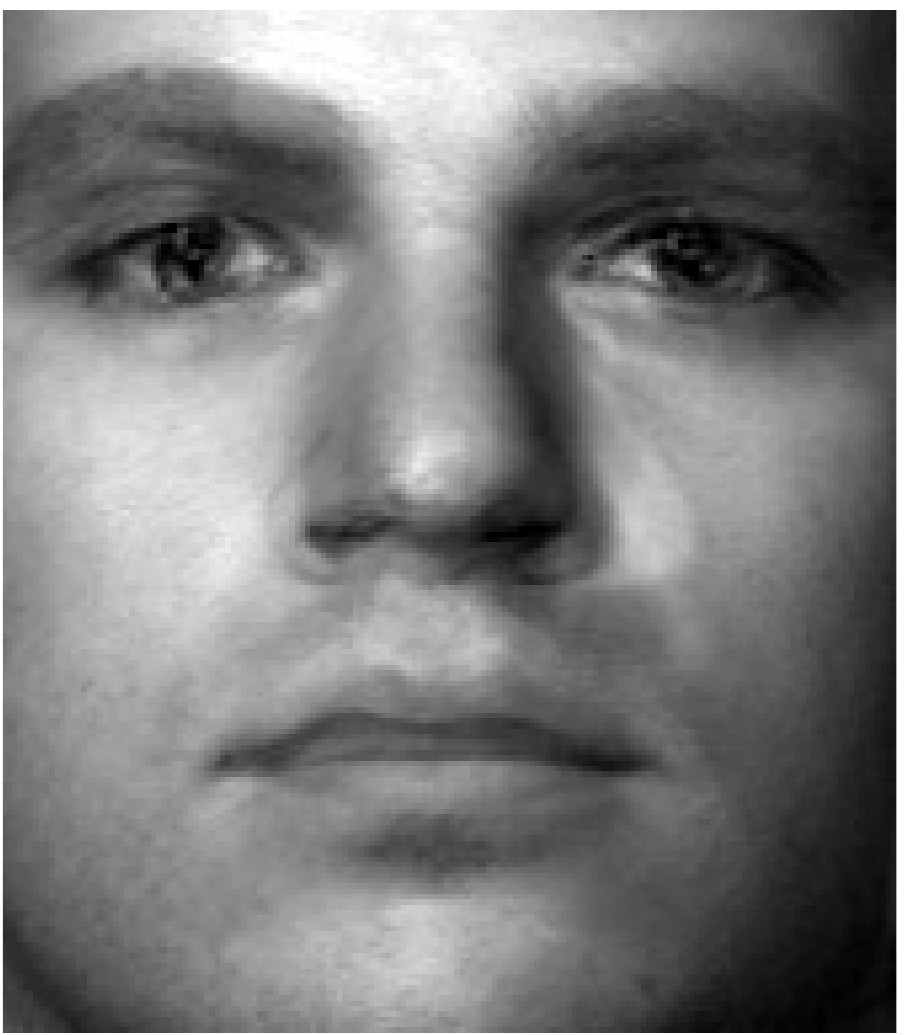} }
\subfloat{ \includegraphics[width=0.1\textwidth]{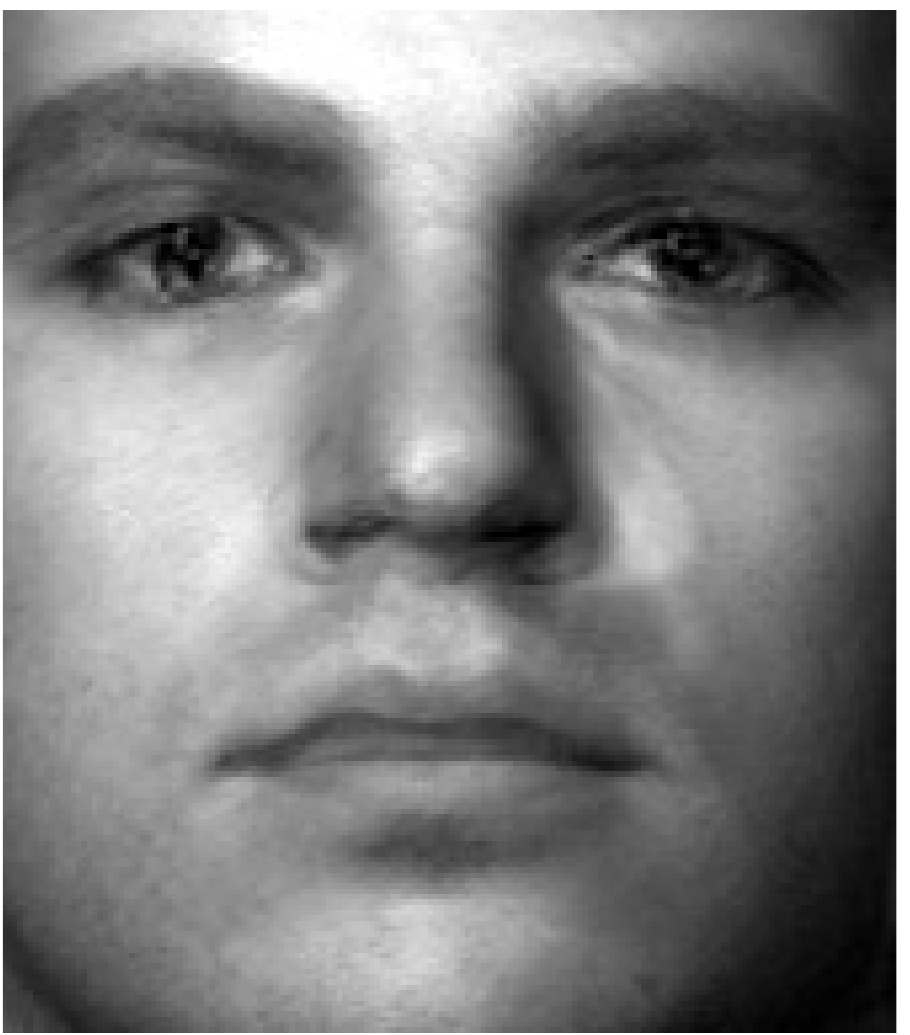} }
\subfloat{ \includegraphics[width=0.1\textwidth]{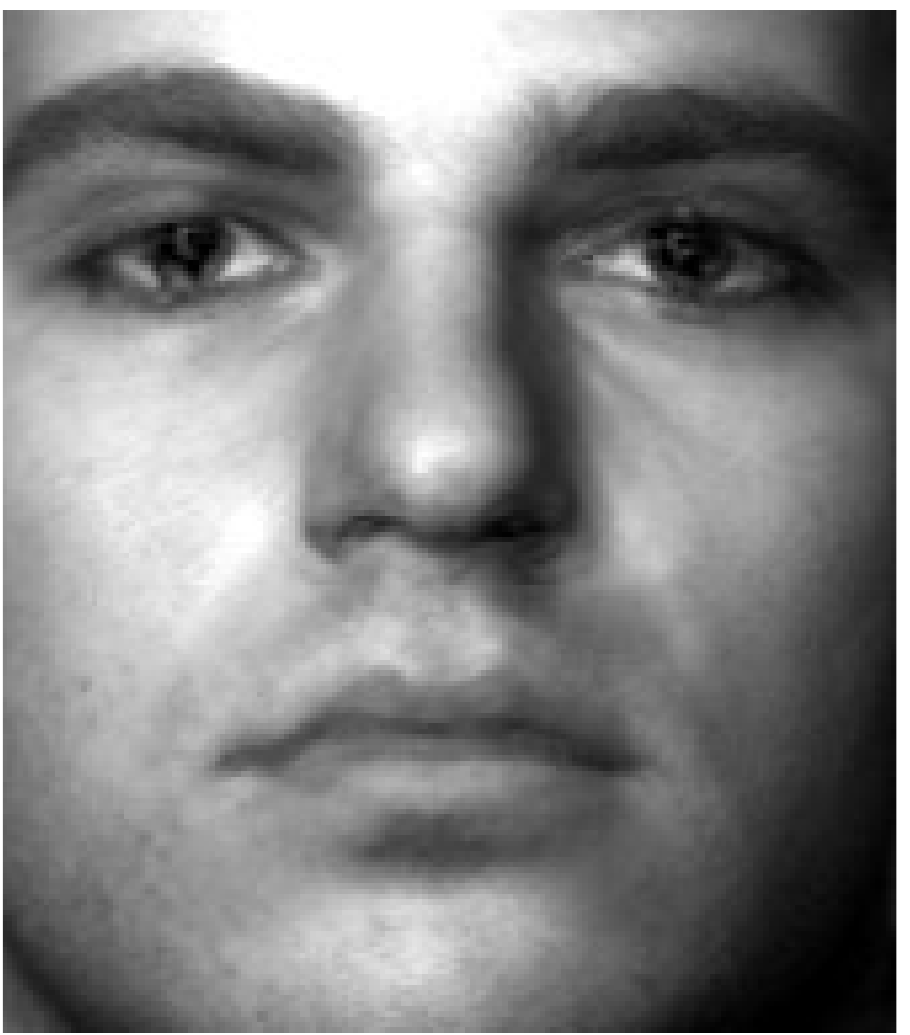} }
\end{center}
\caption[MRD, Yale Faces experiment: novel outputs obtained after structured sampling of the latent space.]{ Sampling inputs to produce novel outputs. First row shows interpolation between positions of the light source in the $x$ coordinate and second row in the $y$ coordinate (elevation), obtained by structured sampling in the shared latent space. Last row shows interpolation between face characteristics to produce a morphing effect, obtained by structured sampling in the private latent space. Note that these images are presented scaled here, the original size is $192 \times 198$ pixels.
}
\label{fig:yale6SetsInterpolation}
\end{figure}

We can also confirm visually the subspaces' properties by sampling a set of novel inputs $\mX_{\text{samp}}$ from each subspace and then mapping back to the observed data space using the likelihoods $p(\viewY|\mX_{\text{samp}})$ or $p(\viewZ|\mX_{\text{samp}})$, thus obtaining novel outputs (images).
To better understand what kind of information is encoded in each of the dimensions of the shared or private spaces, we sampled new latent points by varying only one dimension at a time, while keeping the rest fixed. The first two rows of \fig \ref{fig:yale6SetsInterpolation} show some of the outputs obtained after sampling across each of the shared dimensions $1$ and $3$ respectively, which clearly encode the coordinates of the light source, whereas dimension $2$ was found to model the overall brightness. The sampling procedure can intuitively be thought as a walk in the space shown in Figure \ref{fig:yale6SetsLatentSpace}\subref{fig:yale6SetsX13} from left to right and from the bottom to the top. Although the set of learned
latent inputs is discrete, the corresponding latent subspace is continuous, and we can interpolate images in new illumination conditions by sampling from areas where there are no training inputs (\ie in between the red crosses shown in Figure \ref{fig:yale6SetsLatentSpace}).
Similarly, we can sample from the private subspaces and obtain novel outputs which interpolate the non-shared characteristics of the involved data.  This results in a morphing effect across different faces, which is shown in the last row of \fig \ref{fig:yale6SetsInterpolation}. The two interpolation effects can be combined. Specifically, we can interactively obtain a set of shared dimensions corresponding to a specific lighting direction, and by fixing these dimensions we can then sample in the private dimensions, effectively obtaining interpolations between faces under the desired lighting condition. This demonstration, and the rest of the results, are illustrated in the online videos (\url{http://git.io/vwLhH}). As can be seen, MRD allows structured generation of novel high-dimensional outputs (images) by using low-dimensional inputs (latent points) as ``controls''.

\begin{figure}[t]
\begin{center}
\subfloat{ \includegraphics[width=1\textwidth]{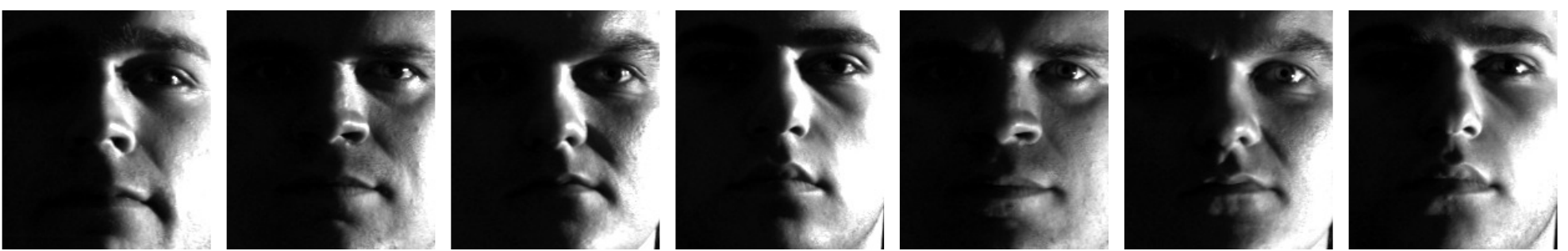} }
 \vspace{-6pt}
 \newline
\subfloat{ \includegraphics[width=1\textwidth]{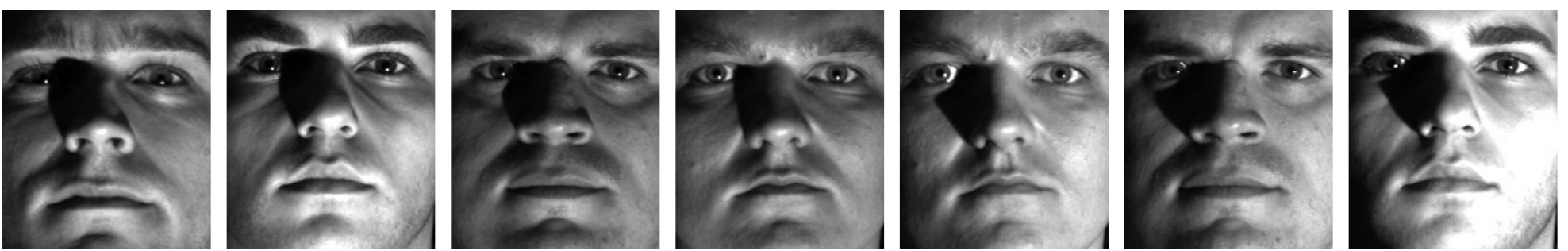} }
 \vspace{-6pt}
 \newline
\subfloat{ \includegraphics[width=1\textwidth]{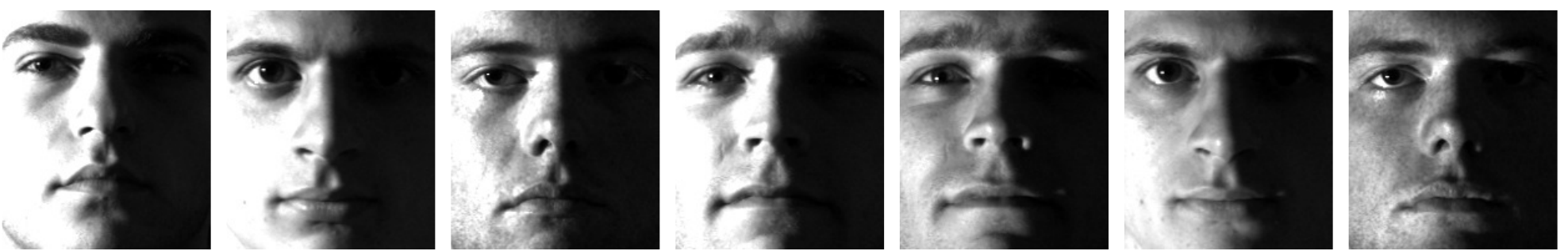} }
 \vspace{-6pt}
 \newline
\subfloat{ \includegraphics[width=1\textwidth]{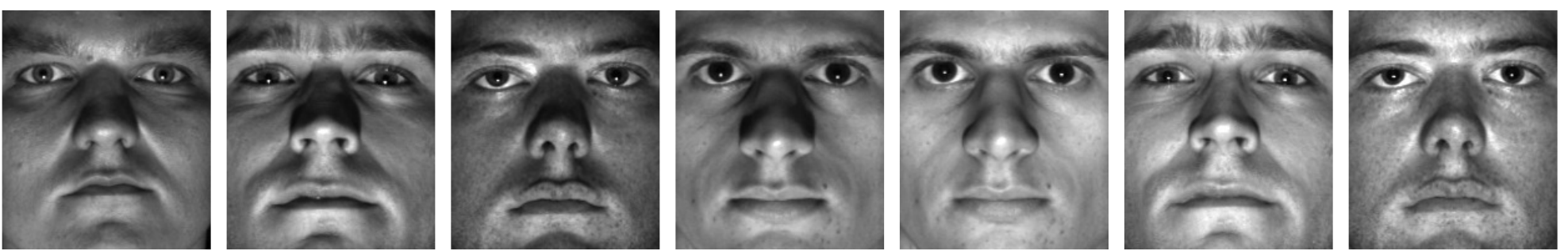} }
 \vspace{-6pt}
 \newline
\end{center}
\caption[MRD, Yale Faces experiment: correspondence problem solution.]{
Solving the correspondence problem: given the images of the first column, the model searches only in the shared latent space to find the pictures of the opposite view which have the same illumination condition. The images found, are sorted in columns $2$ - $7$ by relevance.
}
\label{fig:yale6SetsGrouping}
\end{figure}

As a final test, we confirm the efficient factorization of the latent space into private and shared parts by automatically recovering all the illumination similarities found in the training set.  More specifically, given a data point $\yn^\viewYi$ from the first view, we search the whole space of training inputs $\mX$ to find the $6$ nearest neigbours to the latent representation $\xn$ of $\yn^\viewYi$, \emph{based only on the shared dimensions}.
 From these latent points, we can then obtain points in the output space of the second view, by using the likelihood $p(\viewZ | \mX)$. This procedure is a special case of Algorithm \ref{algorithm:MRD_inference} where the test point given is already in the training set.  As can be seen in Figure \ref{fig:yale6SetsGrouping}, the model returns images with matching illumination condition.  Moreover, the fact that, typically, the first neighbours of each given point correspond to outputs belonging to different faces, indicates that the shared latent space is ``pure'', and is not polluted by information that encodes the face appearance.

\subsection{Pose Estimation and Ambiguity Modelling\label{sec:MRD_experiment_poseSilhouette}}

\begin{figure}[t]
\begin{center}
\includegraphics[width=0.5\textwidth]{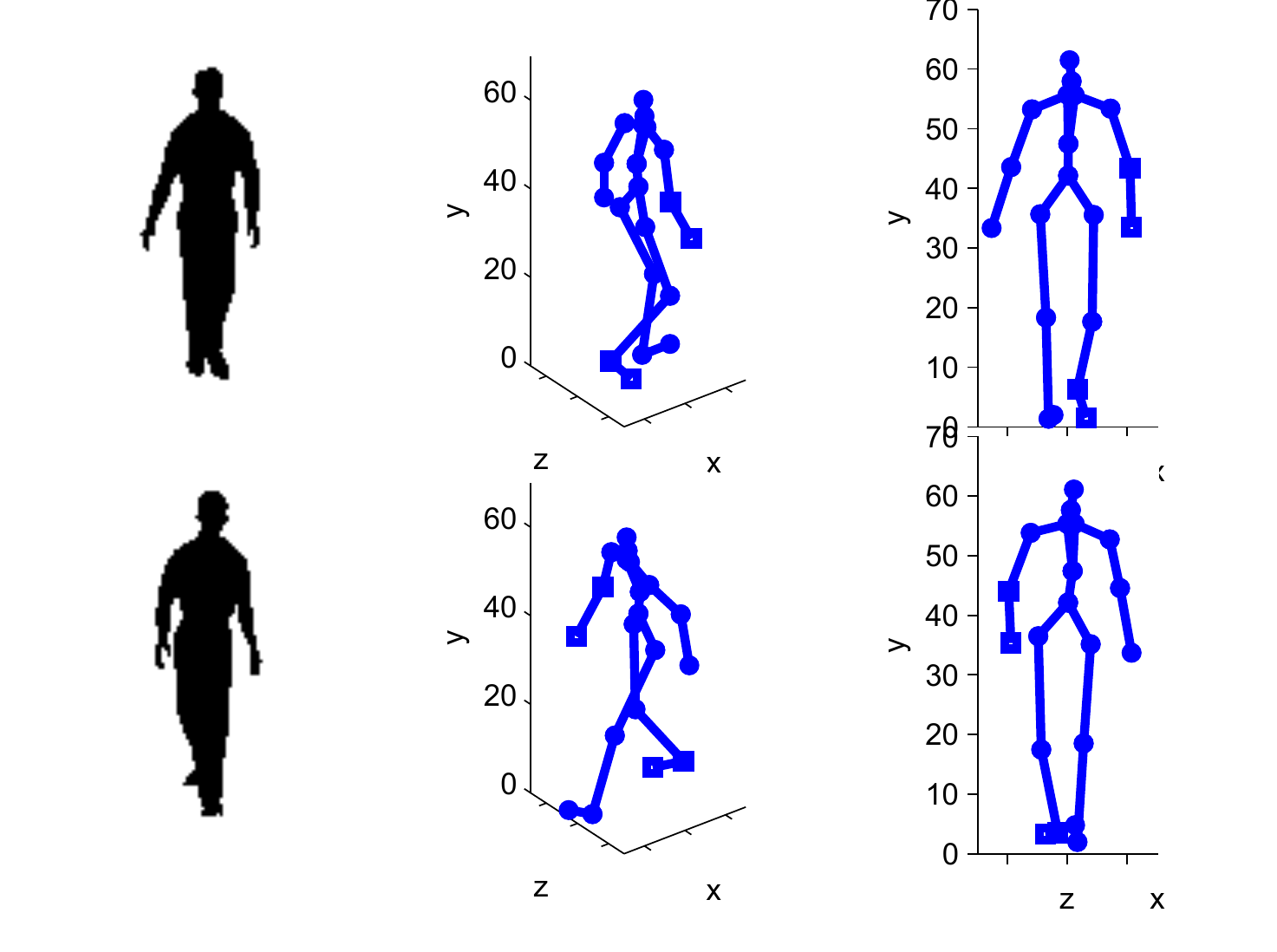}
\end{center}
\caption[MRD, human pose experiment: ambiguity example.]{
Although the two poses in the second column are very dissimilar, they correspond to resembling silhouettes
that have similar feature vectors. This happens because the $3$D information is lost in the silhouette space,
as can also be seen in the third column, depicting the same poses from the silhouettes' viewpoint.
}
\label{fig:humanPoseAmbiguity}
\end{figure}

For our next experiment we will use the MRD model to perform human pose estimation from silhouette data. The purpose of this experiment is to show how a factorized latent variable model can be used to perform efficient inference when the task is ambiguous.  We consider a set of $3$D human poses and associated silhouettes, coming from the data set of \cite{Agarwal:2006hh}. We used a subset of $5$ sequences, totaling $649$ frames, corresponding to walking motions in various directions and patterns.  A separate walking sequence of $158$ frames was used as a test set.  Each pose is represented by a $63-$dimensional vector of joint locations and each silhouette is represented by a $100-$dimensional vector of HoG (histogram of oriented gradients) features. Given the test silhouette features $\{ \viewynTs \}_{\jn=1}^{\nN\Ts}$, we used our model to generate the corresponding poses $\{ \viewznTs \}_{\jn=1}^{\nN\Ts}$. This is challenging, as the data are multi-modal and ambiguous, \ie a silhouette representation may be generated from more than one pose (\eg \fig \ref{fig:humanPoseAmbiguity}).

The inference procedure proceeds as described in Algorithm \ref{algorithm:MRD_inference}. Specifically, given a test point $\vy\nTs^\viewYi$ we firstly estimate the corresponding latent point $\vx\nTs$ and then through a nearest neighbour search we seek the training latent point $\tilde{\vx}$ which is closest to $\vx\nTs$ in the shared dimensions. 
It is interesting to investigate which latent points $\tilde{\vx}$ are returned by the nearest neighbour search, as this will reveal properties of the shared latent space. While exploring this aspect, we found that the training points suggested as most similar in the shared space typically correspond to silhouettes (outputs) similar to the given test one, $\viewynTs$. This confirms that the factorization of the latent space is efficient in representing the correct kind of information in each subspace. However, when ambiguities arise, as the example shown in Figure \ref{fig:humanPoseAmbiguity}, the non-dynamical version of our model has no way of selecting the correct input, since all points of the test sequence are treated independently. Intuitively, this means that two very similar test givens $(\viewynTs, \viewyinnTs{\jn+1})$ can be mapped to generated latent vectors $(\tilde{\vx}\nTs, \tilde{\vx}_{(\jn+1),*})$ from which predictions $(\viewzinnTs{\jn}, \viewzinnTs{\jn+1})$ in the other modality can be drastically different.
 But when the dynamical version is employed, the model forces the whole set of training and test inputs (and, therefore, also the test outputs) to form smooth paths.
 In other words, the dynamics disambiguate the model.

\begin{figure}[h]
\begin{center}
  \includegraphics[width=0.55\textwidth]{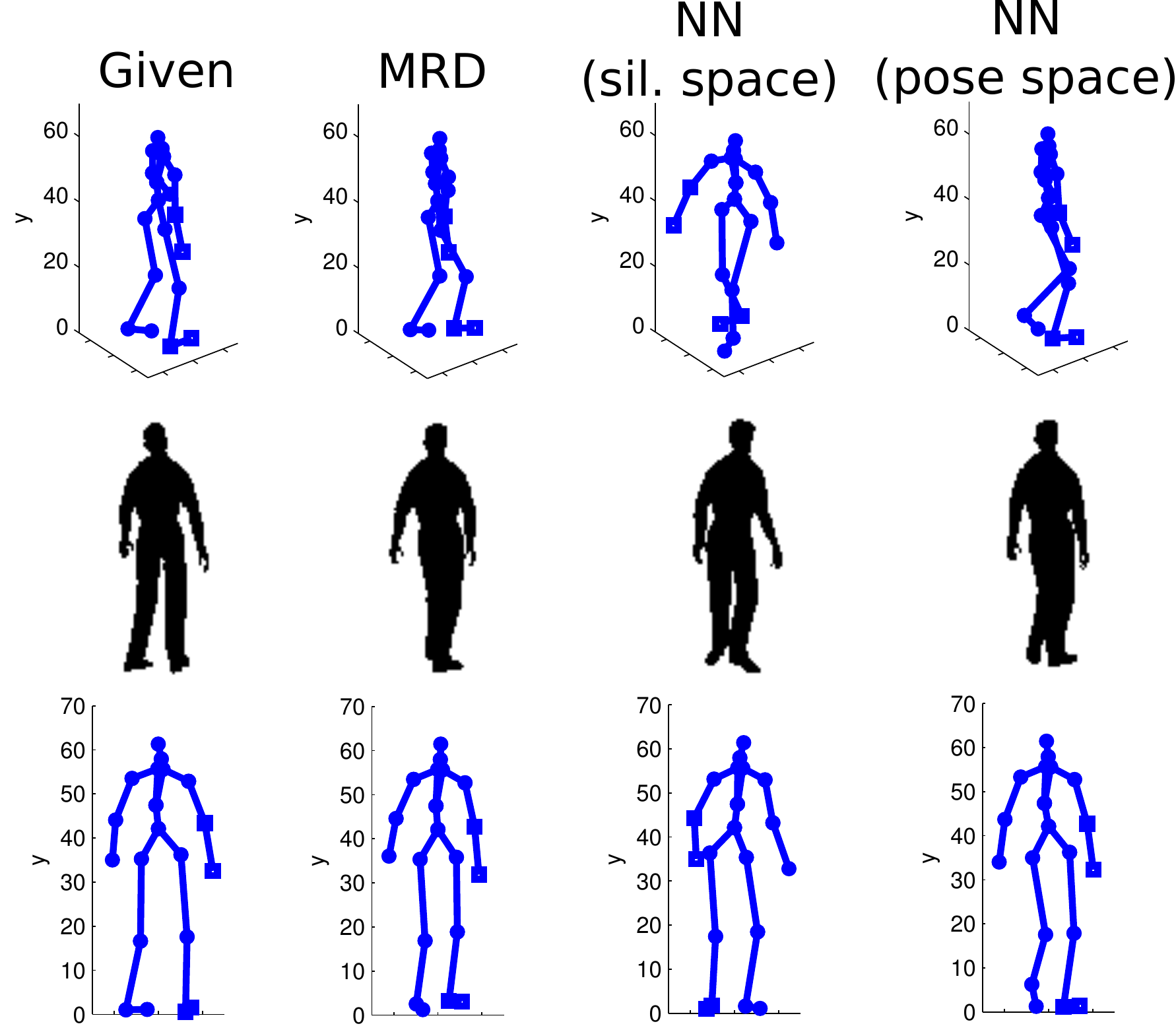}
\end{center}
\caption[MRD, human pose experiment: predictions examples.]{
Given the HoG features $\viewynTs$ for the test silhouette in column one, we predict the corresponding pose $\viewznTs$ using the dynamical MRD and nearest neighbour (NN) in the silhouette space
obtaining the results in the first row, columns 2 and 3 respectively. The last row is the same as the first one, but the poses are rotated to highlight the ambiguities.
 Notice that the silhouette shown in the second row for MRD does not correspond exactly to the pose
of the first row, as the model generates a \emph{novel} pose given a test silhouette. Instead, it is the training silhouette found by performing NN in the shared latent space to obtain $\tilde{\vx}\n$.
As some form of ``ground truth'', in column $4$ we plot the NN of the training \emph{pose} $\viewy\n$ given the test pose $\viewznTs$ (which is normally unknown during the test phase).
}
\label{fig:humanPoseAmbiguityTest}
\end{figure}

Therefore, in the dynamical scenario the given test silhouette $\viewynTs$ is accompanied by its timestamp, $\bft\nTs$, which is used to disambiguate the temporal approximate posterior $q(\vx\nTs | \bft\nTs)$ and, consequently, the prediction $\viewznTs$. 
This temporal disambiguation effect is demonstrated in \fig \ref{fig:humanPoseAmbiguityTest}, where from a set of test silhouettes $\viewY\Ts$ we find the corresponding set of nearest training silhouettes $\widetilde{\bfY}^{\mathcal{\viewYi}}$ through the shared latent space which respects dynamics. In this case, we see that each $\viewynTs$ is not necessarily the most similar to the corresponding $\tilde{\bfy}^\mathcal{\viewYi}\n$, because that would mean that the dynamics would ``break'', as in \fig \ref{fig:humanPoseAmbiguityTest}, column 1 versus column 3 of row 2. Instead, our model treats the whole test set as a sequence, so in column 2 of \fig \ref{fig:humanPoseAmbiguityTest} we see that the silhouette is more dissimilar to the given one (column 1) but it represents a walk in the same direction. 
What is more, if we assume that the test \emph{pose} $\viewznTs$ is  known and look for its nearest training neighbour \emph{in the pose space}, we find that the corresponding silhouette is very similar to the one found by our model, which is only given information in the silhouette space. 

After the above analysis regarding the properties of the latent space, we now proceed to quantitatively evaluate the generation of test poses $\viewzTs$ from test silhouettes $\viewy\Ts$. \fig \ref{fig:humanPoseAmbiguity} shows one encouraging example of this result. To more reliably quantify the results, we compare our method
with linear and Gaussian process regression and with nearest neighbour in
the silhouette space. We also compared against the shared GP-LVM
\citep{Ek:2009vv} which optimizes the latent points using
MAP and, thus, requires an initial factorization of the inputs to
be given a priori.  Finally, we compared to a dynamical version of nearest neighbour where
we kept multiple nearest neighbours and selected the coherent ones over a sequence. 
The errors shown in table \ref{tab:humanMotionTable}
as well as the on-line videos (\url{http://git.io/vwLhH}) show that MRD
performs better than the other methods in this task.

\begin{table}[h]
\begin{small}
\begin{center}
\begin{tabular}{ l | l }
                               & Error \\ \hline 
Mean Training Pose             & 6.16   \\ 
Linear Regression              & 5.86   \\ 
GP Regression                  & 4.27   \\ 
Nearest Neighbour (sil. space) & 4.88  \\ 
Nearest Neighbour with sequences (sil. space) & 4.04  \\ 
\textcolor{light-gray}{Nearest Neighbour (pose space)} & \textcolor{light-gray}{2.08}   \\ 
Shared GP-LVM                      & 5.13    \\ 
MRD without Dynamics       & 4.67   \\ 
MRD with Dynamics          & \textbf{2.94}    \\ \hline
\end{tabular}
\end{center}
\end{small}
\caption[MRD, human pose experiment: MSE and comparisons.]{
The mean of the Euclidean distances of the joint locations between the predicted and the true poses. The nearest neighbour in the pose space is not a fair comparison (since the test pose is supposed to be unseen), but is reported as it provides some insight about the lower bound on the error that can be achieved for this task.
}
\label{tab:humanMotionTable}
\end{table}

\subsection{Discriminative - Generative MRD\label{sec:oilData}}
So far the experiments have considered two continuous views and the task was either to generate novel data or to transfer information between the views. We now consider a hybrid discriminative - generative model and task, where one view contains labels of the features in the other view. This experimental setting is quite different from the ones considered so far, since the two views contain very diverse types of data; in particular, the class-label view contains discrete, low dimensional features. Further, these features are noise-free and very informative for the task at hand and, therefore, applying MRD in this data set can be seen as a form of supervised dimensionality reduction. The challenge for the model is to successfully cope with the different levels of noise in the views, while managing to recover a continuous shared latent space from two very diverse information sources, one of which is discriminative. In particular, we would ideally expect to obtain a shared latent space which encodes class information and a nonexistent private space for the class-label modality.

\begin{figure}[ht]
\begin{center}
  \includegraphics[width=0.67\textwidth]{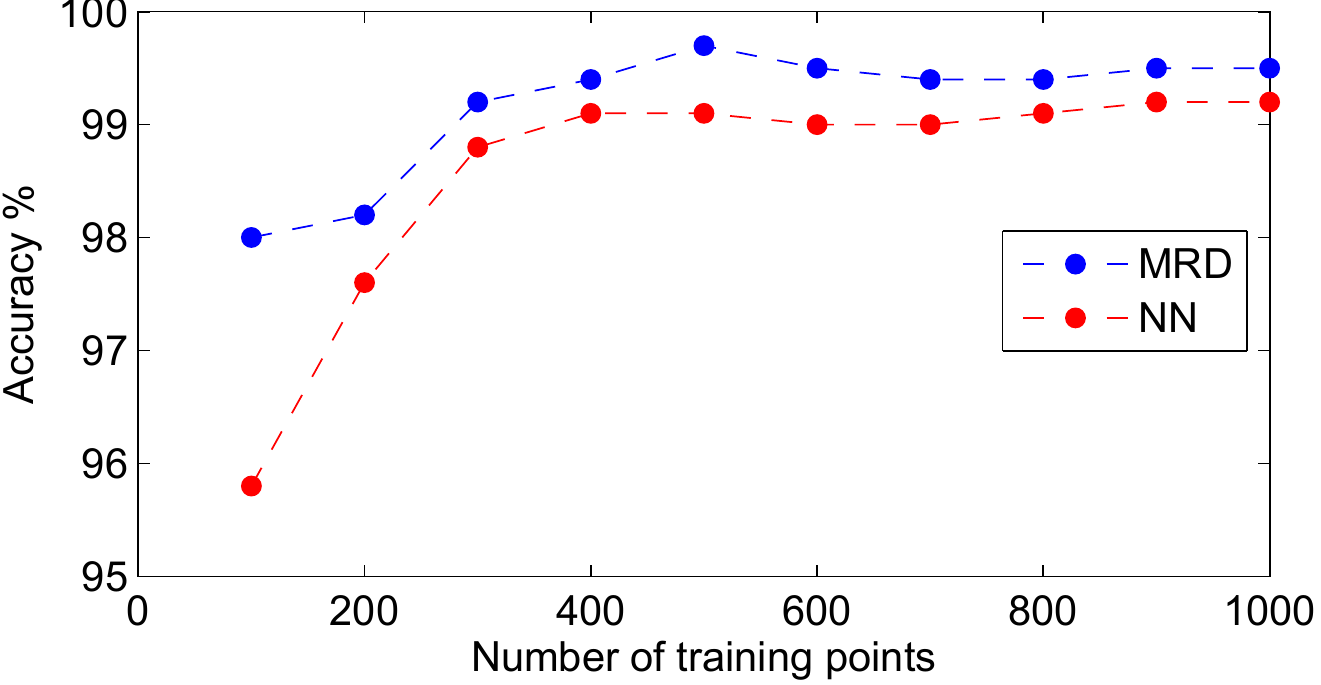}
\end{center}
\caption[Accuracy obtained after using MRD as a generative classifier.]{
Accuracy obtained after testing MRD and NN on the full test set of the ``oil flow'' data set.
}
\label{fig:oilErrors}
\end{figure}

To test our hypotheses, we used the ``oil flow'' database \citep{Bishop:oil93} which contains $1000$ $12-$dimensional examples split in $3$ classes.  We selected $10$ random subsets of the data with increasing number of training examples and compared to the nearest neighbor (NN) method in the data space. The label $\viewz\n = [\viewzs\ind{\jn}{1}, \viewzs\ind{\jn}{2}, \viewzs\ind{\jn}{3}]^\top$ corresponding to the training instance $\viewy\n$ was encoded so that $\viewzs\nd = -1$ if $\viewy\n$ does not belong to class $\jd$, and $\viewzs\nd=1$ otherwise. Given a test instance $\viewynTs$, we predict the corresponding label vector $\viewznTs$ as before. Since this vector contains continuous values, we use $0$ as a threshold to obtain values $\viewzs\nd \in \{-1,1\}$. With this technique, we can perform multi-class and multi-label classification, where an instance can belong to more than one classes. The specific data set considered in this section, however, is not multi-label. To evaluate this technique, we computed the classification accuracy as a proportion of correctly classified instances. As can be seen in Figure \ref{fig:oilErrors}, MRD successfully determines the shared information between the data and the label space and outperforms NN. This result suggests that MRD manages to factor out the non class-specific information found in $\viewY$ and perform classification based on more informative features (\ie the shared latent space).

\begin{figure}[ht]
\begin{center}
\subfloat[Relevance weights]{
  \includegraphics[width=0.465\textwidth]{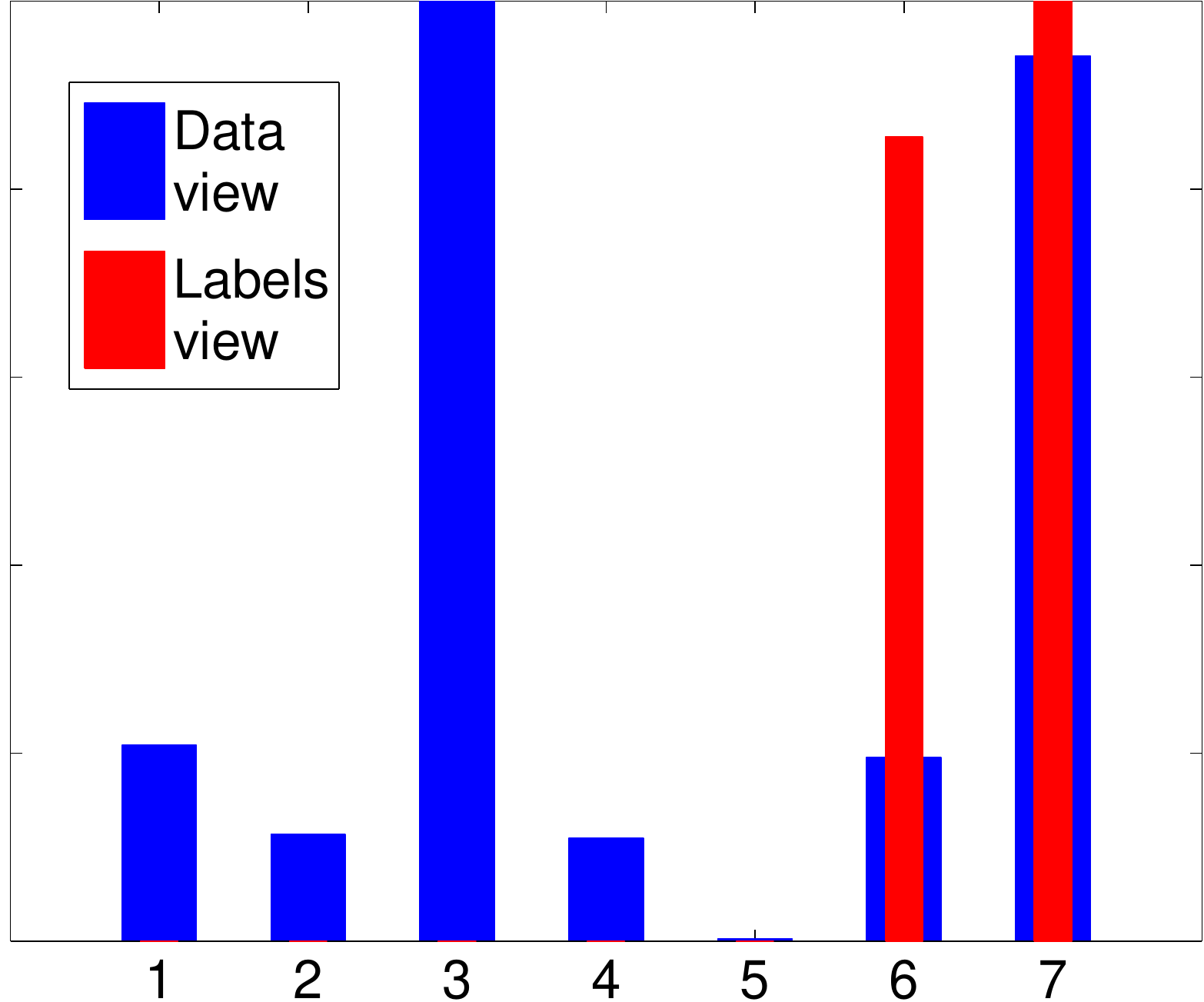}
  }
  \hfill
\subfloat[Shared latent space.]{
  \includegraphics[width=0.50\textwidth]{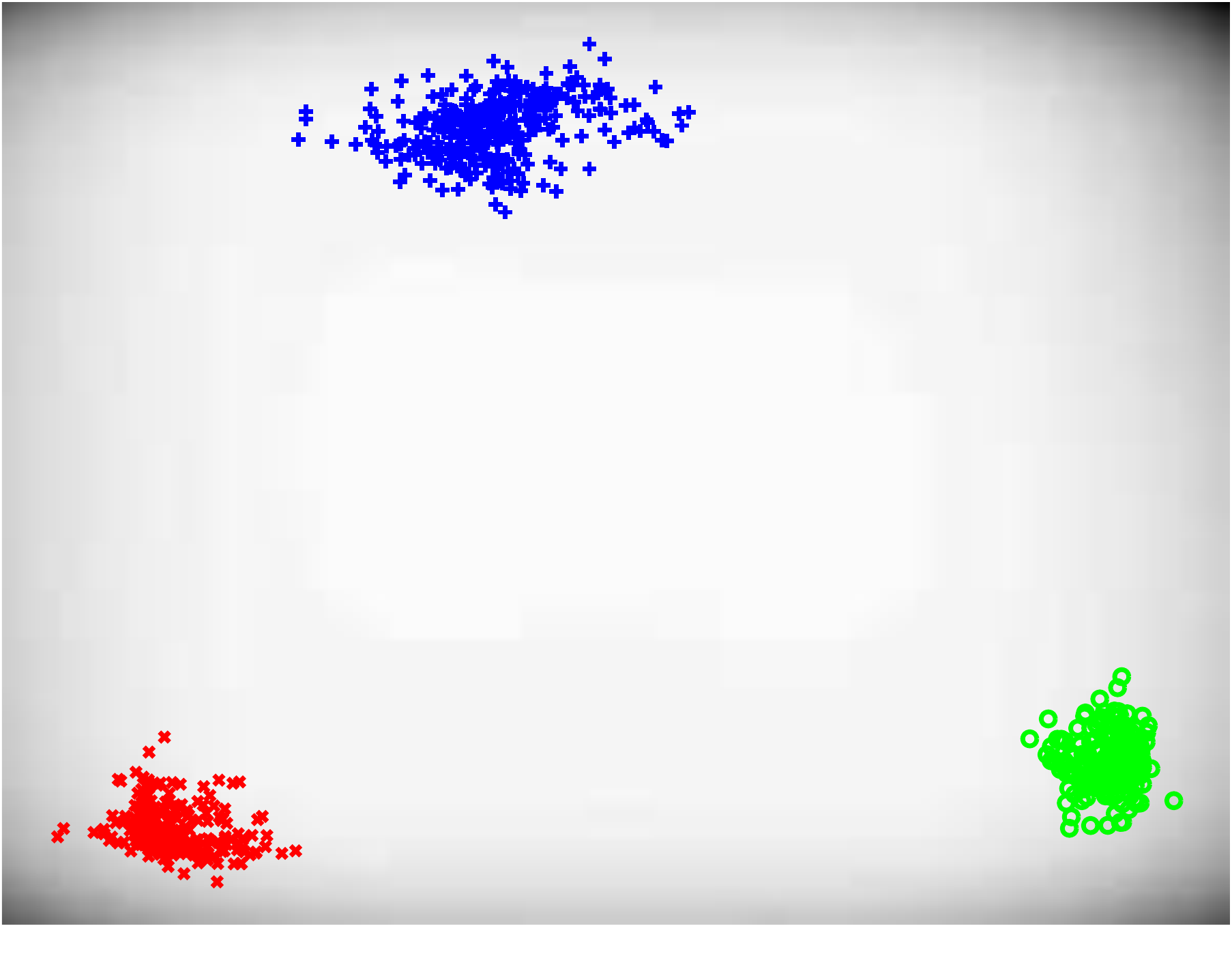}
}
\end{center}
\caption[Latent space projection and weights for the MRD classification experiment.]{
Results from testing the MRD as a classifier on the full ``oil flow'' data set, where one of the views was the data and one the class labels. Three important observations can be made: firstly, both views ``agree'' on using dimensions 6 and 7 (and the data view even switches off one other dimension). Secondly, the labels' view has no private space. Thirdly, the shared latent space projection clearly clusters the data according to their class label.
}
\label{fig:oilLatent}
\end{figure}

It is worth mentioning that, as expected, the models trained in each experimental trial defined a latent space factorization where there is no private space for the label view, whereas the shared space is one or two dimensional and is composed of three clusters (corresponding to the three distinct labels). Therefore, by factorizing out signal in $\viewY$ that is irrelevant to the classification task, we manage to obtain a better classification accuracy. The above are confirmed in \fig \ref{fig:oilLatent}, where we plot the shared latent space and the relevance weights for the model trained on the full data set.

\subsection{Multi-view Models and Data Exploration}
\begin{figure}[ht]
\begin{center}
  \includegraphics[width=0.2385\textwidth]{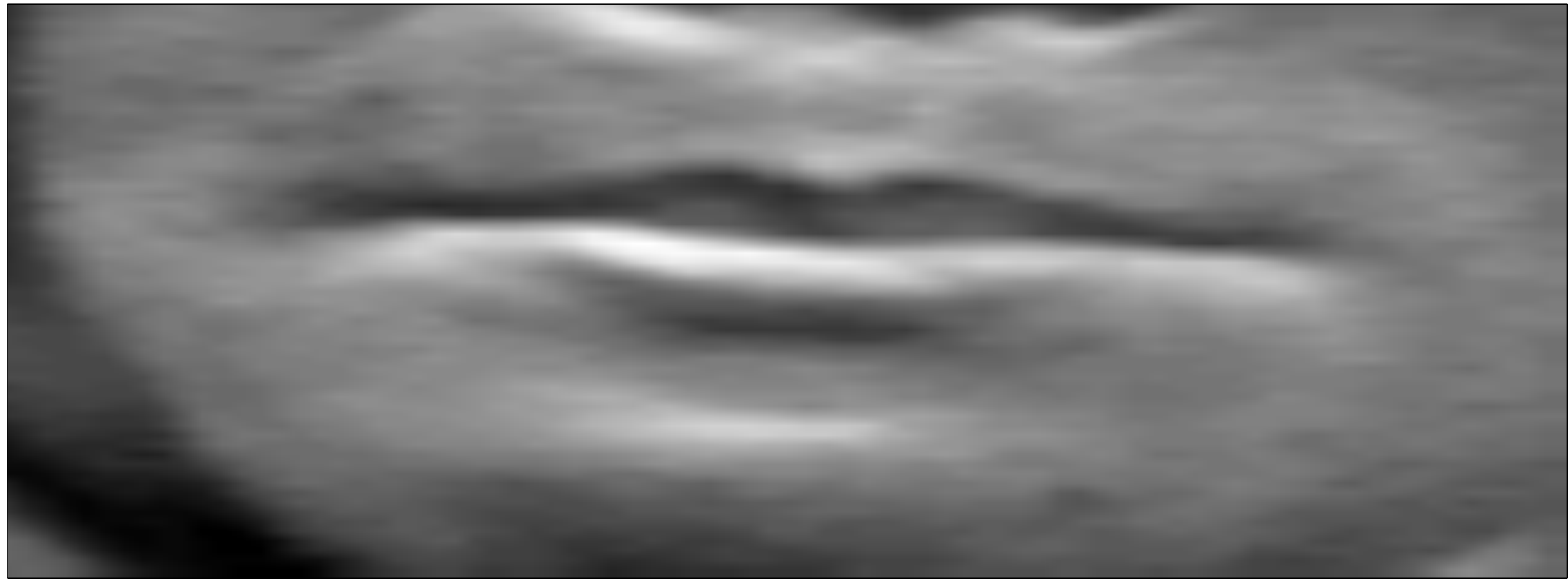}
  \includegraphics[width=0.242\textwidth]{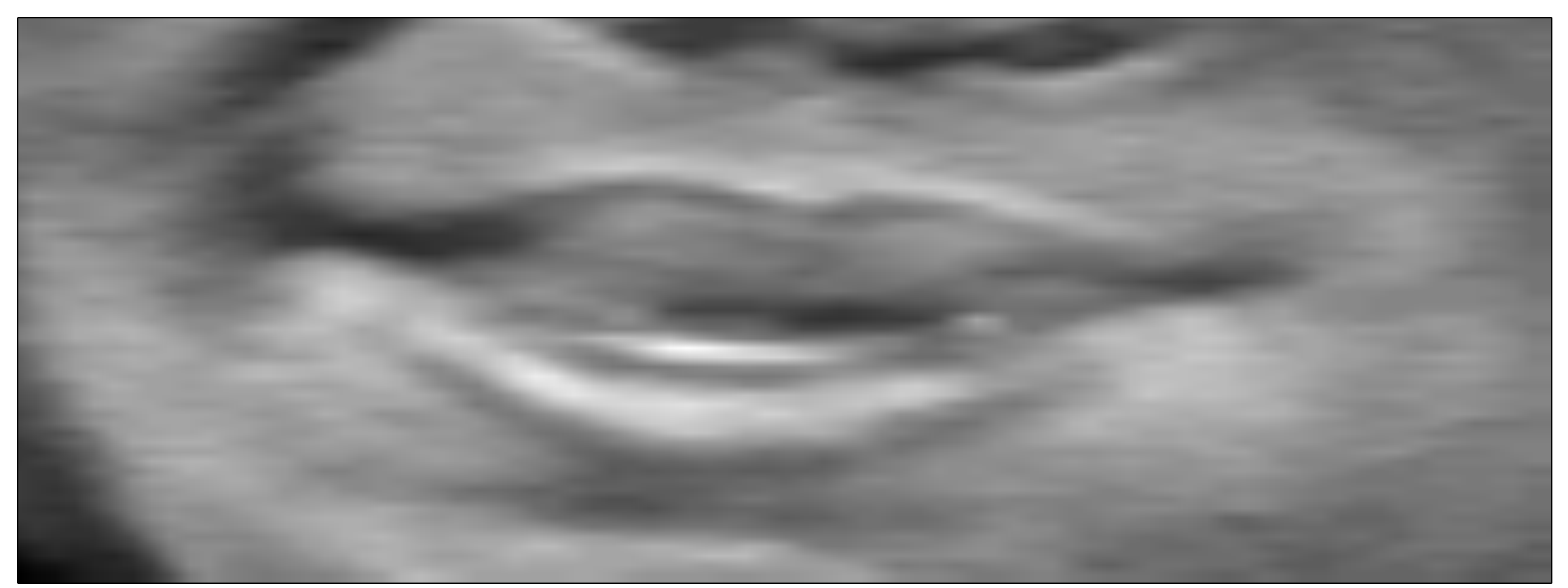}
\end{center}
\vspace{-0.1cm}
\caption{
Two example frames of the AVletters data set.
}
\label{fig:lipsExample}
\end{figure}

We have so far demonstrated MRD in data sets with two modalities. However, there is no theoretical constraint on the number of modalities that can be handled by MRD, it naturally extends beyond two views. Even when multiple views of scarce data are considered, the principled Bayesian framework will provide strong regularization. This is one of the important and powerful aspects of MRD compared to previous work. In this section we will use the AVletters database \citep{Matthews:2002ec} to generate multiple views of data. This audio-visual data set was generated by recording the audio and visual signals of $10$ speakers that uttered the letters A to Z three times each (\ie three \emph{trials}). The audio signal was processed to obtain a $299-$ dimensional vector per utterance. The video signal per utterance is a sequence of $24$  frames, each being represented by the raw values of the $60 \times 80$ pixels around the lips, as can be seen in Figure \ref{fig:lipsExample}. Thus, a single instance of the video modality of this data set is a $115200-$dimensional vector. With different formulation of these data into views we construct three different scenarios, detailed in the following, in order to demonstrate MRD with large number of views.

\subsubsection*{Data Exploration}
\begin{figure}[t]
  \centering
  \includegraphics[height=0.24\textheight]{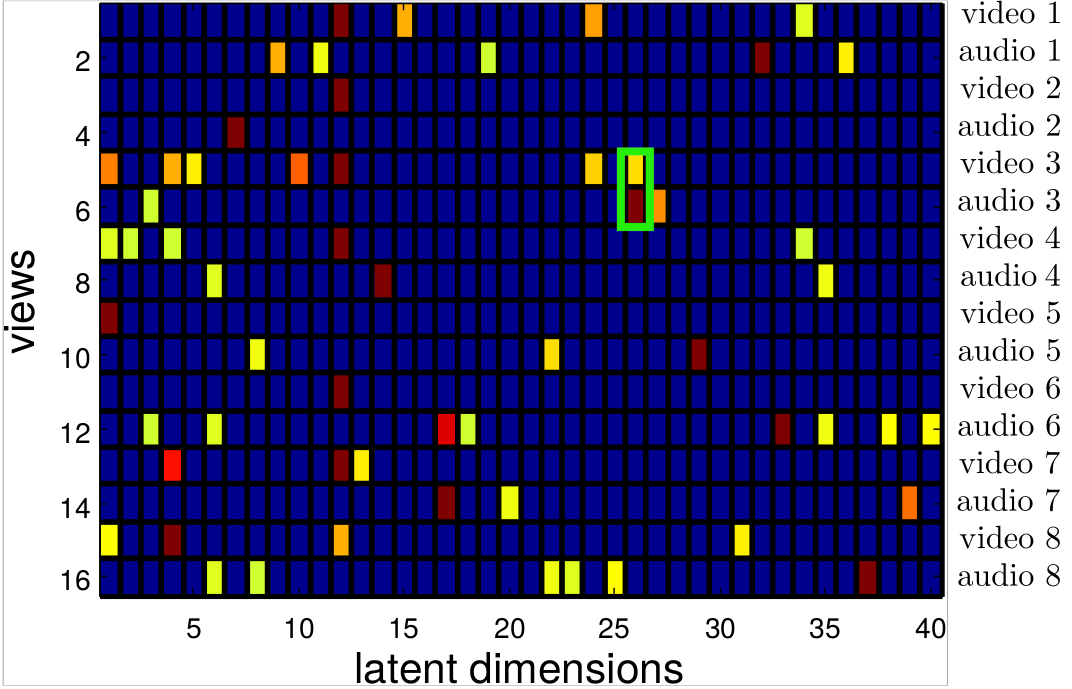}
\caption[MRD, ``AVletters'' experiment 1 (16 views): optimized ARD weights.]{
The optimized weights for the first version of the AVletters experiment represented as a heat-map. ``Warm'' (red) colors on column $i,j$ indicate a large weight for latent dimension $j$ and modality $i$. Notice that for visualization of these weights we normalized them to be between 0 and 1 and used a threshold so that $w_{i,j}<\varepsilon$ (with $\varepsilon \rightarrow 0$) was set to zero. The green box highlights a shared latent space (dimension 26) for views 5 and 6 (subject 3).
}
\label{fig:AVletters2Scales}
\end{figure}

Depending on the desired predictive or exploratory task, different subsets of the data can be split across different views. To explore the connections and commonalities in the information encoded in different subjects, letters and type of signal (video or audio), we first performed data exploration by considering the following generic setting: we created a data set where the modalities were split across all subjects and across type of signal. We only considered 8 of the subjects. Thus, we ended up with $16$ different modalities, where modalities $i, i+1$ contained the video and audio signal respectively for the $i-$th subject. The alignment was therefore made with respect to the different letters. We used all three available trials but letters ``B'', ``M'' and ``T'' were left out of the training set completely to be used at test time. For each modality, we thus had 69 rows (23 letters $\times$ 3 trials). The split across instances and modalities is summarized in Table \ref{tab:AVlettersExperiment1}. In the test set, each modality had only 9 rows (3 letters $\times$ 3 trials). Notice that this is a rather extreme scenario: the number of training instances is only 4.3 times larger than the number of modalities. We applied MRD to reveal the strength of commonality between signal corresponding to different subjects and to different recording type (video/audio). The visualization of the ARD weights can be seen in \fig \ref{fig:AVletters2Scales}.

\begin{table}
\begin{footnotesize}
\begin{center}
\begin{tabular}{r|*6{c|}*1{c}}
 \multicolumn{2}{c}{} &         view 1  &         view 2  & $\cdots$ & view 15    &        view 16 \\
 \multicolumn{2}{c}{} & (subj.1, video) & (subj.1, audio) &          & (subj.8, video) & (subj.8, audio)\\
\hline \hline
1 & `A' trial 1 & & & & &  \\
2 & `A' trial 2 & & & & & \\
3 & `A' trial 3 & & & & & \\
$\vdots$ & $\cdots$ & & & & & \\
$\nN\!=\!69$ & `Z' trial 3 & & & & & \\ \hline
\end{tabular}
\end{center}
\end{footnotesize}
\caption[View/row split for the first AVletters experiment.]{View/row split for the first AVletters experiment. This is only an illustration, thus cells are empty. Column $k$ represents $\bfY^{(k)}$ and row $i$ represents $[\bfy^{(1)}\n, \dots, \bfy^{(K)}\n]$.
}
\label{tab:AVlettersExperiment1}
\end{table}

This figure shows that similar weights (particularly in dimension $12$) are typically found for modalities $1, 3, 5, ...$, \ie the ones that correspond to the video signal. This visualization is instructive, as it reveals that to predict the lip movements in a test scenario, the other pieces of information that can help the most in this prediction is the lip movements of the rest of the subjects. We can also draw other sorts of conclusions from this kind of data exploration. For example, we can see that the subject number $3$ is the one that has the best variance alignment between the video and audio signal. This can be understood by observing that the $5$th and $6$th row of the matrix in \fig \ref{fig:AVletters2Scales} share some weights (highlighted with a green box), \ie modality $5$ and $6$ share a latent subspace.

\subsubsection*{Generation Task}
Given the above analysis, we attempted to recover the lip movements of subject $3$ for uttering the three test letters ``B'', ``M'', ``T''. The given information was the audio signal of this subject as well as the video signal of all the rest (corresponding to the same letters). The RMSE error for MRD was  $0.3$ while for NN it was $0.35$.

\subsubsection*{Incorporating Labels}
\begin{figure}[ht]
\begin{center}
  \includegraphics[width=0.65\textwidth]{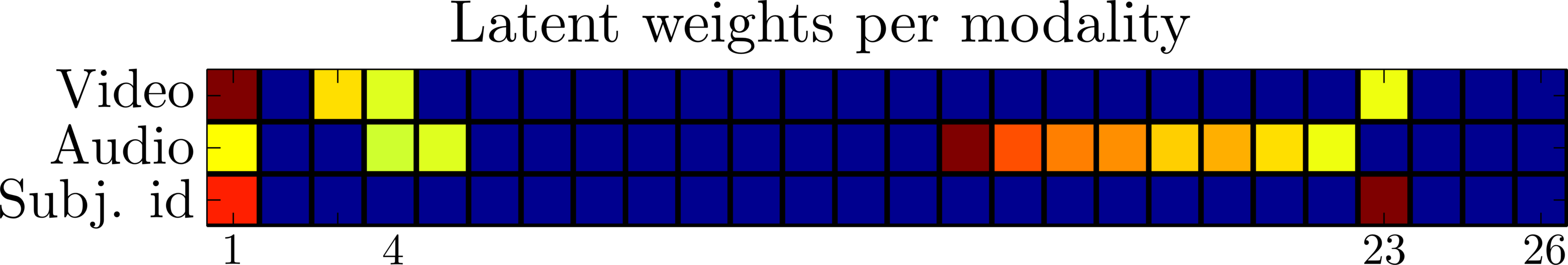}
\end{center}
\vspace{-0.1cm}
\caption[MRD, ``AVletters'' experiment 2 (3 views): optimized ARD weights.]{
The optimized weights for the second version of the AVletters experiment represented as a heat-map which has the same format as for \fig \protect\ref{fig:AVletters2Scales}.}
\label{fig:AVletters1Scales}
\end{figure}

We subsequently considered a different scenario for modelling the AVletters data base in which we also wanted to include some sort of label information, following the promising results of the discriminative model detailed in Section \ref{sec:oilData}. Specifically,  we selected the utterances of the first three subjects for the first two trials and for letters A to Q and constructed three views as follows; view 1 contained the audio signal by stacking the relevant information for all considered subjects, trials and letters. Similarly, view 2 contained the video signal. Each row in views 1 and 2 corresponds to one of three subjects, and to encode this information we use a third view. Thus, view 3 contains the discrete labels $C \in \{000, 010, 100\}$ which specify the subject identity.  This construction resulted in a training set of $102$ data points per view (3 subjects $\times$ 17 letters $\times$ 2 trials) and is summarized in Table \ref{tab:AVlettersExperiment2}.
Therefore, the subject identity is directly encoded by a single view containing discrete labels corresponding to each row of all the other views. This comes in contrast to the representation described in the previous paragraph, where the subject identity was implicitly encoded by having a separate view per subject.
The ordering of the rows in the three views does not matter, as long as the same permutation is applied to all three views.
\begin{table}[ht]
\begin{footnotesize}
\begin{center}
\begin{tabular}{r|c||*2{c|}*1{c}}
\multicolumn{2}{c}{} & View 1 & View 2 & View 3\\ \hline \hline
1 & subj 1, `A', trial 1 & audio & video & \texttt{001}\\
2 & subj 1, `A', trial 2 & audio & video & \texttt{001}\\
$\vdots$ & $\cdots$ & $\cdots$ & $\cdots$ & $\cdots$ \\
34 & subj 1, `Q', trial 2 & audio & video & \texttt{001}\\
35 & subj 2, `A', trial 1 & audio & video & \texttt{010}\\
$\vdots$ & $\cdots$ & $\cdots$ & $\cdots$ & $\cdots$ \\
$\nN\!=\!102$ & subj 3, `Q', trial 2 & audio & video & \texttt{100}\\ \hline
\end{tabular}
\end{center}
\end{footnotesize}
\caption[View/row split for the second `AVletters' experiment.]{View/row split for the second `AVletters' experiment.
}
\label{tab:AVlettersExperiment2}
\end{table}
With the above setting we are, thus, interested in modelling the commonality between the two types of signal for each person with regards to global characteristics, like accent, voice, lip movement and shape. This is because the data were aligned across modalities so that audio and video utterances were matched irrespective of the pronounced letter.
The factorization learned by MRD is shown by the optimized weights in Figure \ref{fig:AVletters1Scales}.

 As can be seen, in this scenario the video and audio signals only share one latent dimension ($4$) while the subject identifier modality shares weights with both of the other two signals (dimensions $1$ and $23$). This means that, given this small training set, MRD manages to reveal the commonality between the audio and video signals and, at the same time, learn to differentiate between subjects in both the video and the audio domain. To further demonstrate the power of MRD in representation learning, we attempted to transfer information between views. To do that, we created a test set in a similar way as for the training one but for the third trial of the recordings. From this test data set we attempted to predict the video signal in two ways: firstly by giving only the corresponding audio signal and, secondly, by giving both the audio and subject identity information. For comparison, we used nearest neighbour and standard Gaussian process regression, as can be seen in Table \ref{table:MRD_AVletters} which presents the corresponding RMSE.

\begin{table}[h]
\begin{center}
  \begin{tabular}{ c | c || c | c | c}
    \hline
    Given         & Predicted &   MRD &    NN &  GP  \\ \hline \hline
    audio         & video     & 0.498 & 0.485 & 0.494 \\ \hline
    audio, labels & video     & 0.434 & 0.485 & 0.472 \\ \hline
  \end{tabular}
\end{center}
  \caption[RMSE for test predictions on ``AVletters'': MRD and comparisons.]{RMSE of predicting the video information of test data, given only the audio or the audio and subject id information.}
  \label{table:MRD_AVletters}
\end{table}

Notice that the model could also end up with a completely separate space for the third modality (labels); the fact that it didn't means that the way in which video and audio are associated in this data set is also dependent on the subject, something which is expected, especially since two of the subjects are females and one is male. One can see from Table \ref{table:MRD_AVletters} that when MRD is given the label information, it can disambiguate better in the prediction phase and produce smaller error.

Finally, we tested the model in the opposite direction: specifically, we presented the video and audio signal of the test (third) trial of the recordings to the model and tried to recover the identity of the subject. To make this more challenging, we randomly erased $50\%$ of the information from each of the given views. The vast dimensionality of the given space did not allow us to compare against a standard GP regression model, so we only compared against nearest neighbour. The result was  an F-measure\footnote{The F-measure is given by $F_1 = 2 \cdot \frac{\text{precision} \cdot \text{recall}}{\text{precision} + \text{recall}} = \frac{2 \  \cdot \text{ true positives }}{2\  \cdot \text{ true positives } + \text{ false negatives } + \text{ false positives}}.$} of $0.92$ for MRD compared to $0.76$ for NN.

\subsection{Automatic Correlation Learning of Output Dimensions\label{sec:FI-MRD}}

GP-LVM-based models make the assumption that all dimensions of the latent space, $\{ \xq \}_{\jq=1}^\nQ$, affect all output dimensions, $\{ \yd \}_{\jd=1}^\nD$, since a single GP mapping is used\footnote{More precisely, multiple GP mappings with \emph{shared parameters} are used.}. In MRD we make conditional independence assumptions for \emph{subspaces} of $\mX$, so that different views are generated by different GP mappings. In this section, we consider the extreme case where we want to learn automatically \emph{all} conditional independencies of a multivariate data set's dimensions. In some sense, this is a means of simultaneously performing graphical model learning (learn conditional independencies) and manifold learning (learn manifolds). To achieve this goal, we reformulate the observed data set $\mY \in \Re^{\nN \times \nD}$ so that each dimension $\yd \in \Re^{\nN}$ constitutes a separate, one-dimensional view of the data. Then, following the MRD formulation, we consider a single latent space $\mX$ which encapsulates output correlations through the $\nD$ sets of ARD weights. These weights constitute the hyperparameters of $\nD$ separate, independent Gaussian processes.

The above formulation, from now on referred to as \emph{fully independent MRD (FI-MRD)}, allows us to discover correlations of the output dimensions via the latent space. We have already seen examples of this task in \fig \ref{fig:yale6SetsInterpolation}, where we sampled from specific latent dimensions and observed the obtained variance in the outputs. A similar task could be achieved with the motion capture data set of \fig \ref{fig:humanPoseAmbiguity}. For example, we might observe that a specific latent dimension is responsible for encoding the variance in both legs' movements. In this section we are interested in discovering this kind of effects \emph{automatically} (without the need to sample) and also by considering all possible \emph{subsets} of latent dimensions at the same time. This task reminds the objectives of multi-output Gaussian process inference \citep[see \eg][]{Alvarez:efficient10}, according to which the GP output function correlations are sought to be modelled explicitly by defining special covariance functions. However, in our framework the inputs to the covariance functions are latent, and the output correlations are rather captured by defining a special noise model.

To test this model, we considered again the motion capture data set introduced in Section \ref{sec:MRD_experiment_poseSilhouette} (without the silhouette views). We used a smaller subset of $120$ frames that represent $4$ distinct walking motions (two facing to the North, one facing to the South, and a semi-circular motion). In this motion capture data set, the human subjects are represented in the $3$D coordinate space. That is, each of the $21$ body joints is represented as a $3$ dimensional vector, corresponding to the degrees of freedom along the $(x,y,z)$ axes. Therefore, we have $\nD=63$ columns in the original data set which we use to form $63$ single-dimensional views. The motion is centered with respect to the root node, so that the subject moves in place.
Each view has its own set of ARD weights, but since every 3 views (and thus every 3 sets of weights) correspond to the same joint (for different degrees of freedom), we can group these 63 weight sets as
$\vw^{(j)}_{xyz} = [\vw^{(j)}_x \vw^{(j)}_y \vw^{(j)}_z], j=1, \cdots, 21$.
Notice, however, that each of the 63 sets of weights is a priori independent and learned separately, we just group them by 3 according to their corresponding joints \emph{after optimization}, just for performing our data analysis (explained below) more easily. For the data analysis, we perform $\mathcal{K}$-means clustering on the 21 vectors $\vw^{(j)}_{xyz}, j=1, ..., 21$, where each vector has dimensionality $3\nQ \times 1$. In this way, we can investigate which parts of the latent space are operating together to control a particular joint.

\begin{figure}[t]
\begin{center}
  \includegraphics[width=1\textwidth]{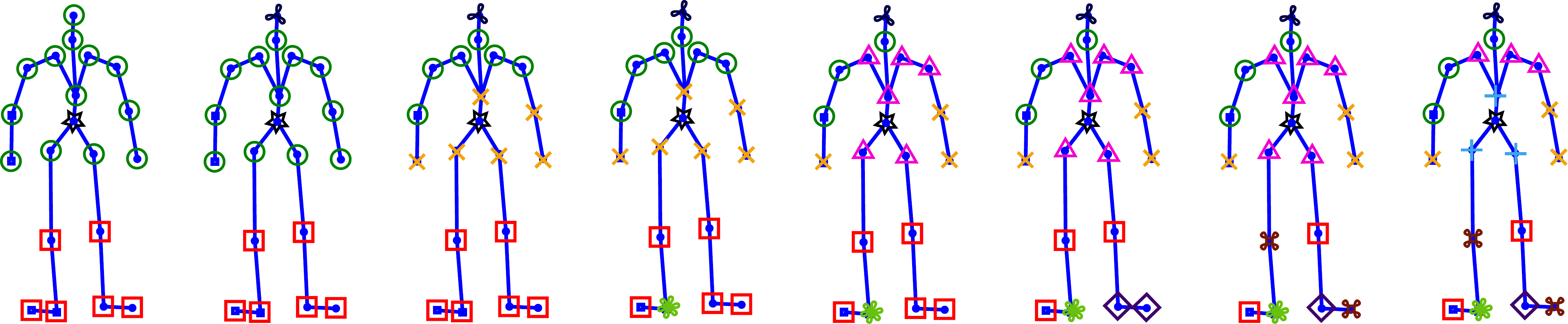}
\end{center}
\caption[Fully independent MRD experiment.]{In the fully independent MRD experiment we have $21$ joints $\times$ $3$ degrees of freedom ($x$,$y$,$z$ coordinates) $= 63$ one-dimensional views. The 3 sets of ARD weights for each joint are clustered in $\mathcal{K}$ clusters. The figure shows the cluster assignments for each joint, for $\mathcal{K}=3$ (far left) to $\mathcal{K}=10$ (far right). Each cluster is represented as a separate combination of symbol/color and corresponds to output dimensions controlled by similar latent subspaces.}
\label{fig:MRD_multOutput}
\end{figure}

The $\nQ=10$ dimensional latent space was not constrained with dynamics, and was initialized by performing PCA on the whole $\mY$ data set. As can be seen in \fig \ref{fig:MRD_multOutput} the model uncovers very intuitive correlations in the joints of the body, via the cluster assignments. For example, the model tends to represent the head with a separate latent subspace, whereas joints belonging to the same limb typically share parts of the latent space.
Moreover, the discovered clusters are similar (but not identical) to the ones obtained if we directly cluster the output dimensions corresponding to each joint (after we group them by 3, aggregating information for the $x,y,z$ coordinates). Therefore, the fully independent MRD formulation manages to maintain and uncover the cluster related properties of the original data set in the efficiently factorized latent space. Achieving this via a low-dimensional latent space is important, since very high-dimensional output spaces might not be easily clustered in the high-dimensional Euclidean space. Further, the MRD formulation allows us to transfer information between output dimensions (which here constitute separate views), a task which is typically solved (in supervised learning) using multi-output GPs. Finally, this experiment highlights the advantage of our Bayesian framework, since the number of data points is only less than double of the number of views ($120$ datapoints, $63$ views). To the best of our knowledge, MRD is the only multi-view model operating well in this kind of extreme scenarios.

\section{Conclusions}
\label{sec:conclusions}
In this paper we have presented a model capable of consolidating several different views using a single factorized latent variable. The model is fully probabilistic, handles a large variety of different data and is capable of modelling very high dimensional observations. We have shown how the model can recover the generating parameters from extremely high-dimensional data and how the factorized latent space can be used to model ambiguities in the data. Its power has been demonstrated in challenging cases of representation learning, generation of novel data, discriminative tasks and information transfer between views. We also highlighted the utility of the model in data exploration and developed the FI-MRD (Section \ref{sec:FI-MRD}) as a particular extension. 

The model is trained in a Bayesian manner using an efficient variational approximation which makes it possible to learn even from very small amounts of data and naturally allows for additional priors to be included. Compared to previous work, our approach significantly advances the state-of-the-art by equipping the standard IBFA model with two very important properties: non-linearity and non-parametric mappings. We also provided an alternative Bayesian formulation and link to IBFA, in a similar vein to the work of \citet{Klami:group11,Klami:2012tf,Damianou:2012wu}. Previous IBFA models have typically been demonstrated in scenarios with two or, more rarely, very few views compared to the number of data. To the best of our knowledge, our paper is the first to present results in cases comprising a truly large number of views. 

The presented model can be readily used in many interesting application scenarios, especially when data is scarce and comes from several noisy views that individually cannot disambiguate the task. Currently we are exploring the use of MRD in robotics and computational biology. Exploiting the typically rich prior knowledge in tasks associated with these fields is possible with future research that targets sophisticated priors for MRD. Scaling up our model to tens of thousands of datapoints is also possible, following recent work in distributed \citep{Gal:Distributed14,Dai:Parallel14} and stochastic \citep{Hoffman:2013tz,Hensman:bigdata13} variational inference. This promising direction is left for future work.

\acks{
We thank the financial support of the WYSIWYD project (EU FP7-ICT, Ref 612139). We also thank Michalis Titsias, Max Zwiessele, Zhenwen Dai for helpful discussions related to this research.
}

\appendix

\chapter


\section{Details on the Derivation of the Variational Lower Bound\label{appendix:bound}}

In this appendix we demonstrate the detailed derivation of the variational lower bound. For simplicity, we will assume that we only have two views, $\viewY \triangleq \mY^{(1)}$ and $\viewZ \triangleq \mY^{(2)}$, and the extensions to multiple views follows trivially.

As explained in the main paper, we assume Gaussian process priors on the mappings, so that:
\begin{align}
f^\viewYi \sim \mathcal{GP}(\viewy, k^\viewYi) & \Rightarrow p(\mF^\viewYi|\mX, \bftheta^\viewYi) = \prod_{\jd=1}^{\nD_\viewYi} \mathcal{N}(\vf\d^\viewYi | \bfzero, \mK^\viewYi) \nonumber \\
f^\viewZi \sim \mathcal{GP}(\viewz, k^\viewZi) & \Rightarrow p(\mF^\viewZi|\mX, \bftheta^\viewZi) = \prod_{\jd=1}^{\nD_\viewZi} \mathcal{N}(\vf\d^\viewZi | \bfzero, \mK^\viewZi) \label{GPpriorsSuppl} ,
\end{align}
where $\mK^{\{\viewYi,\viewZi\}}  = k^{\{\viewYi,\viewZi\}}(\vx, \vx')$ are the covariance matrices evaluated at the latent points.

The first step in defining a Bayesian training procedure, is to place a prior distribution $p(\mX|\bftheta_x)$ over the latent space,
where $\bftheta_x$ denotes any parameters associated with this prior.
For the moment we will not assume any particular form for this distribution and we will omit the conditioning on $\bftheta_x$. Then, the
joint distribution of the model is written as
\begin{align}
p(\viewY, \viewZ,\mF^\viewYi,\mF^\viewZi, \mX) = & p(\viewY|\mF^\viewYi) p(\mF^\viewYi|\mX) p(\viewZ|\mF^\viewZi) p(\mF^\viewZi|\mX) p(\mX) \nonumber \\
      = & p(\mX) \prod_{\jd=1}^{\nD_\viewYi}  p(\viewy\d | \vf\d^\viewYi) p(\vf\d^\viewYi | \mX)
         \prod_{\jd=1}^{\nD_\viewZi}  p(\viewz\d | \vf\d^\viewZi) p(\vf\d^\viewZi | \mX) , \label{jointSuppl}
\end{align}

\noindent where we assume independence in the data features given the latent variables.
Then, we seek to optimize the model by computing the marginal likelihood

\begin{equation}
\label{marginalLikelihood}
p(\viewY,\viewZ) =  \int_{\mX, \mF^\viewYi, \mF^\viewZi} p( \viewY | \mF^\viewYi ) p(\mF^\viewYi | \mX ) p( \viewZ | \mF^\viewZi ) p(\mF^\viewZi | \mX ) p(\mX) .
\end{equation}
 
\noindent The key difficulty with this Bayesian approach is propagating the prior
density $p(\mX)$ through the nonlinear mapping. This
mapping gives the expressive power to the model, but simultaneously
renders the associated marginal likelihood \protect\eqref{marginalLikelihood} intractable. 

We now invoke the variational Bayesian methodology to
approximate the integral. Following a standard variational inference procedure,
 we introduce a variational distribution which we assume to factorise as $q(\bfTheta)q(\mX)$ 
where $q(\mX) \sim \mathcal{N}(\bfmu, \bfS)$ and $\bfU$ denotes additional variational parameters which we will define later on. We now compute the Jensen's lower bound $\mathcal{L}_{\viewYi,\viewZi}$ on the logarithm of
\eqref{marginalLikelihood},
\begin{align}
\mathcal{L}_{\viewYi,\viewZi}
= {}&
  \int_{\mX,\bfU} q(\bfU)q(\mX) \log 
  \frac{ p(\viewY, \viewZ |\mX) p(\mX)}
     {q(\bfU)q(\mX)}  \label{jensens0Suppl} \\
= {}& 
  \int_{\mX,\bfU, \mF^\viewYi, \mF^\viewZi} q(\bfU)q(\mX) \log \left(
  \frac{p( \viewY | \mF^\viewYi ) p(\mF^\viewYi | \mX ) p( \viewY | \mF^\viewZi ) p(\mF^\viewZi | \mX )}{q(\bfU)}
        \frac{p(\mX)}{q(\mX)} \right) \nonumber \\
 ={}&
      \int_{\mF^\viewYi,\mX,\bfU} q(\bfU) q(\mX) \log \frac{p( \viewY | \mF^\viewYi ) p(\mF^\viewYi | \mX )}{q(\bfU)}   \nonumber \\
 +{}& \int_{\mF^\viewZi, \mX,\bfU } q(\bfU) q(\mX) \log \frac{p( \viewY | \mF^\viewZi ) p(\mF^\viewZi | \mX )}{q(\bfU)}  \nonumber \\
 -{}& \int_{\bfX} \cancel{q(\bfU)} q(\mX) \log \frac{q(\mX)}{p(\mX)} \nonumber \\
={}& \tilde{\mathcal{L}}_\viewYi + \tilde{\mathcal{L}}_\viewZi - \KL{q(\mX)}{p(\mX)},
     \label{jensens1Suppl}
\end{align}

\noindent where $\boldsymbol \theta$ denotes the model's parameters $\bftheta = \{\bftheta^\viewYi, \bftheta^\viewZi\}$.  However,
the above form of the lower bound is problematic because $\mX$ (in the
GP terms $p(\mF^\viewYi|\mX)$ and $p(\mF^\viewZi|\mX)$) appears non-linearly inside the kernel matrices
$\mK^\viewYi$ and $\mK^\viewZi$ of equation \eqref{GPpriorsSuppl}, making the integration over $\mX$ difficult.
It is, thus, obvious that standard mean field variational methodologies do not lead to an analytically
tractable algorithm.

In contrast, our framework allows us to compute
a closed-form Jensen's lower bound
by applying variational inference after expanding the GP prior so as to include auxiliary inducing
variables. Originally, inducing variables were introduced for computational speed ups in GP regression models.
 In our approach, these extra variables are used as in the variational sparse GP method of 
\cite{Titsias:2009vf}. More specifically, we expand the joint
 probability model in equation \protect\eqref{jointSuppl} with $\nM$ extra samples $\mU^\viewYi$ and $\mU^\viewZi$ of the latent functions $\vf^\viewYi$ and $\vf^\viewZi$ evaluated at a set of pseudo-inputs
(known as ``inducing points'')
$\bfZ^\viewYi$ and $\bfZ^\viewZi$ respectively. Here, 
$\mU^\viewYi \in \mathbb{R}^{\nM_\viewYi \times \nD_\viewYi}$, $\mU^\viewZi \in \mathbb{R}^{\nM_\viewZi \times \nD_\viewZi}$,
 $\bfZ^\viewYi \in \mathbb{R}^{\nM_\viewYi \times \nQ}$,  $\bfZ^\viewZi \in \mathbb{R}^{\nM_\viewZi \times \nQ}$
and $\nM = \nM_\viewYi+\nM_\viewZi$. Typically we select $\nM \ll \nN$ to also gain in computational speed.

 The augmented joint probability density takes the form
 \begin{align}
 p(\viewY,\viewZ,\mF^\viewYi, \mF^\viewZi, \mU^\viewYi, \mU^\viewZi, \mX | \bfZ^\viewYi,\bfZ^\viewZi) =
    p(\mX) & \prod_{\jd=1}^{\nD_\viewYi} p(\viewy\d | \vf\d^\viewYi) p(\vf\d^\viewYi | \bfu\d^\viewYi, \mX) p(\bfu\d^\viewYi | \bfZ^\viewYi)  & \nonumber \\
         & \prod_{\jd=1}^{\nD_\viewZi} p(\viewz\d | \vf\d^\viewZi) p(\vf\d^\viewZi | \bfu\d^\viewZi, \mX) p(\bfu\d^\viewZi | \bfZ^\viewZi) \label{augmentedJointSuppl}
 \end{align}
 where $p(\bfu^{\{\viewYi,\viewZi\}}\d | \bfZ^{\{\viewYi,\viewZi\}})$ are zero-mean Gaussians with
 covariance matrices $\Kuu^\viewYi$ and $\Kuu^\viewZi$ respectively, constructed using the same functions as for
 the GP priors \eqref{GPpriorsSuppl}.
  We write the augmented GP prior analytically (see
 \cite{Rasmussen:2005te}) as
 \begin{equation}
  \label{priorF2Suppl}
 p(\vf\d^\viewYi | \bfu\d^\viewYi, \mX) =  \mathcal{N}  \left( \vf\d^\viewYi | \Kfu^\viewYi \left(\Kuu^\viewYi\right)^{-1}
  \bfu\d^\viewYi , \Kff^\viewYi - \Kfu^\viewYi \left(\Kuu^\viewYi\right)^{-1} \Kuf^\viewYi \right),
 \end{equation}
 and similarly for $\viewZi$. Here, $\Kff^\viewYi = k^\viewYi(\mX,\mX)$ and $\Kfu^\viewYi$ denotes the cross-covariance between the function values of $k^\viewYi$ evaluated at
the latent points $\mX$ and the inducing points $\bfZ^\viewYi$, \ie $\Kfu^\viewYi = k^\viewYi(\mX, \bfZ^\viewYi)$.

Analogously to \cite{Titsias:2010tb}, we are now able to obtain a tractable lower bound through
the variational density:
\begin{equation}
\label{qThetaSuppl}
q(\bfU)q(\mX) = \{q(\mU^\mathcal{K}) p(\mF^\mathcal{K}|\mU^\mathcal{K},\mX,\bfZ^\mathcal{K})\}_{\mathcal{K}=\{\viewYi,\viewZi\}} q(\mX),
\end{equation} 
where $q(\mU^{\{\viewYi,\viewZi\}})$ are free form distributions and $q(\mX)$ a Gaussian with parameters $\bfmu$ and $\bfS$.
Notice also that \eqref{qThetaSuppl} factorises across dimensions.
Optimization of the variational lower bound provides an approximation to the true posterior $p(\mX|\viewY,\viewZ)$
by $q(\mX)$. 

After defining a variational distribution, we can continue our derivation by returning to
the expression for the lower bound \eqref{jensens1Suppl} and replacing the joint distribution with
 its augmented version \eqref{augmentedJointSuppl} and the variational distribution with its
 factorised version \eqref{qThetaSuppl}. Since the variational bound breaks to separate terms for each
of the observations spaces, here we will drop the subscripts $\viewYi$ and $\viewZi$ and show how, in general, we
can calculate the $\tilde{\mathcal{L}}$ terms of equation \eqref{jensens1Suppl} for a general observation space $\bfY$. We have:

\begin{align}
\tilde{\mathcal{L}} = {}& \int_{\mX, \mF, \mU} q(\bfU)q(\mX) \log 
    \frac{ p(\mY,\mF, \mU | \mX, \bfZ)}
       {q(\bfU)}  \nonumber \\
= {}& \int_{\mX,\mF,\mU} \prod_{\jd=1}^\nD p(\vf\d | \bfu\d, \mX, \bfZ)q(\bfu\d)
      \log  \frac{\prod_{\jd=1}^\nD p(\yd | \vf\d) \cancel{p(\vf\d | \bfu\d, \mX, \bfZ)}
            p(\bfu\d | \bfZ)}
   {\prod_{\jd=1}^\nD \cancel{p(\vf\d | \bfu\d, \mX, \bfZ )}q(\bfu\d) q(\mX)}  \nonumber \\
 ={}& \int_{\mX, \mF, \mU} \prod_{\jd=1}^\nD p(\vf\d | \bfu\d, \mX, \bfZ )q(\bfu\d) q(\mX) 
    \log  \frac{\prod_{\jd=1}^\nD p(\yd | \vf\d) p(\bfu\d | \bfZ)}
           {\prod_{\jd=1}^\nD q(\bfu\d) q(\mX))}   \label{eq:Lsuppl0}.
\end{align}

\noindent Dropping $\bfZ$ from our expressions, for simplicity, we finally obtain:
\begin{equation}
\label{LSuppl}
\tilde{\mathcal{L}} =
\sum_{\jd=1}^\nD \left( 
    \int_{\bfu\d, \mX} q(\bfu\d) q(\mX) \mathbb{E}_{p(\vf\d | \bfu\d, \mX)}\left[ \log p(\yd | \vf\d) \right]  +
             \log \mathbb{E}_{q(\bfu\d)}\left[ \frac{p(\bfu\d)}{q(\bfu\d)} \right] 
  \right) = \sum_{\jd=1}^\nD \tilde{\mathcal{L}}\d ,
\end{equation}

\noindent Calculating \eqref{LSuppl} in the same manner for every available observation space and 
replacing back in the variational bound \eqref{jensens1Suppl} 
we obtain the final form of the bound which is now analytically tractable. 
In particular, the $\text{KL}$ term
is tractable and easy for certain priors $p(\mX)$.
As for the $\tilde{\mathcal{L}}$ terms, we can calculate the
expectation over $p(\vf\d | \bfu\d,\mX)$ and reveal that the optimal setting for $q(\bfu\d)$ is 
also a Gaussian. More specifically, we have:
\begin{align}
\tilde{\mathcal{L}}\d={}& \int_{\bfu\d} q(\bfu\d) \log \frac{e^{\mathbb{E}_{q(\mX)}\left[ \log N \left( \yd | \bfa\d, \beta^{-1} \bfI\d \right) \right]}
    p(\bfu\d)}{q(\bfu\d)}  - \bfC , \label{boundFAnalytically5}
\end{align}
where $\bfa\d$ is the mean of \eqref{priorF2Suppl} and 
$\bfC=\frac{\beta}{2} \text{Tr}(\mathbb{E}_{q(\mX)}\left[ \Kff \right]) +
    \frac{\beta}{2} \text{Tr} \left(\Kuu^{-1} \mathbb{E}_{q(\mX)}\left[ \Kuf \Kfu \right] \right) $.
The expression in \eqref{boundFAnalytically5} is a KL-like quantity and, therefore, $q(\bfu\d)$ is optimally set to be the quantity 
appearing in the numerator of the above equation. So:

\begin{equation}
\label{qu}
q(\bfu\d) = e^{\mathbb{E}_{q(\mX)}\left[ \log \mathcal{N} \left( \yd | \bfa\d, \beta^{-1} \bfI\d \right) \right]}
    p(\bfu\d) ,
\end{equation}
exactly as in \cite{Titsias:2010tb}. This is a Gaussian distribution since we have assumed
$p(\bfu\d ) = \mathcal{N} (\bfu\d | \mathbf{0}, \Kuu )$.

After replacing $q(\bfu\d)$ with its optimal value, we can reverse Jensen's inequality (\ie substitute the optimal form back inside the bound) to obtain:

\begin{equation}
\label{boundFAnalyticallyFinalIntegralSuppl}
\tilde{\mathcal{L}}\d \geq
  \log \int_{\bfu\d} e^{\mathbb{E}_{q(\mX)} \left[ \log N \left( \yd | \bfa\d, \beta^{-1} \bfI\d \right) \right]}
    p(\bfu\d) -\bfC .
\end{equation}

\noindent Notice that the expectation appearing above is a standard Gaussian integral and \eqref{boundFAnalyticallyFinalIntegralSuppl} can
be calculated in closed form, which turns out to be:
\begin{equation}
\label{LFinalSuppl}
\tilde{\mathcal{L}}\d(q, \boldsymbol \theta) \geq \log \left[ 
  \frac{(\beta)^{\frac{N}{2}} \vert \Kuu \vert ^\frac{1}{2} }
     {(2\pi)^{\frac{N}{2}} \vert \beta \boldsymbol \Phi + \Kuu  \vert ^\frac{1}{2} } 
   e^{-\frac{1}{2} \bfy_{d}^\top \bfW \yd}
   \right]   -
   \frac{\beta \psi}{2} + \frac{\beta}{2} 
   \text{Tr} \left( \mathit{\Kuu^{-1}} \boldsymbol \Phi \right) 
\end{equation}

\noindent where:
\begin{equation}
\label{psis}
\psi = \text{Tr}(\mathbb{E}_{q(\mX)}\left[ \Kff \right]) \;, \;\;
\boldsymbol \Psi = \mathbb{E}_{q(\mX)}\left[ \Kfu \right] \;, \;\;
\boldsymbol \Phi = \mathbb{E}_{q(\mX)}\left[ \Kuf \Kfu \right]
\end{equation}

\noindent and $\bfW = \beta \bfI - \beta^2 \boldsymbol \Psi (\beta \boldsymbol \Phi + \Kuu)^{-1} \boldsymbol \Psi^\top$. This expression is straight forward to compute, as long as the covariance functions $k^\viewYi$ and $k^\viewZi$
 are selected so that the $\{\psi, \boldsymbol \Psi, \boldsymbol \Phi \}$ quantities of equation \eqref{psis} can be computed analytically. As shown in \cite{Titsias:2010tb}, these
statistics constitute convolutions of the covariance function with Gaussian densities and 
are tractable for many standard covariance functions, such as the ARD squared exponential or the linear one.

Given the above, we obtain the final variational lower bound of equation \eqref{eq:boundCollapsed} by computing equation \eqref{LSuppl} for each modality (this equation is a summation of the terms of equation \protect\eqref{LFinalSuppl}) and subtracting the KL term as explained in equation \eqref{jensens1Suppl}.

\section{Inferring a New Latent Point\label{appendix:latentPosterior}}

Given a model which is trained so as to jointly represent two output spaces $\viewY$ and $\viewZ$ with
a common but factorised input space $\mX$, we wish to generate a new (or infer a training) set of outputs
$\viewZ\Ts \in \mathbb{R}^{\nN_* \times \nD_\viewZi}$ given a set of (potentially partially) observed test points $\viewY\Ts \in \mathbb{R}^{\nN_* \times \nD_\viewYi}$.
This is done in three steps, as explained in the main paper. Here we explain in more detail the first step, where we need to predict
 the set of latent points $\mX^* \in \mathbb{R}^{\nN_* \times \nQ}$
which is most likely to have generated $\viewY\Ts$. 

To achieve this, we use an approximation to the posterior marginal $p(\mX\Ts|\viewY\Ts,\viewY)$,
which has the same form as for the standard Bayesian GP-LVM model \citep{Titsias:2010tb} and is given by a variational distribution
 $q(\mX\Ts)$ which, in turn, is a marginal of $q(\mX,\mX\Ts)$. To find $q(\mX,\mX\Ts)$ we optimise a variational lower bound $\mathcal{L}_{\viewYi,*}  \left(q(\mX,\mX^*) \right)$
on the marginal likelihood $p(\viewY,\viewY\Ts)$ which has analogous form with the training
  objective function \eqref{eq:boundEnd}. In specific, we ignore $\viewZ$ and replace $\viewY$ with $(\viewY,\viewY\Ts)$ and $\mX$ with $(\mX,\mX\Ts)$ in \eqref{jensens0Suppl} so as to get:
\begin{align}
 \mathcal{L}_{\viewYi, *} 
   & = \int_{\mX\Ts, \mX} p(\viewY\Ts,\viewY | \mX\Ts,\mX) p(\mX\Ts,\mX) \nonumber \\
   & \leq \int_{\mX, \mX\Ts,\bfU} q(\mX\Ts,\mX) q(\bfU) \log 
      \frac{p(\viewY\Ts,\viewY | \mX\Ts,\mX) p(\mX\Ts,\mX)}{q(\mX\Ts,\mX) q(\bfU)}  . \label{suppl:predictionsBound}
\end{align}

\noindent This variational lower bound is computed exactly as described in Appendix \ref{appendix:bound}.

What now remains is to define $q(\mX\Ts,\mX)$. At this step, the inference procedure differs depending on the type of prior used for the latent space $\mX$.
Specifically, if we use a prior that does not couple datapoints, such as a standard normal one, then we are allowed to
write that $q(\mX,\mX\Ts) = \prod_{\jn=1}^\nN q(\xn) \prod_{\n=1}^{\nN_*} q(\vx\Ts)$, 
where $q(\vx\Ts) = \mathcal{N}(\vx\Ts | \bfmu\nTs, \bfS\nTs)$. In this case, equation \eqref{suppl:predictionsBound} can be broken into a sum of three terms:
\begin{equation}
\label{suppl:predictionsBoundTerms}
\begin{aligned}
\mathcal{L}_{\viewYi, *} 
& \le \int_{\mX\Ts,\mX,\bfU} q(\mX\Ts)q(\mX)q(\bfU)\log\frac{p(\viewY\Ts,\viewY|\mX\Ts,\mX)}{q(\mX\Ts)q(\mX)q(\bfU)} \\
&- \KL{q(\mX\Ts)}{p(\mX\Ts)} - \KL{q(\mX)}{p(\mX)}.
\end{aligned}
\end{equation}

\noindent The last term is already computed during training time and does not need to be re-computed. The middle term can also be computed cheaply and in parallel. As for the first term, it has the same form as for equation \eqref{LFinalSuppl} but augmented with the test data. Importantly, the expensive computations in this expression are in the $\{\psi,\bfPsi,\bfPhi\}$ statistics. However, these statistics are decomposable across data-points, meaning that we can re-use computations (expectations over $q(\mX)$) performed during training time.

On the other hand, performing inference in the dynamical model is more challenging, since $q(\mX\Ts,\mX)$ is fully
coupled across $\mX$ and $\mX\Ts$. Therefore, if we wish to maintain the correlation of the inputs depending on their times,
we should select this distribution to only factorise across features: 
$q(\mX\Ts,\mX) = \prod_{\jq=1}^\nQ  \mathcal{N} (\bfx\qTs | \bfmu\qTs ,\bfS\qTs)$,
 where $\bfS\inq{q}{n}$ are $(\nN+\nN_*) \times (\nN+\nN_*)$ matrices. In this case, the predictive equation \eqref{suppl:predictionsBound} will not break as in equation \eqref{suppl:predictionsBoundTerms} and, therefore, the computational complexity is increased.

\section{Top-down Predictions\label{appendix:topdown}}

In this section we describe the prediction of an output point in some modality $k$, given latent test posterior marginal $q(\mX\Ts)$ obtained as explained in Appendix \ref{appendix:latentPosterior}. For simplicity, we will drop the superscsript denoting a specific modality and, instead, refer to a generic output space $\bfY$. The quantity of interest is:
\begin{align}
p(\mY) 
& \approx \int_{\mF\Ts} p(\mY\Ts | \mF\Ts) \int_{\mU,\mX\Ts} p(\mF\Ts|\mU,\mX\Ts)q(\mU)q(\mX\Ts) \nonumber \\
& = \int_{\mF\Ts } p(\mY\Ts | \mF\Ts) q(\mF\Ts) \label{eq:qY}
\end{align}

\noindent where $q(\mF\Ts)$ is found as a product of its dimensions with:
\begin{equation}
q(\vf\dTs) = \int_{\vu\d, \mX\Ts} p(\vf\dTs | \vu\d, \mX\Ts)q(\vu\d)q(\mX\Ts) .
\end{equation}

\noindent The above is found by first using the Gaussian $q(\mU)$ and equation \eqref{priorF2Suppl} in order to find the intermediate result $q(\mF\Ts|\mX\Ts)$ and then getting the final result following \citet{Girard:uncertain01} to be:
\begin{align}
 \mathbb{E}(\mF\Ts) ={}&  \bfB^\top \bfPsi\Ts \label{meanFstar} \\
 \text{Cov}(\mF\Ts) ={}& \bfB^\top \left( \bfPhi\Ts - \bfPsi\Ts \bfPsi\Ts^\top \right) \bfB + \psi\Ts \eye - \text{tr} \left( \left( \Kuu^{-1} - \left( \Kuu + \beta \bfPhi \right)^{-1} \right) \bfPhi\Ts \right) \eye , \; \; \label{covFstar}
\end{align}

\noindent where $\psi\Ts = \tr{\mathbb{E}_{q(\mX\Ts)} \left[ \mK_{**} \right]}$, $\bfPsi\Ts = \mathbb{E}_{q(\mX\Ts)} \left[ \mK_{u*} \right]$, $\bfPhi\Ts = \mathbb{E}_{q(\mX\Ts)} \left[ \mK_{u*} \mK_{u*}^\top \right]$. Further, $\bfB = \beta \left( \Kuu + \beta \bfPhi \right)^{-1} \bfPsi^\top \mY$,
 $\mK_{**} = k(\mX\Ts, \mX\Ts)$ and $\mK_{u*} = k(\bfZ,\mX\Ts)$. 
Notice that the above is a moment-matching approach: the true distribution for $q(\mF\Ts)$ is intractable, but all of its moments are analytic (above we computed only the mean and variance). Replacing \protect\eqref{meanFstar} and \protect\eqref{covFstar} into \protect\eqref{eq:qY} we have that  
the predicted mean of $\mY\Ts$ is equal to $\mathbb{E}\left[ \mF\Ts \right]$
and the predicted covariance (for each column of $\mY\Ts$) is equal to 
$\text{Cov}(\mF\Ts) + \beta^{-1} \eye_{\nN_*}$.

\bibliography{carl,../../../bib/lawrence,../../../bib/other,../../../bib/zbooks}

\end{document}